\documentclass{article}

\usepackage{appendix}
\usepackage[sort, authoryear, round]{natbib}
\usepackage{fancyhdr}
\usepackage[format=plain,
            labelfont={it,it},
            textfont=it]{caption}
\usepackage[stable, bottom, flushmargin]{footmisc}
\usepackage{graphicx}
\usepackage{xcolor}    
\usepackage{breakcites}
\usepackage{amsmath,amssymb,amsthm}
\usepackage[ruled,vlined]{algorithm2e}

\definecolor{lightblue}{RGB}{0,102,204}
\usepackage[pagebackref, colorlinks=true, linkcolor=lightblue, citecolor=lightblue, urlcolor=lightblue]{hyperref}

\usepackage{pgfplots}
\pgfplotsset{compat = newest}
\usepackage{wrapfig}

\usepackage{enumerate}
\usepackage{xparse, expl3}
\usepackage{indentfirst}
\usepackage{float}
\usepackage{tikz}
\usepackage{tikzscale}
\usetikzlibrary{arrows.meta,arrows}
\usetikzlibrary{decorations.pathreplacing}
\usetikzlibrary{positioning,chains,fit,shapes,calc,shapes.misc}
\usetikzlibrary{arrows.meta,positioning,shapes.geometric,calc}
\usepackage{amsfonts}
\usepackage{amsmath}
\usepackage[capitalize]{cleveref}

\usepackage{todonotes}

\newtheorem{theorem}{Theorem}[section]

\newtheorem{lemma}[theorem]{Lemma}

\newtheorem{remark}[theorem]{Remark}

\newtheorem{problem}[theorem]{Problem}

\crefname{theorem}{Theorem}{Theorems}
\crefname{proposition}{Proposition}{Propositions}
\crefname{lemma}{Lemma}{Lemmas}
\crefname{corollary}{Corollary}{Corollaries}
\crefname{definition}{Definition}{Definitions}
\crefname{example}{Example}{Examples}
\crefname{remark}{Remark}{Remarks}
\crefname{algorithm}{Algorithm}{Algorithms}
\crefname{equation}{Equation}{Equations}
\crefname{section}{Section}{Sections}
\crefname{subsection}{Section}{Sections}
\crefname{conjecture}{Conjecture}{Conjectures}
\crefname{problem}{Open Problem}{Open Problem}

\usepackage{microtype}
\usepackage{hyperref}
\usepackage{graphicx}

\usepackage{mathtools}
\usepackage{mathrsfs}
\usepackage{bbm,bm}
\usepackage{xfrac}
\usepackage{nicefrac}
\usepackage{float}
\usepackage{setspace}
\usepackage{mathtools}

\newcommand{\nc}{\newcommand}
\newcommand{\rnc}{\renewcommand}

\nc{\R}{\mathbb R}
\nc{\Q}{\mathbb Q}
\nc{\Z}{\mathbb Z}
\nc{\N}{\mathbb N}

\let\E\relax
\DeclareMathOperator*{\E}{\mathbb{E}}
\let\P\relax
\DeclareMathOperator*{\P}{\mathbb{P}}
\DeclareMathOperator*{\e}{\mathbb{E}}
\DeclareMathOperator*{\p}{\mathbb{P}}

\nc{\CA}{\mathcal A}
\nc{\CC}{\mathcal C} 
\nc{\CD}{\mathcal D}
\nc{\CF}{\mathcal F}
\nc{\CG}{\mathcal G}
\nc{\CH}{\mathcal H}
\nc{\CI}{\mathcal I} 
\nc{\CS}{\mathcal S} 
\nc{\CX}{\mathcal X}
\nc{\CY}{\mathcal Y}

\nc{\VC}{\mathrm{VC}}
\nc{\DS}{\mathrm{DS}}
\nc{\Nat}{\mathrm{Nat}}
\nc{\Graph}{\mathrm{Graph}}

\nc{\D}{\mathbb D}
\nc{\defn}[1]{\textbf{#1}}
\nc{\prop}{\normalfont \large \textsf{prop}}
\rnc{\t}[1]{\text{#1}}
\nc{\ERM}{\mathrm{ERM}}
\nc{\idk}{\bot}
\nc{\hmin}{h_{\mathrm{min}}}
\nc{\htie}{h_{\mathrm{tie}}}

\newenvironment{proofof}[1]{
  \renewcommand{\proofname}{Proof of {#1}}\proof}{\endproof}

\DeclareMathOperator{\ls}{\mathcal{L}}
\DeclareMathOperator*{\argmin}{arg\,min}

\newcommand{\ind}{\mathbbm{1}}
\newcommand{\erm}{\mathcal{A}}
\newcommand{\cerm}{\mathrm{ERM}}

\newcommand{\rt}{\mathbf{t}}
\newcommand{\rI}{\mathbf{I}}
\newcommand{\tie}{\mathrm{Tie}}

\newcommand{\rh}{\mathbf{\hat{h}}}

\newcommand{\cX}{\mathcal{X}}
\newcommand{\cH}{\mathcal{H}}
\newcommand{\cY}{\mathcal{Y}}
\newcommand{\cA}{\mathcal{A}}
\newcommand{\cD}{\mathcal{D}}
\newcommand{\rS}{\mathbf{S}}

\newcommand{\rx}{\mathbf{x}}

\newcommand{\cS}{\mathcal{S}}

\newcommand{\ry}{\mathbf{y}}
\newcommand{\rN}{\mathbf{N}}

\newcommand{\cur}{86}

\newcommand{\cb}{c_b}
\newcommand{\cc}{c_c}

\newcommand{\ch}{24}

\newcommand{\cj}{24}

\newcommand{\cu}{c_{u}}

\newcommand{\hs}{h^{\negmedspace\star}}
\newcommand{\sign}{\mathrm{sign}}
\newcommand{\avg}{\mathrm{avg}_{\smash{\not=}}}

\usepackage[final]{neurips_2025}

\usepackage[utf8]{inputenc} 
\usepackage[T1]{fontenc}    
\usepackage{hyperref}       
\usepackage{url}            
\usepackage{booktabs}       
\usepackage{amsfonts}       
\usepackage{nicefrac}       
\usepackage{microtype}      
\usepackage{xcolor}         

\title{On Agnostic PAC Learning in the Small Error Regime}

\author{
Julian Asilis \\ USC \\ \texttt{asilis@usc.edu}
\And
Mikael {M\o ller H\o gsgaard} \\ Aarhus University  \\ \texttt{hogsgaard@cs.au.dk} \And
Grigoris Velegkas\thanks{Part of the work was done while the author was a PhD student at Yale University.} \\ Google Research \\ \texttt{gvelegkas@google.com}
}

\begin{document}

\maketitle

\begin{abstract}
Binary classification in the classic PAC model exhibits a curious phenomenon: Empirical Risk Minimization (ERM) learners are suboptimal in the realizable case yet optimal in the agnostic case. Roughly speaking, this owes itself to the fact that non-realizable distributions $\mathcal{D}$ are more difficult to learn than realizable distributions -- even when one discounts a learner's error by $\mathrm{err}(h^*_\mathcal{D})$, i.e., the error of the best hypothesis in $\mathcal{H}$. Thus, optimal agnostic learners are permitted to incur excess error on (easier-to-learn) distributions $\mathcal{D}$ for which $\tau = \mathrm{err}(h^*_\mathcal{D})$ is small.
Recent work of Hanneke, Larsen, and Zhivotovskiy (FOCS '24) addresses this shortcoming by including $\tau$ itself as a parameter in the agnostic error term. In this more fine-grained model, they demonstrate tightness of the  error lower bound $\tau + \Omega (\sqrt{\nicefrac{\tau (d + \log(1 / \delta))}{m}} + \nicefrac{d + \log(1 / \delta)}{m} )$ in a regime where $\tau > d/m$, and leave open the question of whether there may be a higher lower bound when $\tau \approx d/m$, with $d$ denoting $\VC(\mathcal{H})$.
In this work, we resolve this question by exhibiting a learner which achieves error $c \cdot \tau + O (\sqrt{\nicefrac{\tau (d + \log(1 / \delta))}{m}} + \nicefrac{d + \log(1 / \delta)}{m} )$ for a constant $c \leq 2.1$, matching the lower bound and demonstrating optimality when $\tau =O( d/m)$. Further, our learner is computationally efficient and is based upon careful aggregations of ERM classifiers, making progress on two other questions of Hanneke, Larsen, and Zhivotovskiy (FOCS '24). We leave open the interesting question of whether our approach can be refined to lower the constant from 2.1 to 1, which would completely settle the complexity of agnostic learning.

\end{abstract}

\section{Introduction}\label{Section:Introduction}
\vspace{-0.2 cm}

Binary classification stands as perhaps the most fundamental setting in supervised learning, and one of its best-understood. In Valiant's celebrated Probably Approximately Correct (PAC) framework, learning is formalized using a domain $\CX$ in which unlabeled data points reside, the label space $\CY = \{\pm 1\}$, and a probability distribution $\CD$ over $\CX \times \CY$. The purpose of a learner $\CA$ is to receive a training set $S = (x_i, y_i)_{i=1}^m$ of points drawn i.i.d.\ from $\CD$ and then emit a hypothesis $\CA(S) : \CX \to \CY$ which is unlikely to mispredict the label of a new data point $(x, y)$ drawn from $\CD$.

At the center of a learning problem is a hypothesis class of functions $\CH \subseteq \CY^\CX$, with which the learner $\CA$ must compete. More precisely, $\CA$ is tasked with emitting a hypothesis $f$ whose \emph{error}, denoted $\ls_\CD(f) := \P_{(x, y) \sim \CD} \big( f(x) \neq y \big)$, is nearly as small as that of the best hypothesis in $\CH$. We assume for simplicity such a best hypothesis exists and denote it as $h^*_\CD := \argmin_{h\in \CH} \ls_\CD(h)$ or when $ \cD $ is clear from context with $ h^{*} $. Of course, the distribution $\CD$ is unknown to $\CA$, and $\CA$ is judged on its performance across a broad family of allowable distributions $\D$. In \emph{realizable} learning, $\D = \{\CD : \inf_{h\in\CH} \ls_\CD(h) = 0\}$,
meaning $\CA$ is promised that there always exists a \emph{ground truth} hypothesis $h^* \in \CH$ attaining zero error.
In this case, for $\CA$ to compete with the performance of $h^*_\CD$ simply requires that it attains error $\leq \epsilon$ for all realizable distributions $\CD$ (with high probability over the training set $S \sim \CD^m$, and using the smallest amount of data $m$ possible). In \emph{agnostic} learning, $\D$ is defined as the set of all probability distributions over $\CX \times \CY$, meaning any data-generating process is allowed. Thus, $\CA$ is not equipped with any information concerning $\CD \in \D$ at the outset, but in turn it is only required to emit a hypothesis $f$ with $\ls_\CD (f) \leq \ls_\CD(h^*_\CD) + \epsilon$.

Perhaps the most fundamental questions in PAC learning ask: What is the minimum number of samples required to achieve error at most $\epsilon$? And which learners attain these rates? For realizable binary classification, the optimal sample complexity of learning remained a foremost open question for decades, and was finally resolved by breakthrough work of \citet{hanneke2016optimal} which built upon that of \citet{simon2015almost}. Notably, optimal binary classification requires the use of learners which emit hypotheses $f$ outside of the underlying hypothesis class $\CH$. Such learners are referred to as being \emph{improper}. This has the effect of excluding Empirical Risk Minimization (ERM) from contention as an optimal learner for the realizable case. Recall that ERM learners proceed by selecting a hypothesis $h_S \in \CH$ incurring the fewest errors on the training set $S$. (For realizable learning, note that $h_S$ will then incur zero error on $S$.) ERM, then, is an example of a \emph{proper} learner, which always emits a hypothesis in the underlying class $\CH$.
Agnostic learning, however, exhibits no such properness barrier for optimal learning. In fact, ERM algorithms are themselves optimal agnostic learners, despite their shortcomings on realizable distributions. This somewhat counter-intuitive behavior owes itself to the fact that agnostic learners are judged on a worst-case basis across all possible distributions $\CD$. Simply put, non-realizable distributions are more difficult to learn than realizable distributions --- even when one discounts the error term by $\ls_\CD(h^*_\CD)$ --- and thus ERM learners are permitted to incur some unnecessary error on realizable distributions, so long as they remain within the optimal rates induced by the more difficult (non-realizable) distributions.

In light of this behavior, it is natural to ask for a more refined perspective on agnostic learning, in which the error incurred beyond $\tau := \ls_\CD(h^*_\CD)$ is itself studied as a function of $\tau$. Note that this formalism addresses the previously-described issue by demanding that learners attain superior performance on distributions with smaller values of $\tau$ (such as realizable distributions, for which $\tau = 0$). {Precisely this perspective was recently studied by \citet{hanneke2024revisiting}, who established that all proper learners, including ERM, are sub-optimal for $\tau \in [ \Omega(\ln^{10}{\left(m/d \right)}(d+\ln{\left(1/\delta \right)})/m),o(1)] $, by establishing a new lower bound and an optimal learner in this regime. While settling the sample complexity for a wide range of $\tau$, \citet{hanneke2024revisiting} leave open several interesting questions.}

\begin{problem} \label{Question:1-tau-d/n-setting}
Let $d = \VC(\CH)$ and $m = |S|$. What is the optimal sample complexity of learning in the regime where $\tau \approx d / m$?
\end{problem}

\begin{problem} \label{Question:2-majority-voting}
Are learners based upon majority voting, such as bagging, optimal in the \linebreak $\tau$-based agnostic learning framework?
\end{problem}

\begin{problem} \label{Question:3-computational-efficiency}
Can one design a computationally efficient learner which is optimal in the \linebreak $\tau$-based agnostic learning framework?
\end{problem}

The primary focus of our work is to resolve Open Problem~\ref{Question:1-tau-d/n-setting}, thereby extending the understanding of optimal error rates across a broader range of $\tau$. As part of our efforts, we also make progress on Open Problems \ref{Question:2-majority-voting} and \ref{Question:3-computational-efficiency}, as we now describe.

\subsection{Overview of Main Results}

We now present our primary result and a brief overview of our approach.

\begin{theorem}\label{thm:main-result}
    For any domain $\CX,$ hypothesis class $\CH$ of VC dimension $d,$ number of samples $m,$
    parameter $\delta \in (0,1),$
    there is an algorithm such that for any
    distribution $\CD$ over $\CX \times \{-1,1\}$ it returns a classifier $h_S: \CX \rightarrow \{-1, 1\}$ that, with probability at least $1-\delta$, has error bounded by
    \begin{gather*}
        \resizebox{\hsize}{!}{$
        \min \left\{  2.1 \cdot \tau + O\left(\sqrt{\frac{\tau(d + \ln(1/\delta))}{m}} + \frac{d + \ln(1/\delta)}{m} \right),  \tau + O\left(\sqrt{\frac{\tau(d + \ln(1/\delta))}{m}} + \frac{\ln^5(m/d)\cdot(d + \ln(1/\delta))}{m} \right) \right\} \,,
    $}
    \end{gather*}
    where $\tau$ is the error of the best hypothesis in $\CH.$
\end{theorem}

Notably, \cref{thm:main-result} settles the sample complexity of agnostic learning in the regime $\tau = O(d/m)$, by exhibiting an optimal learner which attains existing lower bounds. Furthermore, it improves upon the learner of \citet{hanneke2024revisiting} when $\tau = o(\ln^5 (m/d) \cdot d/m)$. In light of \cref{thm:main-result},  only the polylog
range $\tau \in [\omega(1), o(\ln^{10}(m/d))]\cdot \tfrac{d+\ln{\left(1/\delta \right)}}{m}$ remains to have its optimal error rates characterized.

Let us briefly describe our approach.
\citet{hanneke2024revisiting}, building upon \citet{devroye1996probabilistic},
states that any learner, upon receiving $m$ i.i.d.\ samples from $\CD$, must produce with probability
at least \linebreak
$\delta$ a hypothesis incurring error
$ \tau + \Omega(\sqrt{\tau(d + \ln(1/\delta))/m} + (d + \ln(1/\delta))/m )$
for worst-case $\CD$.
Our first elementary observation is that, due to this
lower bound, by designing a learner whose error is bounded by
$  c\cdot \tau + O(\sqrt{\tau(d + \ln(1/\delta))/m} + (d + \ln(1/\delta))/m )\,$
for some numerical constant $c \geq 1$, we can get a tight bound
in the regime $\tau =O(\nicefrac{d}{m}),$ thus
resolving \Cref{Question:1-tau-d/n-setting}.
Furthermore, to make progress across the full regime $\tau \in [0,\nicefrac{1}{2}],$ it is crucial to obtain a constant $c$ whose value is close to 1.
Motivated by this observation, our primary result gives an algorithm that
proceeds by taking
majority votes over ERMs trained on carefully crafted subsamples of the training set $S$, and which achieves error  $2.1 \cdot \tau + O(\sqrt{\tau(d + \ln(1/\delta))/m} + (d + \ln(1/\delta))/m)$.
Notably, this brings us within striking distance of the constant $c = 1$ --- we leave open the intriguing question of whether our technique can be refined to lower $c$ to 1, which would completely settle the complexity of agnostic learning.

{Our approach is inspired by \citet{hanneke2016optimal},
but introduces several new algorithmic components and ideas in the analysis. More concretely, we first modify Hanneke's sample splitting scheme,
and then randomly select a small fraction of the resulting subsamples on which to run $\cerm$, rather than running $\cerm$ on all such subsamples. This improves the $\cerm$-oracle efficiency of the algorithm.
In order to decrease the value of the constant multiplying $\tau$, our main insight is to run \emph{two} independent copies of the above classifier, as well as one $\cerm$
that is trained on elements coming from a certain ``region of disagreement'' of the previous two classifiers. For any test point $x,$ if both of the voting
classifiers agree on a label $y$ and have ``confidence'' in their vote, we output $y;$ otherwise we output the prediction of the
$\cerm.$ We hope that this idea of breaking ties between
voting classifiers using $\cerm$s that are trained on their
region of disagreement can find further applications.
}
{Our resulting algorithm employs a voting scheme at its center, making progress on \Cref{Question:2-majority-voting}, and is  computationally efficient with respect to ERM oracle calls, making considerable progress on \Cref{Question:3-computational-efficiency}.}
Finally, we combine this algorithm with the learner of \citet{hanneke2024revisiting} to obtain a ``best-of-both-worlds'' result. {An overview of all the steps is presented in the start of \cref{sec:small-constant-sketch}.}

\subsection{Related Work}

The PAC learning framework for statistical learning theory dates to the seminal work of \citet{valiant1984theory}, with roots in prior work of Vapnik and Chervonenkis \citep{vapnik1964class, vapnik1974theory}. In binary classification, finiteness of the VC dimension was first shown to characterize learnability by \citet{blumer1989learnability}. Tight lower bounds on the sample complexity of learning VC classes in the realizable case were established by \citet{ehrenfeucht1989general} and matched by upper bounds of \citet{hanneke2016optimal}, building upon work of \citet{simon2015almost}. Subsequent works have established different optimal PAC learners for the realizable setting \citep{aden2023optimal, baggingoptimal,aden2024majority,hogsgaard2025efficient}.
For agnostic learning in the standard PAC framework, ERM is known to achieve sample complexity matching existing lower bounds \citep{haussler1992decision, boucheron2005theory, anthony2009neural}. As described, we direct our attention to a more fine-grained view of agnostic learning, in which the error incurred by a learner above the best-in-class hypothesis $h^*_\CD$ is itself studied as a function of $\tau = \ls_\CD(h^*_\CD)$.
Bounds employing $\tau$ in the error term are sometimes referred to as \emph{first-order bounds} and have been previously analyzed in fields such as online learning \citep{Maurer2009EmpiricalBB,wagenmaker2022first}. \citet{hanneke2024revisiting} appear to be the first to consider $\tau$-optimal-dependence for upper bounds in PAC learning; we adopt their perspective in this work.

\vspace{-0.2 cm}
\section{Preliminaries}\label{Section:Preliminaries}
\vspace{-0.2 cm}

\paragraph{Notation}

For a natural number $m \in \N$, $[m]$ denotes the set $\{1, \ldots, m\}$. Random variables are written in bold face (e.g., $\mathbf{x}$) and their realizations in non-bold type face (e.g., $x$). For a set $Z$, $Z^*$ denotes the set of all finite sequences in $Z$, i.e., $Z^* = \bigcup_{i \in \N} Z^i$. For a sequence $S$ of length $m$ and indices $i \leq j \in [m]$, $S[i:j]$ denotes the smallest contiguous subsequence of $S$ which includes both its $i$th and $j$th entries. Furthermore, we employ 1-indexing for sequences. For $S = (a, b, c)$, for instance, $S[1:2] = (a, b)$. The symbol $\sqcup$ is used to denote concatenation of sequences, as in $(a, b) \sqcup (c, d) = (a, b, c, d)$. When $S, S' \in Z^*$ are sequences in $Z$ and each element $s \in S$ appears in $S'$ no less frequently than in $S$, we write $S \sqsubseteq S'$.
For a sequence $S$ and a set $A$
we denote by $S \sqcap A$ the longest subsequence of $S$
that consists solely of elements of $A$.
If $S$ is a finite set, then $\E_{x\sim S} \left[f(x)\right]$ denotes the expected value of $f$ over a uniformly random draw of $x \in S$.

\vspace{-0.2 cm}
\paragraph{Learning Theory}
Let us briefly recall the standard language of supervised learning. Unlabeled data points are drawn from a \defn{domain} $\CX$, which we permit to be arbitrary throughout the paper. We study binary classification, in which data points are labeled by one of two labels in the \defn{label set} $\CY = \{\pm 1\}$. A function $h : \CX \to \CY$ is referred to as a \defn{hypothesis} or \defn{classifier}, and a collection of such functions $\CH \subseteq \CY^\CX$ is a \defn{hypothesis class}. Throughout the paper, we employ the 0-1 loss function
$\ell_{0-1} : \CY \times \CY \to \R_{\geq 0}$ defined by $\ell_{0-1}(y, y') = \ind [y \neq y']$.
A \defn{training set} is a sequence of labeled data points $S = \big((x_1, y_1), \ldots, (x_m, y_m) \big) \in (\CX \times \CY)^*$. A \defn{learner} is a function which receives training sets and emits hypotheses, e.g., $\CA: (\CX \times \CY)^* \to \CY^\CX$. The purpose of a learner is to emit a hypothesis $h$ which attains low \defn{error}, or \emph{true error}, with respect to an unknown probability distribution $\CD$ over $\CX \times \CY$. That is,
$\ls_\CD(h) = \E_{(x, y) \sim \CD} \left[\ind [h(x) \neq y]\right]$.
A natural proxy for the true error of $h$ is its \defn{empirical error} on a training set $S = (x_i, y_i)_{i \in [m]}$, denoted $\ls_S(h) = \E_{(x, y) \sim S} \left[\ind[ h(x) \neq y]\right]$.
If $\CA$ is a learner for $\CH$ with the property that $\CA(S) \in \argmin_{h\in\CH} \ls_S(h)$ for all training sets $S$, then $\CA$ is said to be an \defn{empirical risk minimization} (ERM) learner for $\CH$. Throughout the paper we will use $\CA$ to denote an arbitrary ERM learner.

\section{Proof Sketch}\label{sec:proof-sketch}

We now provide a detailed explanation
of our approach and a comprehensive sketch of the
proof.
We divide our discussion into two parts.
In the first, we present a simple approach
that achieves an error bound of $15\tau + O(\sqrt{\nicefrac{\tau(d + \ln(1/\delta))}{m}} + \nicefrac{(d + \ln(1/\delta))}{m})$. Recall that this resolves the optimal sample complexity
for the regime $\tau \approx \nicefrac{d}{m}$.
In the second part, we describe several modifications to
the algorithm and new ideas in its analysis
which drive the error down to
$ 2.1\tau + O (\sqrt{\nicefrac{\tau(d + \ln(1/\delta))}{m}} + \nicefrac{(d + \ln(1/\delta))}{m} )$.

\vspace{-0.2 cm}
\subsection{First Approach: Multiplicative Constant 15}\label{sec:large-constant-sketch}
\vspace{-0.2 cm}

A crucial component of our algorithm
is a scheme $\CS'$ for recursively splitting the
input training sequence $S$ into subsequences,
which adapts Hanneke's recursive splitting algorithm \citep{hanneke2016optimal} and is formalized in
\Cref{alg:splittingwith3} and depicted in \Cref{fig:splittingwith3}. The algorithm takes two training sequences as input: $ S $, the active set, and $T$, the union of elements chosen in previous recursive calls.
\vspace{-0.2 cm}
\begin{algorithm}
\caption{Splitting algorithm $\cS'$}\label{alg:splittingwith3}
\KwIn{Training sequences $S, T \in (\cX \times \cY)^{*}$, where $|S| = 3^{k}$ for $k \in \mathbb{N}$.}
\KwOut{Family of training sequences.}
\If{$k \geq 6$}{
    Partition $S$ into $S_{1}, S_{2}, S_{3}$, with $S_i$ being the $(i-1)|S|/3+1$ to the $i|S|/3$ training examples of $S$. Set for each $i$
    \newline
     \textcolor{white}{halloooooooooooooo}$S_{i,\sqcup} = S_{i}[1:3^{k-4}],\quad\quad
        S_{i,\sqcap} = S_{i}[3^{k-4}+1:3^{k-1}]$,
    \newline
    \Return{$[\cS'(S_{1,\sqcup}; S_{1,\sqcap} \sqcup T), \cS'(S_{2,\sqcup}; S_{2,\sqcap} \sqcup T), \cS'(S_{3,\sqcup}; S_{3,\sqcap} \sqcup T)]$}
}
\Else{
    \Return{$S \sqcup T$}}
\end{algorithm}
\vspace{-0.2 cm}
\begin{figure}[h]
    \centering
        \usetikzlibrary{backgrounds}
        \resizebox{!}{4cm}{
        \begin{tikzpicture}[
            scale=0.8,
            every node/.style={transform shape},
            dataset/.style={rectangle, draw, minimum height=0.8cm, minimum width=5cm, align=center, font=\small},
            split/.style={rectangle, draw, minimum height=0.8cm, minimum width=2.5cm, align=center, font=\small},
            active/.style={fill=green!20},
            new_active/.style={fill=green!50},
            old_history/.style={fill=gray!30},
            new_history/.style={fill=gray!70}
        ]
        \node[dataset, active] (S) at (-3, 2.5) {Active set $S$};
        \node[dataset, old_history, minimum width=3cm] (T) at (1.3, 2.5) {History $T$};
        \node[] at (-5.9, 2.5) {$\cS'\bigg($};
        \node[font=\large] at (-0.37, 2.1) {,};
        \node[] at (3.025, 2.5) {$\bigg)$};

        \node[split, active, minimum width=1.6cm] (S1) at (-4.8, 0.5) {$S_1$};
        \node[split, active, minimum width=1.6cm] (S2) at (-3, 0.5) {$S_2$};
        \node[split, active, minimum width=1.6cm] (S3) at (-1.2, 0.5) {$S_3$};

        \draw[thick] (S.south west) ++(0.05, -0.1) -- ++(0, -0.2) -- ++(1.5, 0) -- ++(0, 0.2);
        \node[font=\small] at (-4.8, 1.5) {$1/3$};
        \draw[thick] (S.south) ++(-0.75, -0.1) -- ++(0, -0.2) -- ++(1.5, 0) -- ++(0, 0.2);
        \node[font=\small] at (-3, 1.5) {$1/3$};
        \draw[thick] (S.south east) ++(-1.55, -0.1) -- ++(0, -0.2) -- ++(1.5, 0) -- ++(0, 0.2);
        \node[font=\small] at (-1.2, 1.5) {$1/3$};

        \draw[->, thick] (-4.8, 1.3) -- (S1.north);
        \draw[->, thick] (-3, 1.3) -- (S2.north);
        \draw[->, thick] (-1.2, 1.3) -- (S3.north);

        \node[split, new_active, minimum width=0.5cm] (S1_sqcup) at (-6.75, -1.5) {$S_{1,\sqcup}$};
        \node[split, new_history, minimum width=1.5cm] (S1_sqcap) at (-5.4, -1.5) {$S_{1,\sqcap}\cup T$};
        \node[] at (-7.55, -1.5) {$\cS'\bigg($};
        \node[] at (-6.25, -1.95) {$,$};
        \node[] at (-4.45, -1.5) {$\bigg)$};

        \draw[thick] (S1.south west) ++(0.05, -0.1) -- ++(0, -0.2) -- ++(0.4, 0) -- ++(0, 0.2);
        \node[font=\tiny] at (-5.4, -0.4) {$1/27$};
        \draw[thick] (S1.south east) ++(-0.05, -0.1) -- ++(0, -0.2) -- ++(-1.0, 0) -- ++(0, 0.2);
        \node[font=\tiny] at (-4.55, -0.4) {$26/27$};

        \draw[->, thick] (-5.35, -0.5) -- (S1_sqcup.north);
        \draw[->, thick] (-4.55, -0.5) -- (S1_sqcap.north);

        \node[split, new_active, minimum width=0.5cm] (S2_sqcup) at (-3.1, -1.5) {$S_{2,\sqcup}$};
        \node[split, new_history, minimum width=1.5cm] (S2_sqcap) at (-1.75, -1.5) {$S_{2,\sqcap}\sqcup T$};
        \node[] at (-3.9, -1.5) {$\cS'\bigg($};
        \node[] at (-2.60, -1.95) {$,$};
        \node[] at (-0.80, -1.5) {$\bigg)$};

        \draw[thick] (S2.south west) ++(0.05, -0.1) -- ++(0, -0.2) -- ++(0.4, 0) -- ++(0, 0.2);
        \node[font=\tiny] at (-3.55, -0.4) {$1/27$};
        \draw[thick] (S2.south east) ++(-0.05, -0.1) -- ++(0, -0.2) -- ++(-1.0, 0) -- ++(0, 0.2);
        \node[font=\tiny] at (-2.75, -0.4) {$26/27$};

        \draw[->, thick] (-3.55, -0.5) -- (S2_sqcup.north);
        \draw[->, thick] (-2.75, -0.5) -- (S2_sqcap.north);

        \node[split, new_active, minimum width=0.5cm] (S3_sqcup) at (0.55, -1.5) {$S_{3,\sqcup}$};
        \node[split, new_history, minimum width=1.5cm] (S3_sqcap) at (1.90, -1.5) {$S_{3,\sqcap}\cup T$};
        \node[] at (-0.25, -1.5) {$\cS'\bigg($};
        \node[] at (1.05, -1.95) {$,$};
        \node[] at (2.85, -1.5) {$\bigg)$};

        \draw[thick] (S3.south west) ++(0.05, -0.1) -- ++(0, -0.2) -- ++(0.4, 0) -- ++(0, 0.2);
        \node[font=\tiny] at (-1.75, -0.4) {$1/27$};
        \draw[thick] (S3.south east) ++(-0.05, -0.1) -- ++(0, -0.2) -- ++(-1.0, 0) -- ++(0, 0.2);
        \node[font=\tiny] at (-0.9, -0.4) {$26/27$};

        \draw[->, thick] (-1.75, -0.5) -- (S3_sqcup.north);
        \draw[->, thick] (-0.9, -0.5) -- (S3_sqcap.north);

        \begin{scope}[on background layer]
            \draw[->, dashed, gray, opacity=0.7] (T.south) .. controls +(270:0.8) and +(90:0.8) .. (S3_sqcap.north);
                \draw[->, dashed, gray, opacity=0.7] (T.south) .. controls +(270:1.3) and +(90:1) .. (S2_sqcap.north);
                \draw[->, dashed, gray, opacity=0.7] (T.south) .. controls +(270:1.2) and +(90:1.2) .. (S1_sqcap.north);
        \end{scope}
        \begin{scope}[xshift=5.5cm, yshift=0.4cm]
            \node[draw, rectangle, active, minimum width=1cm, label=right:Active set] (legend_active) at (0,0) {};
            \node[draw, rectangle, new_active, minimum width=1cm, label=right:New active set] (legend_new_active) at (0,-0.7) {};
            \node[draw, rectangle, old_history, minimum width=1cm, label=right:History] (legend_history) at (0,-1.4) {};
            \node[draw, rectangle, new_history, minimum width=1cm, label=right:New history] (legend_new_history) at (0,-2.1) {};
        \end{scope}

        \end{tikzpicture}
        }
        \caption{The splitting process of algorithm $\mathcal{S}'$. The active set $S$ is split into three disjoint sets $S_1, S_2, S_3$. Each of these is then split into a new active set (green) and a set of previously recursed-on samples (grey), which are passed down to subsequent recursive calls.}
        \label{fig:splittingwith3}
    \end{figure}

In comparison to Hanneke's scheme, we have made two modifications which allow us to control the constant multiplying $\tau$, as we will explain shortly. First, we create disjoint splits
of the data, in contrast to the overlapping subsets employed by Hanneke. Second, we include more elements in the sets $S_{i,\sqcap}$, which are included in all subsequent training subsequences,
than in $S_{i,\sqcup}$, the  set on which we make a recursive call.

Let us introduce additional notation.
We denote $s'_{\sqcap}\coloneqq\nicefrac{|S|}{|S_{1,\sqcap}|}$,
for $S_{1,\sqcap}$ as defined in \Cref{alg:splittingwith3}.
For $\erm$ an $\cerm$ learner and $ \cS'(S;T) $ the splitting scheme of \Cref{alg:splittingwith3}, we set $\erm'(S;T) := ( \erm(S') )_{S'\in\cS'(S;T)}$, thought of as a multiset.
For an example $ (x,y)$, we define $ \avg(\erm'(S;T))(x,y)=\sum_{h\in \erm'(S;T)}\ind\{ h(x)\not=y \}/|\erm'(S;T)|$, i.e., the fraction of incorrect hypotheses in $\erm'(S;T)$ for $(x, y)$.
For a natural number $ t \in \N$, we let $ \smash{\hat{\erm}}'_{\rt}(S;T)$, be the random multiset of $ t $ hypotheses drawn independently and uniformly at random from $ \erm'(S;T)$.
In a slight overload of notation, we use
$ \rt $ (i.e., $ t $ in bold face) to denote the randomness used to draw the $ t $ hypotheses from $ \erm'(S;T).$
Intuitively, one can think of $ \smash{\hat{\erm}}'_{\rt} $ as a bagging algorithm where the subsampled training sequences are restricted to subsets of $ \cS'(S;T).$
Similarly, we define $\avg(\smash{\hat{\erm}}'_{\rt}(S;T))(x,y)=\sum_{h\in \smash{\hat{\erm}}'_{\rt}(S;T)}\ind\{ h(x)\not=y \}/|\smash{\hat{\erm}}'_{\rt}(S;T)|$.
For a distribution $ \cD $ over $ \cX\times \{ -1,1\},$ training sequences $S, T\in \left(\cX\times\{  -1,1\} \right)^{*},$ and $ \alpha\in[0,1]$,  we let $ \ls_{\cD}^{\alpha}(\erm'(S;T))=\p_{(\rx,\ry)\sim \cD}\left[\avg(\erm'(S;T))(\rx,\ry) \geq \alpha\right],$ i.e., the probability that at least an $ \alpha $-fraction of the hypotheses in $ \erm'(S;T) $ err on a new example drawn from $ \cD $. We will overload the notation $ \ls_{\cD} $ when considering majorities and write $ \ls_{\cD}(\erm'(S;T))=\ls_{\cD}^{0.5}(\erm'(S;T)) $--- the probability of an equal-weighted majority vote fails.
Similarly, we define $ \ls_{\cD}^{\alpha}(\smash{\hat{\erm}}'_{\rt}(S;T))=\p_{(\rx,\ry)\sim \cD}[\avg(\smash{\hat{\erm}}'_{\rt}(S;T))(\rx,\ry) \geq \alpha]$ and $ \ls_{\cD}(\smash{\hat{\erm}}'_{\rt}(S;T))=\ls_{\cD}^{0.5}(\smash{\hat{\erm}}'_{\rt}(S;T)) .$  Finally, we let $ \smash{\hat{\cA}}'_{\rt}(S)=\smash{\hat{\cA}}'_{\rt}(S;\emptyset)$, and $ \smash{\hat{\cA}}'_{\rt}(S)(x)=\sign(\sum_{h\in \smash{\hat{\erm}}'_{\rt}(S;T)}h(x))$.

We now describe our first approach,
which we break into three steps. The first step relates the error of $\ls_{\cD}(\smash{\hat{\cA}}'_{\rt}(\rS))$ to $\ls_{\cD}^{\smash 0.49}(\cA'(\rS,\emptyset))$, while the second and third steps bound the error of $ \ls_{\cD}^{0.49}(\cA'(\rS,\emptyset))$. The first step borrows ideas from \citet{baggingoptimal} and the last step from \citet{hanneke2016optimal}, but there are
several technical bottlenecks in the analysis that do
not appear in these works, as they
consider the realizable setting, for which $\tau = 0$.

\phantomsection
\addcontentsline{toc}{subsubsection}{
Relating the error of $ \smash{\hat{\cA}}'_{\rt}(\rS) $ to $ \cA'(\rS,\emptyset)$
}

\paragraph{Relating the error of $ \smash{\hat{\cA}}'_{\rt}(\rS) $ to $ \cA'(\rS,\emptyset) $:}
{We will} demonstrate how to bound the error of the random classifier $\smash{\hat{\cA}}'_{\rt}(\rS)$
using the error of $ \cA'(\rS,\emptyset)$.
First, let $\rS \sim \cD^m$ and
assume we have shown
\begin{align}\label{eq:proofsketch-1}
    \ls_{\cD}^{0.49}(\cA'(\rS;\emptyset))\leq 15\tau+O\left(\sqrt{\frac{\tau(d+\ln{\left(1/\delta \right)})}{m}}+\frac{d+\ln{\left(1/\delta  \right)}}{m}\right)\,
  \end{align}
with probability at least $ 1-\delta/2.$ Consider the event $ E=\{ (x,y):  \avg(\erm'(\rS;\emptyset))(\rx,\ry) \geq \nicefrac{49}{100}\}.$
By the law of total expectation, we bound the error of $ \smash{\hat{\erm}}'_{\rt}(\rS) $ as  $ \ls_{\cD}(\smash{\hat{\erm}}'_{\rt}(\rS))\leq \p_{(\rx,\ry)\sim\cD}\left(E\right)+\e_{(\rx,\ry)\sim \cD}[\ind\{\smash{\hat{\erm}}'_{\rt}(\rS)(\rx)\not=\ry  \}\mid \bar{E} ].$
We see that the first term on the right-hand side is $ \ls_{\cD}^{0.49}(\cA(\rS;\emptyset)) $, which we have assumed for the moment can be bounded by \Cref{eq:proofsketch-1}. We now argue that the second term can be bounded by $ O((d+\ln{\left(1/\delta \right)})/m).$
Note that under the event $ \bar{E}=\{ (x,y):  \avg(\erm(\rS;\emptyset))(x,y) < \nicefrac{49}{100} \}$, strictly more than half of the hypotheses in $ \erm(\rS;\emptyset)$ --- from which the hypotheses of $ \smash{\hat{\erm}}'_{\rt}(\rS) $ are drawn --- are correct.
Using this, combined with Hoeffding's inequality and switching the order of expectation (as $ \rt $ and $ (\rx,\ry) $ are independent), we get that $ \e_{\rt}[\e_{(\rx,\ry)\sim\cD}[\ind\{\smash{\hat{\erm}}'_{\rt}(\rS)(\rx)\not=\ry  \}\mid \bar{E} ]] \leq \exp\left(-\Theta(t)\right).$
Setting $ t=\Theta(\ln{\left(m/(\delta(\ln{\left(1/\delta \right)}+d)) \right)})$ then gives  $\e_{\rt}[\e_{(\rx,\ry)\sim\cD}[\ind\{\smash{\hat{\erm}}'_{\rt}(\rS)(\rx)\not=\ry  \}\mid \bar{E} ] ]=O\left((\delta(\ln{\left(1/\delta \right)}+d))/m\right).$
By an application of Markov's inequality, this implies with probability at least $ 1-\delta/2 $ over the draws of hypotheses in $ \smash{\hat{\erm}}'_{\rt}(\rS)$ that $ \e_{(\rx,\ry)\sim\cD}[\ind\{\smash{\hat{\erm}}'_{\rt}(\rS)(\rx)\not=\ry  \}\mid \bar{E} ]=O((d+\ln{\left(1/\delta \right)})/m).$
This bounds the second term in the error decomposition of $ \ls_{\cD}(\smash{\hat{\cA}}'_{\rt}(\rS;\emptyset))$ and  gives the claimed bound on $ \smash{\hat{\cA}}'_{\rt}(\rS).$

\phantomsection
\addcontentsline{toc}{subsubsection}{
Bounding the error of $ \ls_{\cD}^{0.49}(\cA'(\rS,\emptyset)) $
}

\paragraph{Bounding the error of $ \ls_{\cD}^{0.49}(\cA'(\rS,\emptyset)) $:}
We now give the proof sketch of \cref{eq:proofsketch-1}.
To this end, for a training sequence $ S' $ and  hypothesis $ h$ we define $\sum_{\not=}(h,S')  =\sum_{(x,y)\in S'}\ind\{h(x)\not=y\}.$
Assume for the moment that for $ h^{\star}\in \argmin\ls_{\cD}(h) $ we have demonstrated that with probability at least $ 1-\delta/4 $ over $ \rS, $ and for some numerical constants $c_b, c_c$,
\begin{align}\label{eq:proofsketch0}
    \resizebox{0.94\hsize}{!}{$ \ls_{\cD}^{0.49}(\erm'(\rS;\emptyset)) \leq  \max\limits_{S'\in\cS'(\rS,\emptyset)} \frac{14\Sigma_{\not=}(\hs\negmedspace,S')}{m/s_{\sqcap}'}+\sqrt{\frac{\cb\left(d+\ln{(4/\delta )}\right)\frac{14\Sigma_{\not=}(\hs\negmedspace,S')}{(m/s_{\sqcap}')}}{m}}+\frac{\cc\left(d+\ln{(4/\delta )}\right)}{m} \,.$}
\end{align}
In order to exploit \Cref{eq:proofsketch0}, we first observe that for each $ i\in\{  1,2,3\}  $, we have that
\begin{align}\label{eq:proofsketchconstant1}
   \max_{S'\in\cS'(\rS_{i,\sqcup};\rS_{i,\sqcap}\sqcup\emptyset)}\frac{14\Sigma_{\not=}(\hs\negmedspace,S')}{m/s_{\sqcap}'} \leq \frac{14|\rS_{i}|\Sigma_{\not=}(\hs\negmedspace,\rS_{i})}{|\rS_{i}|m/s_{\sqcap}'} \leq 15\ls_{\rS_{i}}(\hs),
\end{align}
as $ S'\sqsubseteq \rS_{i}$ and $ |\rS_{i}|s_{\sqcap}'/m=|\rS_{i}|/|\rS_{i,\sqcap}|=1/(1-\frac{1}{27})$.
We briefly remark that the multiplication with $ |\rS_{i}|/|\rS_{i,\sqcap}| $  is a source of a multiplicative factor on $ \tau$. However, it can be made arbitrarily close to 1 by making the split between $ \rS_{i,\sqcup} $ and $ \rS_{i,\sqcap} $ more imbalanced in the direction of $ \rS_{1,\sqcap},$ at the cost of larger constants $ \cb, \cc.$
Now, by an application of Bernstein's inequality on the hypothesis $ \hs$ (\cref{lem:additiveerrorhstar})
for each $ i\in\{ 1,2,3 \}  $, we have that $ 15\ls_{\rS_{i}}(\hs) $ is at most  $ 15(\tau+O(\sqrt{\tau\ln{(1/\delta )}/m}+\ln{(1/\delta )}/m)),$ with probability at least $ 1-\delta/4$ over $ \rS_{i}.$
Invoking the union bound and using that $\max_{S'\in \cS'(\rS,\empty)}=\max_{i\in\left\{  1,2,3\right\}}\max_{S'\in \cS'(\rS_{i,\sqcup},\rS_{i,\sqcap}\sqcup \emptyset)} $, we then have that  $ \max_{S'\in\cS'(\rS,\emptyset)}14\Sigma_{\not=}(\hs\negmedspace,S')/(m/s_{\sqcap}')=15(\tau+O(\sqrt{\tau\ln{\left(1/\delta \right)}/m}+\ln{\left(1/\delta \right)}/m))$ with probability at least $ 1- 3\delta/4$ over $ \rS.$
Finally, union bounding with the event in \cref{eq:proofsketch0} yields that both events hold with probability at least $ 1-\delta $ over $ \rS,$ from which the error bound of \cref{eq:proofsketch-1} follows by inserting the bound of $ \max_{S'\in\cS'(\rS,\emptyset)}14\Sigma_{\not=}(\hs\negmedspace,S')/(m/s_{\sqcap}') $ into \cref{eq:proofsketch0}.
\phantomsection
\addcontentsline{toc}{subsubsection}{
Relating the error of $ \ls_{\cD}^{0.49}(\cA'(\rS,\emptyset)) $ to the empirical error of $ \hs $
}

\paragraph{Relating the error of $ \ls_{\cD}^{0.49}(\cA'(\rS,\emptyset)) $ to the empirical error of $ \hs $:}
Here we draw inspiration from the seminal ideas of \cite{hanneke2016optimal} by analyzing the learner's loss recursively. However, certain aspects of our analysis  will diverge from Hanneke's approach, which is tailored to the realizable setting. As in \citet{hanneke2016optimal}, we have the splitting scheme $ \cS'(\, \cdot \,, \, \cdot \,) $ receive two arguments, the second of which can be thought of as the concatenation of all previous training sequences created by the recursive calls of \Cref{alg:splittingwith3}. As we are in the agnostic case, this second argument can be any training sequence $ T\in (\cX\times \{-1,1  \} )^{*}$. We will first demonstrate that
with probability at least $ 1-\delta $ over $ \rS $,
\begin{align}\label{eq:proofsketch3}
    \resizebox{0.94\hsize}{!}{$\ls_{\cD}^{0.49}(\erm'(\rS;T))
    \leq  \max\limits_{S'\in\cS(\rS;T)}\frac{14\Sigma_{\not=}(\hs\negmedspace,S')}{m/s_{\sqcap}'}+\sqrt{\frac{\cb\left(d+\ln{(1/\delta )}\right)\frac{14\Sigma_{\not=}(\hs\negmedspace,S')}{(m/s_{\sqcap}')}}{m}}+\frac{\cc\left(d+\ln{(1/\delta )}\right)}{m}$}.
\end{align}
Setting $ T=\emptyset $ and rescaling $\delta$ then yields \cref{eq:proofsketch0}.

The first step of our analysis is to relate the error of $ \ls_{\cD}^{0.49}(\cA'(\rS;T)) $ to that of the previous calls $ \cA'(\rS_{i,\sqcup};\rS_{i,\sqcap}\sqcup T). $
First note that for any $ (x,y) $ such that $ \avg(\cA'(\rS;T))(x,y)\geq \nicefrac{49}{100}$, at least one of the calls $ \cA'(\rS_{i,\sqcup};\rS_{i,\sqcap}\sqcup T) $ for $ i\in\{ 1,2,3 \}  $ must have $ \avg(\cA'(\rS_{i,\sqcup};\rS_{i,\sqcap}\sqcup T))(x,y)\geq \nicefrac{49}{100} $. Further, there must be at least $ (\frac{49}{100}-\frac{1}{3})\frac{3}{2} =\frac{47}{200}$ of the hypotheses in $ \cup_{j\in\{  1,2,3\}\backslash i } \cA'(\rS_{j,\sqcup};\rS_{j,\sqcap}\sqcup T)$ that also fail on $ (x,y).$
Using these observations, we have that if we draw a random index $ \rI\in\{  1,2,3\}  $ and a random hypothesis $ \rh \in \sqcup_{j\in\{  1,2,3\}\backslash \rI } \cA'(\rS_{j,\sqcup};\rS_{j,\sqcap}\sqcup T),$ then for $(x,y) $ such that  $   \avg(\cA'(\rS;T))(x,y)\geq \nicefrac{49}{100}  $  it holds that $ \p_{\rI,\rh}[\avg(\cA(\rS_{\rI,\sqcup};\rS_{\rI,\sqcap}\sqcup T))(x,y)\geq \nicefrac{49}{100},\rh(x)\not=y]\geq\frac{1}{3}\frac{47}{200}\geq \frac{1}{13}.$  Hence, we have that $13\e_{\rI,\rh}[\p_{(\rx,\ry)\sim\cD}[\avg(\cA(\rS_{\rI,\sqcup};\rS_{\rI,\sqcap}\sqcup T))(\rx,\ry)\geq \nicefrac{49}{100},\rh(\rx)\not=\ry]]\geq \ls_{\cD}^{0.49}(\cA'(\rS;T)).$
Furthermore, by the definition of $ \rI\in\{  1,2,3\}  $ and $ \rh$ being a random hypothesis from $ \sqcup_{j\in\{  1,2,3\}\backslash \rI } \cA'(\rS_{j,\sqcup};\rS_{j,\sqcap}\sqcup T),$ we conclude that
\begin{gather}
 \ls_{\cD}^{0.49}(\cA'(\rS;T))
\leq 13\e_{\rI,\rh}\left[\p_{(\rx,\ry)\sim\cD}\left[\avg(\cA'(\rS_{\rI,\sqcup};\rS_{\rI,\sqcap}\sqcup T))(\rx,\ry)\geq \nicefrac{49}{100},\rh(\rx)\not=\ry\right]\right]\nonumber  \\
\leq 13\negmedspace\negmedspace\negmedspace\negmedspace\negmedspace
    \max_{\substack{i \neq j\in\{1,2,3  \}, \\ h\in\cA'(\rS_{j,\sqcup};\rS_{j,\sqcap}\sqcup T)}} \negmedspace \negmedspace  \p_{(\rx,\ry)\sim\cD}\Big[\avg(\cA'(\rS_{i,\sqcup};\rS_{i,\sqcap}\sqcup T))(\rx,\ry)\geq \nicefrac{49}{100},h(\rx)\not=\ry\Big].\label{eq:proofsketch9}
\end{gather}

We will demonstrate \cref{eq:proofsketch3} by induction on $|\rS|=3^{k}$.
By setting $ \cb $ and $ \cc $ sufficiently large, we can assume the claim holds for $k < 9$.
By \cref{eq:proofsketch9} and a union bound, it suffices to bound each of the six terms for different combinations of $i, j$ by \cref{eq:proofsketch3} with probability at least $ 1-\delta/6$. Using symmetry, we can consider the case $ j=1$ and $ i=2.$
Note that $13\max_{h\in\cA'(\rS_{1,\sqcup};\rS_{1,\sqcap}\sqcup T)} \p_{(\rx,\ry)\sim\cD}\left[\avg(\cA'(\rS_{2,\sqcup};\rS_{2,\sqcap}\sqcup T))(\rx,\ry)\geq \nicefrac{49}{100},h(\rx)\not=\ry\right] $ equals
\vspace{0.2cm}
\begin{align}\label{eq:proofsketch1}
\hspace{-2.1cm}\max_{\substack{\phantom{texttttttttttttttttt}\raisebox{-0.255cm}{$\scriptstyle h\in\cA'(\rS_{1,\sqcup};\rS_{1,\sqcap}\sqcup T)$}}}\hspace{-2.1cm}\begin{Bmatrix}\hspace{-0.05cm}13 \hspace{-0.35cm} \p\limits_{(\rx,\ry)\sim\cD}\hspace{-0.40cm}\left[h(\rx)\not=\ry\hspace{-0.05cm}\mid \hspace{-0.05cm}\avg(\cA'(\rS_{2,\sqcup};\rS_{2,\sqcap}\sqcup T))(\rx,\ry)\geq 0.49\right]\ls_{\cD}^{0.49}(\cA'(\rS_{2,\sqcup};\rS_{2,\sqcap}\sqcup T))\hspace{-0.05cm}\end{Bmatrix}\hspace{-0.05cm}.
\end{align}
Supposing that $\ls_{\cD}^{0.49}(\cA'(\rS_{2,\sqcup};\rS_{2,\sqcap}\sqcup T))$ is upper bound by
\begin{align}\label{eq:proofsketch5}
 \frac{1}{13}\Bigg(\max\limits_{S'\in\cS(\rS;T)}\frac{14\Sigma_{\not=}(\hs\negmedspace,S')}{m/s_{\sqcap}'}+\sqrt{\frac{\cb\left(d+\ln{(1/\delta )}\right)\frac{14\Sigma_{\not=}(\hs\negmedspace,S')}{(m/s_{\sqcap}')}}{m}}+\frac{\cc\left(d+\ln{(1/\delta )}\right)}{m}\Bigg),
\end{align}
then \cref{eq:proofsketch1} is bounded as in \cref{eq:proofsketch3} and we are done.
Using that $ |\rS_{2,\sqcup}|=m/3^{4}$, the inductive hypothesis gives that with probability at least $ 1-\delta/16 $ over $ \rS_{2}$ that $ \ls_{\cD}^{0.49}(\cA'(\rS_{2,\sqcup};\rS_{2,\sqcap}\sqcup T)) $ is at most
\begin{align}\label{eq:proofsketch6}
 \max\limits_{S'\in\cS(\rS;T)}\frac{14\Sigma_{\not=}(\hs\negmedspace,S')}{m/(3^{4}s_{\sqcap}')}+\sqrt{\frac{\cb\left(d+\ln{(16/\delta )}\right)\frac{14\Sigma_{\not=}(\hs\negmedspace,S')}{(m/s_{\sqcap}')}}{m/3^{4}}}+\frac{\cc\left(d+\ln{(16/\delta )}\right)}{m/3^{4}}.
\end{align}
Where we have used that $\cS(\rS_{2,\sqcup};\rS_{2,\sqcap}\sqcup T) \subseteq \cS(\rS;T)$.
Thus, it suffices to consider the case in which $ \ls_{\cD}^{0.49}(\cA'(\rS_{2,\sqcup};\rS_{2,\sqcap}\sqcup T)) $ lies between the expressions in \cref{eq:proofsketch5} and \cref{eq:proofsketch6}.

Let $ A=\{  (x,y)| \avg(\cA'(\rS_{2,\sqcup};\rS_{2,\sqcap}\sqcup T))(x,y)\geq \nicefrac{49}{100}\}  $ and  $ \rN=|\rS_{1,\sqcap}\sqcap A|$.
Using a Chernoff bound, one has that $ \rN\geq  13m/(14s_{\sqcap}')\ls_{\cD}^{0.49}(\cA'(\rS_{2,\sqcup};\rS_{2,\sqcap}\sqcup T))$ with probability at least $1 - \delta / 16$.
As $ \rS_{1,\sqcap}\sqcap A \sim \cD(\cdot\mid \avg(\cA'(\rS_{2,\sqcup};\rS_{2,\sqcap}\sqcup T))(\rx,\ry)\geq \nicefrac{49}{100}) $, uniform convergence over $ \cH $ (see \cref{lem:fundamentalheoremoflearning}) yields that with probability $ 1-\delta/16 $ over $ \rS_{1,\sqcap}\sqcap A $, \cref{eq:proofsketch1} is bounded by
\begin{align}\label{eq:proofsketch2}
13\negmedspace\negmedspace\negmedspace\negmedspace\negmedspace\negmedspace\max_{h\in\cA'(\rS_{1,\sqcup};\rS_{1,\sqcap}\sqcup T)} \left( \ls_{\rS_{1,\sqcap}\sqcap A}(h)+\sqrt{\frac{C(d+2\ln{\left(48/\delta \right)})}{\rN}}\right)\ls_{\cD}^{0.49}(\cA'(\rS_{2,\sqcup};\rS_{2,\sqcap}\sqcup T)) \,.
\end{align}
Where $ C >0$  is a universal constant.
We now bound each term in \cref{eq:proofsketch2}, considered after multiplying by $\ls_{\cD}^{0.49}$. Notably, this component of the analysis diverges considerably from that of \citet{hanneke2016optimal} for the realizable case, in which one is assured that $ \ls_{\rS_{1,\sqcap}\sqcap A}(h) = 0$.
For the first, we use the definitions of $\rN$ and $\ls_{\rS_{1,\sqcap}\sqcap A}$ (and abbreviate $h\in\cA'(\rS_{1,\sqcup};\rS_{1,\sqcap}\sqcup T)$ as $h \in \cA'$) to observe that
\begin{align}\label{eq:proofsketch7}
    13 \, \max_{h\in\cA'}  \ls_{\rS_{1,\sqcap}\sqcap A}(h)\ls_{\cD}^{0.49}(\cA'(\rS_{2,\sqcup};\rS_{2,\sqcap}\sqcup T)) \leq 14 \, \max_{h\in\cA'} \sum_{(x,y)\in\rS_{1,\sqcap}\sqcap A}\frac{\ind\{h(x)\not=y  \} }{m/s_{\sqcap}'}.
\end{align}
Further, using that for any $S'\in \cS'(\rS_{1,\sqcup};\rS_{1,\sqcap}\sqcup T)$ we have $ \rS_{1,\sqcap}\sqcap A \subseteq S'$, and $ h=\cA(S') $ is a minimizer of $ \sum_{(x,y)\in S'}\ind\{h(x)\not=y  \}$, combined with that $ \cS(\rS_{1,\sqcup};\rS_{1,\sqcap}\sqcup T) \sqsubseteq \cS(\rS;T)$, one can invoke \cref{eq:proofsketch7} to conclude that
\begin{align}
    13\max_{h\in\cA'(\rS_{1,\sqcup};\rS_{1,\sqcap}\sqcup T)}  \ls_{\rS_{1,\sqcap}\sqcap A}(h)\ls_{\cD}^{0.49}(\cA'(\rS_{2,\sqcup};\rS_{2,\sqcap}\sqcup T)) \leq 14 \max\limits_{S'\in\cS(\rS;T)}\frac{\Sigma_{\not=}(\hs\negmedspace,S')}{m/s_{\sqcap}'}.\label{eq:proofsketch10}
\end{align}
We note that the above step introduces a factor of $\frac{14}{13}$ on $\tau$, which can be brought arbitrarily close to 1 by using a tighter Chernoff bound for the size of $ \rN $, at the cost of larger constants $ \cc $ and $ \cb $ in \cref{eq:proofsketch3}.
For the second term $ \sqrt{C(d+2\ln{\left(48/\delta \right)})/\rN}\ls_{\cD}^{0.49}(\cA'(\rS_{2,\sqcup};\rS_{2,\sqcap}\sqcup T)) $ of \cref{eq:proofsketch2}, we again use that $ \rN\geq  (13m)/(14s_{\sqcap}')\ls_{\cD}^{0.49}(\cA'(\rS_{2,\sqcup};\rS_{2,\sqcap}\sqcup T))$ to bound it from above by
$\sqrt{(14C(d+2\ln{\left(48/\delta \right)})\ls_{\cD}^{0.49}(\cA'(\rS_{2,\sqcup};\rS_{2,\sqcap}\sqcup T)))/(13m/s_{\sqcap}')}$.
As we are considering the case in which $\ls_{\cD}^{0.49}$ is bounded by the expression of \cref{eq:proofsketch6}, one can show that the following upper bound holds for sufficiently large $ \cb $ and $ \cc $:
\[ \sqrt{\frac{\cb\left(d+\ln{(1/\delta )}\right)\frac{14\Sigma_{\not=}(\hs\negmedspace,S')}{(m/s_{\sqcap}')}}{m}}+\frac{\cc\left(d+\ln{(1/\delta )}\right)}{m}. \]
Combining these bounds on each term of \cref{eq:proofsketch2} yields an upper bound of the form
\begin{align}\label{eq:proofsketch4}
    \max\limits_{S'\in\cS'(\rS;T)}\frac{14\Sigma_{\not=}(\hs\negmedspace,S')}{m/s_{\sqcap}'}+\sqrt{\frac{\cb\left(d+\ln{(1/\delta )}\right)\frac{14\Sigma_{\not=}(\hs\negmedspace,S')}{(m/s_{\sqcap}')}}{m}}+\frac{\cc\left(d+\ln{(1/\delta )}\right)}{m} \,,
\end{align}
completing the inductive step
and establishing \cref{eq:proofsketch3}, as desired.

\subsection{Improved Approach: Multiplicative Constant 2.1}\label{sec:small-constant-sketch}

Roughly speaking, we apply two modifications to the previous approach in order to achieve a considerably smaller constant factor on $\tau$. First, we use a different splitting scheme (\cref{alg:splitting27if}) which recursively splits the dataset into 27 subsamples, rather than 3. Second,
rather than taking a majority vote over $\cerm$ learners on a single instance of the splitting algorithm,
we split the training set into three parts, run two independent instances
of the voting classifier on the first two parts, and one instance of $\cerm$
on a subsample of the third part. The subsample we train the $\cerm$ on
is carefully chosen and depends on a certain ``region of disagreement''
of the two voting classifiers.
The final prediction at a given datapoint is as follows: If both voting classifiers
agree on the predicted label with a certain notion of ``margin,'' then we output
that label, otherwise we output the prediction of the tiebreaker $\cerm.$
Lastly, we combine our approach
with that of \citet{hanneke2024revisiting} to achieve a best-of-both-worlds bound.
We summarize our approach in the following point form and \Cref{fig:main_algorithm}.
\begin{enumerate}
    \item Split the training set $\rS$ into three equally-sized parts $\rS_1, \rS_2, \rS_3.$

    \item Split $\rS_1$ into three equally-sized parts,
    $\rS_{1,1}, \rS_{1,2}, \rS_{1,3}.$
    \begin{enumerate}
    \item Run the splitting scheme of \Cref{alg:splittingwith27} on $(S_{1,1}, \emptyset)$ and $(S_{1,2}, \emptyset)$.
    Let $\CS_1, \CS_2$ be the resulting sets of training subsequences. 

    \item Sample $t= O \big(\ln{\left(m/(\delta(d+\ln{\left(1/\delta \right)})) \right)} \big) $  sequences from each of $\CS_1$ and $\CS_2$
    uniformly at random. Let $\hat \CS_1, \hat \CS_2$
    be the resulting collections of sequences.

    \item Train $\cerm$ learners on each sequence
    appearing in $\hat \CS_1, \hat \CS_2$. Denote by $\hat \CA_{\rt_1}, \hat \CA_{\rt_2}$
    the resulting collections of classifiers produced by these $\cerm$s, respectively.

    \item Define the set $\rS_3^{\not=}$ as follows:
    for any $(x,y) \in \rS_{1,3}$
    if at least an $11/243$-fraction of
    the classifiers in $\hat \CA_{\rt_1}$ do not predict the label $y$ for $x$, or likewise for $\hat \CA_{\rt_2}$, then $(x,y) \in  \rS_3^{\not=}$.
    Train an $\cerm$ on $\rS_3^{\not=}$,
    producing the classifier $\htie.$

    \item Let $\tilde \CA_1$ be the classifier which acts as follows on any given $x$: if a $232/243$-fraction of the classifiers of both $\hat \CA_{\rt_1}$ and $\hat \CA_{\rt_2}$ agree on a label $y$ for $x$, then $\tilde \CA_1$ predicts $y.$ Otherwise, it predicts $\htie(x)$.
    \end{enumerate}

    \item Use $\rS_2$ to train the algorithm of \citet{hanneke2024revisiting}. Let $\tilde \CA_2$ be the resulting classifier.

    \item Output the classifier among $\tilde \CA_1$ and $\tilde \CA_2$ which attains superior performance on $S_3$.

 \end{enumerate}
\begin{figure}[h!]
\centering
\resizebox{\textwidth}{!}{
\begin{tikzpicture}[
    auto,
    block/.style={rectangle, draw=black, thick, rounded corners, fill=blue!20, text centered, inner sep=3pt, align=center},
    subblock/.style={rectangle, draw=black, thick, rounded corners, fill=green!20, text centered, inner sep=3pt},
    aggregator/.style={ellipse, draw=black, thick, fill=red!20, text centered, inner sep=3pt},
    test/.style={trapezium, trapezium left angle=70, trapezium right angle=110, draw=black, thick, fill=orange!20, text centered, inner sep=3pt, align=center},
    line/.style={draw, thick, -latex},
    node distance=2cm and 2cm
]

\node[block] (S) {$\rS$};

\node[subblock, right=of S, xshift=-1cm] (S1) {$\rS_1$};
\path[line] (S.east) -- node[above] {Split} (S1.west);

\node[subblock, right=of S1, xshift=-1cm] (S11) {$\rS_{1,1}$};
\node[subblock, below=0.5cm of S11] (S12) {$\rS_{1,2}$};
\node[subblock, below=0.5cm of S12] (S13) {$\rS_{1,3}$};
\path[line] (S1.east) -- ++(0.5,0) node (S1fork) {} |- (S11.west);
\path[line] (S1fork) |- (S12.west);
\path[line] (S1fork) |- (S13.west);
\node at (S1fork |- S11.west) [above] {Split};

\node[subblock, right=of S11] (CS1) {$\CS_1$};
\node[subblock, right=of S12] (CS2) {$\CS_2$};
\path [line] (S11.east) -- node[above] {\Cref{alg:splitting27if}} (CS1.west);
\path [line] (S12.east) -- node[above] {\Cref{alg:splitting27if}} (CS2.west);

\node[subblock, right=of CS1] (hatCS1) {$\hat{\CS}_1$};
\node[subblock, right=of CS2] (hatCS2) {$\hat{\CS}_2$};
\path [line] (CS1.east) -- node[above] {Sample $t_1$} (hatCS1.west);
\path [line] (CS2.east) -- node[above] {Sample $t_2$} (hatCS2.west);

\node[aggregator, right=of hatCS1,xshift=0.2cm] (hatA1) {$\hat{\CA}_{\rt_1}$};
\node[aggregator, right=of hatCS2,xshift=0.2cm] (hatA2) {$\hat{\CA}_{\rt_2}$};
\path [line] (hatCS1.east) -- node[above] {Train Majority} (hatA1.west);
\path [line] (hatCS2.east) -- node[above] {Train Majority} (hatA2.west);

\node[subblock, right=of S13, xshift=5.62cm] (S3neq) {$\rS_{3}^{\not=}$};
\node[aggregator, right=of S3neq,xshift=-1cm] (htie) {$h_{\mathrm{tie}}$};
\path [line] (S13.east) |- (S3neq.west);
\path [line] (S3neq.east) -- node[above] {Train} (htie.west);
\path [line] (hatA2.south) -- ++(0,-0.25) -| node[midway, left,xshift=-0.2cm,yshift=0.17cm] {Filter} (S3neq.north);
\draw[thick] (hatA1.south) |- (hatA2.north);

\node[aggregator, right=4cm of hatA2,xshift=-1.5cm] (tildeA2) {$\tilde{\CA}_{1}$};
\path [line] (hatA1.east) -- node[above,xshift=0.3cm,yshift=0cm] {Aggregate} (tildeA2.west);
\path [line] (hatA2.east) |- (tildeA2.west);
\path [line] (htie.east) -- (tildeA2.west);

\node[subblock, below=2.5cm of S1] (S2) {$\rS_2$};
\node[aggregator, right=of S2, xshift=4.5cm] (tildeA1) {$\tilde{\CA}_{2}$};
\path[line] (S.east) -- ++(0.5,0) |- (S2.west);
\path[line] (S2.east) -- node[above] {Train \cite{hanneke2024revisiting}} (tildeA1.west);

\node[subblock, below=0.2cm of S2] (S3) {$\rS_3$};
\path[line] (S.east) -- ++(0.5,0) |- (S3.west);

\node[test, right=of tildeA1, xshift=3cm] (testS3) {Test on $\rS_3$};
\node[block, right=of testS3, xshift=0cm, text width=5em] (final) {Final Classifier};

\path [line] (tildeA1.east) -- (testS3.west);
\path [line] (tildeA2.south) -| (testS3.north);
\path [line] (S3.east) -| (testS3.south);
\path [line] (testS3.east) -- node[above] {Best on Test} (final.west);

\end{tikzpicture}
}
\caption{A flowchart of the final algorithm. The initial sample $\rS$ is split into three parts. $\rS_1$ is used to construct our tie-breaking classifier $\tilde{\CA}_1$. $\rS_2$ is used to train the algorithm of \citet{hanneke2024revisiting}, yielding $\tilde{\CA}_2$. Finally, $\rS_3$ is used as a hold-out set to select the better of the two classifiers.}
\label{fig:main_algorithm}
\end{figure}

\begin{remark}
Intuitively, $\htie$ can be seen as ``stabilizing'' the predictions produced by the classifiers in the collections $\hat \CA_{\rt_1}$ and $\hat \CA_{\rt_2}$ when at least one collection is judged to have low confidence. However, strictly speaking, we cannot demonstrate that the learner $\tilde \CA_1$ described by Step (2.) is stable in the sense of \citet{bousquet2002stability}. In particular, the classifiers in each of $\hat \CA_{\rt_1}$ and $\hat \CA_{\rt_2}$ are trained on overlapping samples, meaning that many classifiers could change by altering even one example in $\rS_{1,1}$ or $\rS_{1,2}$, thereby altering $\rS_3^{\not=}$
as well. 
\end{remark}

We now derive the generalization error for the steps under points $2.1-2.5$. The derivation is presented in a bottom-up fashion, starting by examining the failure modes of the approach in \cref{sec:large-constant-sketch} (which leads to the bound with $15\tau$), and amending these to obtain the approach of $2.1-2.5$ bounded by $2.1\tau$.
To this end, let us recall the three sources leading to multiplicative factors on $\tau$ in \cref{sec:large-constant-sketch}. First is the balance between the splits of $ S_{i} $ in \cref{alg:splittingwith3}, i.e. $ |S_{i}|/|S_{i,\sqcap}|=1/(1-\nicefrac{1}{27})$ in \cref{eq:proofsketchconstant1}.
Recall that this constant can be driven down to 1 by considering even more imbalanced splits of the dataset, at the expense
of larger constants multiplying the remaining terms of the bound.
Second is the error arising from \cref{eq:proofsketch10}, which controls the size of $ \rN $ via a Chernoff bound. This can similarly be driven arbitrarily close to 1. Hence, the primary multiplicative overhead results from the third source: the argument in \cref{eq:proofsketch9} relating the error of $ \cA'(\rS,T) $ to that of its recursive calls.

The constant arises from relating $ \ls_{\cD}^{0.49}(\cA'(\rS;T)) $ to one of its previous iterates erring on a $\nicefrac{49}{100}$-fraction of its classifiers, and one hypothesis from the remaining iterates also erring.
Recall that to get this bound we first observe that, with probability at least $ \nicefrac{1}{3}$ over a randomly drawn index $ \rI\sim\left\{  1,2,3\right\} ,$ $ \avg(\cA'(\rS_{\rI,\sqcup};\rS_{\rI,\sqcap}\sqcap T))(\rx,\ry) \geq 0.49$ and a random hypothesis $\rh\in \sqcup_{j\in\{  1,2,3\}\backslash \rI } \cA'(\rS_{j,\sqcup};\rS_{j,\sqcap}\sqcup T),$ would, with probability at least  $ \nicefrac{3}{2}(\nicefrac{49}{100}-\nicefrac{1}{3}),$ make an error. Thus, we get the inequality in \cref{eq:proofsketch9}, which multiplies $ \tau $ by $ \approx 13.$
To decrease this constant we first make the following observation: assume that we have two voting classifiers $ \cA'(\rS_{1};\emptyset),  \cA'(\rS_{2};\emptyset) $ and an $ (x,y) $ such that $ \avg( \cA'(\rS_{1};\emptyset))(x,y)\geq \nicefrac{232}{243} $ and $ \avg( \cA'(\rS_{2};\emptyset))(x,y)\geq \nicefrac{232}{243}$.

Using a similar analysis as
in \cref{eq:proofsketch9}, we have
\begin{align}
 \p_{(\rx,\ry)\sim \cD}\big[\avg(  \cA'(\rS_{1};\emptyset))(\rx,\ry)\geq \nicefrac{232}{243}, \avg( \cA'(\rS_{2};\emptyset))(\rx,\ry)\geq \nicefrac{232}{243}\big]\label{eq:proofsketch12}
 \\
 \leq \frac{243}{232}\max_{h\in \cA'(\rS_{1};\emptyset)}\p_{(\rx,\ry)\sim \cD}\left[h(\rx)\not=\ry, \avg( \cA'(\rS_{2};\emptyset))(\rx,\ry)\geq \nicefrac{232}{243}\right].\label{eq:proofsketch13}
\end{align}
Following a similar approach to that used to obtain \cref{eq:proofsketch-1}, we can bound \cref{eq:proofsketch13} by
$1.1\tau+ O(\sqrt{\tau(d+\ln{\left(1/\delta \right)})/m}+(d+\ln{\left(1/\delta \right)})/m)$.
Thus, if we could bound the error using the event in \cref{eq:proofsketch12}, we could obtain a better bound for the overall error of our algorithm.

To this end, consider a third independent training sequence $ \rS_{3}$ and let $ \rS_{3}^{\not=} $ be the training examples $ (x,y) $ in $ \rS_{3} $ such that $ \avg( \cA'(\rS_{1};\emptyset))(x,y)\geq \nicefrac{11}{243} $ or $ \avg( \cA'(\rS_{2};\emptyset))(x,y)\geq \nicefrac{11}{243}$. Let $ \htie=\erm(\rS_{3}^{\not=})$, i.e. $\htie$ is the output of ERM-learner trained on $\rS_3^{\not=}$. We now introduce our tie-breaking idea. For a point $ x $ we let $ \tie^{\nicefrac{11}{243}}\left(\cA'(\rS_{1};\emptyset),\cA'(\rS_{2};\emptyset); \htie\right)(x) = y$, where $y \in \{\pm 1\}$ is the unique number
for which  $ \sum_{h\in \cA'(\rS_{1};\emptyset)}\ind\{h(x)=y\}/|\cA'(\rS_{1};\emptyset)|\geq \nicefrac{232}{243}$ and $\sum_{h\in \cA'(\rS_{2};\emptyset)}\ind\{h(x)=y\}/|\cA'(\rS_{2};\emptyset)|\geq \nicefrac{232}{243}$. If such a number does not exist, we set $y = \htie(x).$ Thus, this classifier errs on $ (x,y) $ if 1) $ \avg( \cA'(\rS_{1};\emptyset))(x,y)\geq \nicefrac{232}{243} $ and $ \avg( \cA'(\rS_{2};\emptyset))(x,y)\geq \nicefrac{232}{243}$,  or 2) for all $y'\in \{ \pm 1 \}$,  $\sum_{h\in \cA'(\rS_{1};\emptyset)}\ind\{h(x)=y'\}/|\cA'(\rS_{1};\emptyset)|< \nicefrac{232}{243}$ or $\sum_{h\in \cA'(\rS_{2};\emptyset)}\ind\{h(x)=y'\}/|\cA'(\rS_{2};\emptyset)|< \nicefrac{232}{243} $ and $ \htie(x)\not=y.$
Using the definition of condition 2), we have that
\begin{gather*}
\ls_{\cD}  \big(\tie^{\nicefrac{11}{243}}\left(\cA'(\rS_{1};\emptyset),\cA'(\rS_{1};\emptyset); \htie\right) \big)
\\
 \leq
 \p_{(\rx,\ry)\sim \cD}\negmedspace\left[\avg( \cA'(\rS_{1};\emptyset))(\rx,\ry)\geq \nicefrac{232}{243}, \avg( \cA'(\rS_{2};\emptyset))(\rx,\ry)\geq \nicefrac{232}{243}\right]
 \\
+  \p_{(\rx,\ry)\sim \cD}\negmedspace\left[ \avg( \cA'(\rS_{1};\emptyset))(\rx,\ry)\geq \nicefrac{11}{243}\text{ or } \avg( \cA'(\rS_{2};\emptyset))(\rx,\ry)\geq \nicefrac{11}{243} \text{ and } \htie(\rx)\not=\ry\right]. 
\end{gather*}
As observed below \cref{eq:proofsketch12}, the first term  can be bounded by $ 1.1\tau+ O(\sqrt{\tau(d+\ln{\left(1/\delta \right)})/m}+(d+\ln{\left(1/\delta \right)})/m)$. Thus, if we could bound the second term by  $\tau+ O(\sqrt{\tau(d+\ln{\left(1/\delta \right)})/m}+(d+\ln{\left(1/\delta \right)})/m)$, we would be done.
Proceeding as in \cref{eq:proofsketch1}, let $ \cD_{\not=} $ be the conditional distribution given $ \avg( \cA'(\rS_{1};\emptyset))(\rx,\ry)\geq \nicefrac{11}{243}$ or $ \avg( \cA'(\rS_{2};\emptyset))(\rx,\ry)\geq \nicefrac{11}{243}.$
Then,
\begin{align*}
\p_{(\rx,\ry)\sim \cD}\negmedspace\left[ \avg( \cA'(\rS_{1};\emptyset))(\rx,\ry)\geq \nicefrac{11}{243}\text{ or } \avg( \cA'(\rS_{2};\emptyset))(\rx,\ry)\geq \nicefrac{11}{243} \text{ and } \htie(\rx)\not=\ry\right]
   =  \\
\p_{(\rx,\ry)\sim \cD_{\not=}}\negmedspace\negmedspace\negmedspace\left[ \htie (\rx)\not=\ry\right]
    \negmedspace\p_{(\rx,\ry)\sim \cD}\negmedspace\left[ \avg( \cA'(\rS_{1};\emptyset))(\rx,\ry)\geq \nicefrac{11}{243}\text{ or } \avg( \cA'(\rS_{2};\emptyset))(\rx,\ry)\geq \nicefrac{11}{243}\right].
\end{align*}
We notice that $\htie$ is the output of an $ \cerm $ learner trained on $ \rS_{3}^{\not=} \sim \cD_{\not=},$ thus, by standard guarantees on ERM learners (\Cref{thm:ermtheoremunderstanding}), it holds with high probability that
\begin{align}
    \p_{(\rx,\ry)\sim \cD_{\not=}}\left[ \htie(\rx)\not=\ry\right]\leq \inf_{h\in \cH}\ls_{\cD_{\not=}}(h)+\sqrt{\frac{d+\ln{\left(1/\delta \right)}}{|\rS_{3}^{\not=}|}}\leq \ls_{\cD_{\not=}}(\hs)+\sqrt{\frac{d+\ln{\left(1/\delta \right)}}{|\rS_{3}^{\not=}|}} \,,\nonumber
\end{align}
where the second inequality follows from that $\hs\in\cH$. Thus, we have that
\begin{align*}
&  \footnotesize \p_{(\rx,\ry)\sim \cD}\left[ \avg( \cA'(\rS_{1};\emptyset))(\rx,\ry)\geq \nicefrac{11}{243}\text{ or } \avg( \cA'(\rS_{2};\emptyset))(\rx,\ry)\geq \nicefrac{11}{243}\text{ and } \htie(\rx)\not=\ry\right] \leq \\
& \footnotesize  \tau+\sqrt{\frac{\left(d+\ln{\left(1/\delta \right)}\right)\p_{\rx,\ry\sim \cD}\left[\avg( \cA'(\rS_{1};\emptyset))(\rx,\ry)\geq \nicefrac{11}{243}\text{ or } \avg( \cA'(\rS_{2};\emptyset))(\rx,\ry)\geq \nicefrac{11}{243}\right]^{2}}{|\rS_{3}^{\not=}|}}
\end{align*}
Using a similar argument as in \cref{eq:proofsketch2}, a Chernoff bound yields that $ |\rS_{3}^{\not=}| =\Theta(m\p_{\rx,\ry\sim \cD}\left[\avg( \cA'(\rS_{1};\emptyset))(\rx,\ry)\geq \nicefrac{11}{243}\text{ or } \avg( \cA'(\rS_{2};\emptyset))(\rx,\ry)\geq \nicefrac{11}{243}\right])$ with high probability. Thus, if the probability term were at most $ c(\tau +(d+\ln{\left(1/\delta \right)})/m)$, the above argument would be complete. However, the analysis from \cref{sec:large-constant-sketch} only bounds the probability term when it is $\nicefrac{49}{100}$, instead of $\nicefrac{11}{243},$ which is not sufficient. Thus, we will now remedy this using a different splitting algorithm that allows for showing the more fine-grained error is small.

\phantomsection
\addcontentsline{toc}{subsubsection}{
An approach with more than $ 3 $ splits
}

\paragraph{An approach with more than $ 3 $ splits:}
We have seen that the above argument requires that $ \p\left[\avg( \cA'(\rS;\emptyset))(\rx,\ry)\geq \nicefrac{11}{243}\right]\leq c(\tau +(d+\ln{\left(1/\delta \right)})/m)$ with high probability.
However, attempting to prove this by induction, as in \cref{sec:large-constant-sketch}, breaks down in the step at which we derived \cref{eq:proofsketch9}. Recall that this is where we relate the condition $\avg( \cA'(\rS;T))(x,y)\geq \nicefrac{11}{243} $ to a previous recursive call also erring on an $ \nicefrac{11}{243} $-fraction of its voters and one of the hypotheses in the remaining recursive calls erring on $ (x,y) $.
To see why this argument fails, now consider $ (x,y) $ such that $\avg( \cA'(\rS;T))(x,y)\geq \nicefrac{11}{243}.$ Picking an index $ \rI\sim \left\{  1,2,3\right\}  $ with probability at least $ \nicefrac{1}{3} $ still returns  $\avg( \cA'(\rS_{\rI,\sqcup};T\sqcup \rS_{\rI,\sqcap}))(x,y)\geq \nicefrac{11}{243}.$ However, when picking a random hypothesis $\rh\in \sqcup_{j\in\{  1,2,3\}\backslash \rI } \cA'(\rS_{1,j,\sqcup};\rS_{j,\sqcap}\sqcup T),$ the probability of $ \rh $ erring can only be lower bounded by $ \nicefrac{3}{2}(\nicefrac{11}{243}-\nicefrac{1}{3}),$ which is negative! (all the errors might be in the recursive call $ \rI $).
Thus, we cannot guarantee a lower bound on this probability when making only $ 3 $ recursive calls in \cref{alg:splittingwith3}.
However, if we make more recursive calls, e.g., $ 27 $ (chosen large enough to make the argument work), we get that with probability at least $ 1/27 $ over $ \rI\sim \left\{  1,\ldots,27\right\}  $, it holds that $ \cA(\rS_{\rI,\sqcup};\rS_{\rI,\sqcap}\sqcup T) \geq \nicefrac{11}{243}$ and with probability at least $ 27/26(11/243-1/27)\approx0.009 $ over  $\rh\in \sqcup_{j\in\{  1,\ldots,27\}\backslash \rI } \cA(\rS_{1,j,\sqcup};\rS_{j,\sqcap}\sqcup T),$ we have that $\rh(x)\not=y.$ By $27\cdot (1/0.009) \leq 3160$, this gives 
\begin{align*}
    \footnotesize \ls_{\cD}^{\frac{11}{243}} (\erm(S;T))
    \leq 3160\negmedspace\negmedspace\negmedspace\negmedspace\negmedspace\negmedspace \max_{\stackrel{i,j\in \{1,\ldots,27  \}, i\not=j}{h'\in \erm(S_{j,\sqcup};S_{j,\sqcap}\sqcup T)}}\negmedspace\p_{(\rx,\ry)\sim\cD}\negmedspace\left[h'(\rx)\not=\ry,\avg(\erm(S_{i,\sqcup}; S_{i,\sqcap}\sqcup T))(\rx,\ry) \geq \frac{11}{243}\right]\negmedspace.
\end{align*}
This is precisely the scheme we propose in \cref{alg:splittingwith27}, with $\cA(\rS;T)=\{\cA(S')\}_{S'\in \cS(\rS;T)}$. Now, defining $ \widehat{\cA}_{\rt}(\rS) $ as $t$ voters drawn from $\cA(\rS;\emptyset)$  with $t=\Theta(\ln{\left(m/(\delta(d+\ln{\left(1/\delta \right)})) \right)}),$ and mimicking the analysis of \cref{sec:large-constant-sketch} yields that $ \ls_{\cD}^{\nicefrac{11}{243}}(\widehat{\cA}_{\rt}) \leq c(\tau+(d+\ln{\left(1/\delta \right)}))/m $. Finally, roughly following the above arguments for $\ls_{\cD}(\tie^{\nicefrac{11}{243}}(\cA_{\rt_{1}}'(\rS_{1}),\cA'_{\rt_{2}}(\rS_{2}); \htie )),$ but now with $\ls_{\cD}(\tie^{\nicefrac{11}{243}}(\widehat{\cA}_{\rt_{1}}(\rS_{1}),\widehat{\cA}_{\rt_{2}}(\rS_{2}); \htie )),$ we obtain, for the latter, the claimed generalization error of $ 2.1\tau+O(\sqrt{\tau(d+\ln{\left(1/\delta \right)})/m}+(d+\ln{\left(1/\delta \right)})/m).$

\vspace{-0.2 cm}
\begin{algorithm}
    \caption{Splitting algorithm $\cS$}\label{alg:splittingwith27}
    \KwIn{Training sequences $S, T \in (\cX \times \cY)^{*}$, where $|S| = 3^{k}$ for $k \in \mathbb{N}$.}
    \KwOut{Family of training sequences.}
    \If{$k \geq 6$\label{alg:splitting27if}}{
        Partition $S$ into $S_{1}, \ldots, S_{27}$, with $S_i$ being the $(i-1)|S|/27+1$ to the $i|S|/27$ training examples of $S$. Set for each $ i $
        \newline
        \textcolor{white}{halloooooooooooooo}   $
             S_{i,\sqcup} = S_{i}[1:3^{k-6}],\quad \quad
            S_{i,\sqcap} = S_{i}[3^{k-6}+1:3^{k-3}], $
            \newline
        \Return{$[\cS(S_{1,\sqcup}; S_{1,\sqcap} \sqcup T), \ldots, \cS(S_{27,\sqcup}; S_{27,\sqcap} \sqcup T)]$}
    }
    \Else{
        \Return{$S \sqcup T$}
    }
\end{algorithm}
\vspace{-0.2 cm}
With the above classifier in hand, we can now, on two new independent training sequences, obtain the classifier of \cite{hanneke2024revisiting} using the first sequences, and then on the second sequences choose the best of our classifier and the classifier of \cite{hanneke2024revisiting} as the final classifier.

\section{Conclusion}

We study the fundamental problem of agnostic PAC learning and provide
improved sample complexity bounds parametrized by $\tau$, the error of the best-in-class hypothesis.
Our results resolve the question of \citet{hanneke2024revisiting} asking for optimal error rates in the regime $\tau \approx \nicefrac{d}{m}$, and make progress on their questions regarding optimal learners for the full range of $\tau$ and efficient learners based upon majority votes of $\cerm$s. The most interesting future direction is whether an improved analysis of our voting scheme or a modification of it can lead to optimal algorithms for the full range of $\tau.$

\begin{ack}
    While this work was carried out, Mikael Møller Høgsgaard was supported by DFF Grant No. 9064-00068B, and the European Union (ERC,TUCLA, 101125203). Views and opinions expressed are however those of the author(s) only and do not necessarily reflect those of the European Union or the European Research Council. Neither the European Union nor the granting authority can be held responsible for them. 
    Julian Asilis was supported by the NSF CAREER Award CCF-223926 and the National Science Foundation Graduate
Research Fellowship Program under Grant No.\ DGE-1842487. Part of the work was done while Grigoris Velegkas was a PhD student at Yale University supported in part by the AI Institute for Learning-Enabled Optimization at Scale (TILOS).
\end{ack}

\bibliographystyle{plainnat}
\bibliography{refs.bib}

@article{hanneke2016optimal,
  title={The optimal sample complexity of PAC learning},
  author={Hanneke, Steve},
  journal={Journal of Machine Learning Research},
  volume={17},
  number={38},
  pages={1--15},
  year={2016}
}

@inproceedings{simon2015almost,
  title={An almost optimal PAC algorithm},
  author={Simon, Hans U},
  booktitle={Conference on Learning Theory},
  pages={1552--1563},
  year={2015},
  organization={PMLR}
}

@inproceedings{hanneke2024revisiting,
  title={Revisiting Agnostic PAC Learning},
  author={Hanneke, Steve and Larsen, Kasper Green and Zhivotovskiy, Nikita},
  booktitle={2024 IEEE 65th Annual Symposium on Foundations of Computer Science (FOCS)},
  pages={1968--1982},
  year={2024},
  organization={IEEE}
}

@article{valiant1984theory,
  title={A theory of the learnable},
  author={Valiant, Leslie G},
  journal={Communications of the ACM},
  volume={27},
  number={11},
  pages={1134--1142},
  year={1984},
  publisher={ACM New York, NY, USA}
}

@article{vapnik1964class,
  title={A class of algorithms for pattern recognition learning},
  author={Vapnik, Vladimir and Chervonenkis, A Ya},
  journal={Avtomat. i Telemekh},
  volume={25},
  number={6},
  pages={937--945},
  year={1964}
}

@misc{vapnik1974theory,
  title={Theory of pattern recognition},
  author={Vapnik, Vladimir and Chervonenkis, Alexey},
  year={1974},
  publisher={Nauka, Moscow}
}

@article{blumer1989learnability,
  title={Learnability and the Vapnik-Chervonenkis dimension},
  author={Blumer, Anselm and Ehrenfeucht, Andrzej and Haussler, David and Warmuth, Manfred K},
  journal={Journal of the ACM (JACM)},
  volume={36},
  number={4},
  pages={929--965},
  year={1989},
  publisher={ACM New York, NY, USA}
}

@article{ehrenfeucht1989general,
  title={A general lower bound on the number of examples needed for learning},
  author={Ehrenfeucht, Andrzej and Haussler, David and Kearns, Michael and Valiant, Leslie},
  journal={Information and Computation},
  volume={82},
  number={3},
  pages={247--261},
  year={1989},
  publisher={Elsevier}
}

@article{haussler1992decision,
  title={Decision theoretic generalizations of the PAC model for neural net and other learning applications},
  author={Haussler, David},
  journal={Information and computation},
  volume={100},
  number={1},
  pages={78--150},
  year={1992},
  publisher={Elsevier}
}

@inproceedings{aden2023optimal,
  title={Optimal pac bounds without uniform convergence},
  author={Aden-Ali, Ishaq and Cherapanamjeri, Yeshwanth and Shetty, Abhishek and Zhivotovskiy, Nikita},
  booktitle={2023 IEEE 64th Annual Symposium on Foundations of Computer Science (FOCS)},
  pages={1203--1223},
  year={2023},
  organization={IEEE}
}

@article{hogsgaard2025efficient,
  title={Efficient Optimal PAC Learning},
  author={H{\o}gsgaard, Mikael M{\o}ller},
  journal={arXiv preprint arXiv:2502.03620},
  year={2025}
}

@inproceedings{aden2024majority,
  title={Majority-of-three: The simplest optimal learner?},
  author={Aden-Ali, Ishaq and H{\o}andgsgaard, Mikael M{\o}ller and Larsen, Kasper Green and Zhivotovskiy, Nikita},
  booktitle={The Thirty Seventh Annual Conference on Learning Theory},
  pages={22--45},
  year={2024},
  organization={PMLR}
}

@inproceedings{baggingoptimal,
  author       = {Kasper Green Larsen},
  editor       = {Gergely Neu and
                  Lorenzo Rosasco},
  title        = {Bagging is an Optimal {PAC} Learner},
  booktitle    = {The Thirty Sixth Annual Conference on Learning Theory, {COLT} 2023,
                  12-15 July 2023, Bangalore, India},
  series       = {Proceedings of Machine Learning Research},
  volume       = {195},
  pages        = {450--468},
  publisher    = {{PMLR}},
  year         = {2023},
  url          = {https://proceedings.mlr.press/v195/larsen23a.html},
  bibsource    = {dblp computer science bibliography, https://dblp.org}
}

@book{anthony2009neural,
  title={Neural network learning: Theoretical foundations},
  author={Anthony, Martin and Bartlett, Peter L},
  year={2009},
  publisher={cambridge university press}
}

@article{boucheron2005theory,
  title={Theory of classification: A survey of some recent advances},
  author={Boucheron, St{\'e}phane and Bousquet, Olivier and Lugosi, G{\'a}bor},
  journal={ESAIM: probability and statistics},
  volume={9},
  pages={323--375},
  year={2005},
  publisher={EDP Sciences}
}

@book{devroye1996probabilistic,
  title={A Probabilistic Theory of Pattern Recognition},
  author={Devroye, Luc and Gy{\"o}rfi, L{\'a}szl{\'o} and Lugosi, G{\'a}bor},
  publisher={Springer},
  year={1996}
}

@book{Understandingmachinelearningfromtheory, place={Cambridge}, title={Understanding Machine Learning: From Theory to Algorithms}, publisher={Cambridge University Press}, author={Shalev-Shwartz, Shai and Ben-David, Shai}, year={2014}}

@inproceedings{wagenmaker2022first,
  title={First-order regret in reinforcement learning with linear function approximation: A robust estimation approach},
  author={Wagenmaker, Andrew J and Chen, Yifang and Simchowitz, Max and Du, Simon and Jamieson, Kevin},
  booktitle={International Conference on Machine Learning},
  pages={22384--22429},
  year={2022},
  organization={PMLR}
}

@inproceedings{Maurer2009EmpiricalBB,
  title={Empirical Bernstein Bounds and Sample-Variance Penalization},
  author={Andreas Maurer and Massimiliano Pontil},
  booktitle={Annual Conference Computational Learning Theory},
  year={2009},
  url={https://api.semanticscholar.org/CorpusID:17090214}
}

@inproceedings{bousquet2020proper,
  title={Proper learning, Helly number, and an optimal SVM bound},
  author={Bousquet, Olivier and Hanneke, Steve and Moran, Shay and Zhivotovskiy, Nikita},
  booktitle={Conference on Learning Theory},
  pages={582--609},
  year={2020},
  organization={PMLR}
}

@article{bousquet2002stability,
  title={Stability and generalization},
  author={Bousquet, Olivier and Elisseeff, Andr{\'e}},
  journal={Journal of machine learning research},
  volume={2},
  number={Mar},
  pages={499--526},
  year={2002}
}
\appendix
\section{Preliminaries for Proof}\label{sec:preliminariesproof}
In this section we give the preliminaries for the proof. For the reader's convenience, we restate \cref{alg:splitting27if}.
\setcounter{algocf}{1}
\begin{algorithm}
    \caption{Splitting algorithm $\cS$}
    \KwIn{Training sequences $S, T \in (\cX \times \cY)^{*}$, where $|S| = 3^{k}$ for $k \in \mathbb{N}$.}
    \KwOut{Family of training sequences.}
    \If{$k \geq 6$}{
        Partition $S$ into $S_{1}, \ldots, S_{27}$, with $S_i$ being the $(i-1)|S|/27+1$ to the $i|S|/27$ training examples of $S$. Set for each $ i $ 
        \newline 
        \textcolor{white}{halloooooooooooooo}   $
             S_{i,\sqcup} = S_{i}[1:3^{k-6}],\quad \quad
            S_{i,\sqcap} = S_{i}[3^{k-6}+1:3^{k-3}], $
            \newline 
        \Return{$[\cS(S_{1,\sqcup}; S_{1,\sqcap} \sqcup T), \ldots, \cS(S_{27,\sqcup}; S_{27,\sqcap} \sqcup T)]$}
    }
    \Else{
        \Return{$S \sqcup T$}
    }
\end{algorithm}  

We first observe that for an input training sequence $ m=|S|=3^{k} $,  the above algorithm makes $ l$ recursive calls where $ l\in\mathbb{N} $ satisfies $ k-6(l-1)\geq 6 $ and $ k-6l<6 $, that is, $ l $ is the largest number such that  $ k/6\geq l$. 
As $ l $ is a natural number, we get that $ l=\lfloor k/6\rfloor $. 
Furthermore, since $k=\log_{3}(m)  $ we get that $ l=\lfloor \log_{3}(m)/6\rfloor $. 
For each of the $ l $ recursive calls 27 recursions are made. Thus, the total number of training sequences created in $ \cS $ is $ 27^{l}\leq  3^{3\log_{3}(m)/(2\cdot6)}=m^{1/(4\ln(3))}\geq m^{0.22}$.
In what follows, we will use the quantity $ s_{\sqcap}$, which we define as $ \frac{|S|}{|S_{1,\sqcap}|}$ when running $ \cS(S;T) $ with $ S;T\in (\cX\times \cY)^{*} $ such that $ |S|=3^{k} $ and $ k\geq6 $. We notice that  
\begin{align}\label{lem:ratios}
    s_{\sqcap}\coloneqq\frac{|S|}{|S_{1,\sqcap}|}=\frac{3^{k}}{3^{k-3}-3^{k-6}}=\frac{3^{k}}{3^{k-3}(1-\frac{1}{3^{3}})}=\frac{27}{1-\frac{1}{27}}.
\end{align}
This ratio $ s_{\sqcap} $ will later in the proof show up as a constant, $ |S_{i}|/(|S|s_{{\sqcap}})=(|S_{i,\sqcup}|+|S_{i,\sqcap}|)/|S_{i,\sqcap}|=1/(1-1/27) $,  multiplied onto $ \tau.$ Thus, from this relation we see that if the split of $S_{i}$ is imbalanced so that $|S_{i,\sqcap}| $ is larger than $ |S_{i,\sqcup}| $ the constant multiplied on to $ \tau $ become smaller.      

Furthermore, in what follows, for the set of training sequences generated by $ \cS(S;T) $ and a $ \cerm $-algorithm $ \cA $, we  write $ \erm(S;T) $ for the set of classifiers the $ \cerm $-algorithm outputs when run on the training sequences in $ \cS(S;T) $, i.e., $ \erm(S;T)=\{ \erm(S') \}_{S'\in\cS(S;T)}  $, where this is understood as a multiset if the output of the $ \cerm $-algorithm is the same for different training sequences in $ \cS(S;T).$ 
Furthermore for an example $ (x,y) $ , we define $ \avg(\erm(S;T))(x,y)=\sum_{h\in \erm(S;T)}\ind\{ h(x)\not=y \}/|\erm(S;T)| $, i.e., the average number of incorrect hypotheses in $ \erm(S;T) .$  
We notice that by the above comment about $ \cS(S;T) $ having size at least $ m^{0.22},$ we have that $ \erm(S;T) $ contains just as many hypothesis, each of which is the output of an $ \cerm $ run on a training sequence of $ \cS(S;T).$ Thus as allotted to earlier our algorithm do not run on all the sub training sequences created by $ \cS(S;T),$ as it calls the $ \erm $ algorithm $ O(\ln{\left(m/(\delta(d+\ln{\left(1/\delta \right)})) \right)}) $-times. Which leads us considering the following classifier.

For a natural number $ t,$ we let $ \widehat{\erm}_{\rt}(S;T),$ be the random collection of $ t $ hypotheses drawn uniformly with replacement from $ \erm(S;T),$ with the draws being independent, where we see $ \widehat{\erm}_{\rt}(S;T) $ as a multiset so allowing for repetitions. 
We remark here that we will overload notation and use $ \rt $ (so $ t $ in bold font) to denote the randomness used to draw the $ t $ hypotheses from $ \erm(S;T)$ in the following analysis of $ \widehat{\erm}_{\rt}(S;T).$ 
Intuitively one can think of $ \widehat{\erm}_{\rt} $ as a bagging algorithm where the subsampled training sequences are restricted to subsets of $ \cS(S;T) $ rather than sampling with replacement from the training examples of $ S $ and $ T.$ In what follows we will consider this algorithm parametrized by $ t=O(\ln{\left(m/(\delta(d+\ln{\left(1/\delta \right)})) \right)}) $ leading to a classifier with the same order of call to the $ \cerm $ as stated in. Similarly to $ \cA(S;T) $ we also define $ \avg(\widehat{\erm}_{\rt}(S;T))(x,y)=\sum_{h\in \widehat{\erm}_{\rt}(S;T)}\ind\{ h(x)\not=y \}/|\widehat{\erm}_{\rt}(S;T)| $

Now, for a distribution $ \cD $ over $ \cX\times \{ -1,1\},$ training sequences $ S;T\in \left(\cX\times\{  -1,1\} \right)^{*},$ and $ \alpha\in[0,1] $  we will use $ \ls_{\cD}^{\alpha}(\erm(S;T))=\p_{(\rx,\ry)\sim \cD}\left[\avg(\erm(S;T))(\rx,\ry) \geq \alpha\right],$ i.e., the probability of at least a $ \alpha $-fraction of the hypotheses in $ \erm(S;T) $ erroring on a new example drawn according to $ \cD $. As above we also define $ \ls_{\cD}^{\alpha}(\widehat{\erm}_{\rt}(S;T))=\p_{(\rx,\ry)\sim \cD}\left[\avg(\widehat{\erm}_{\rt}(S;T))(\rx,\ry) \geq \alpha\right],$ for $ \widehat{\erm}_{\rt}(S;T). $ In the following we will for the case where $ T $ is the empty training sequence $ \emptyset $   us $ \widehat{\cA}_{\rt}(S)=\widehat{\cA}_{\rt}(S;\emptyset)$.

\subsection{Difficulties of the Proof}

Let us take the opportunity to briefly describe the challenges associated with settling the sample complexity of agnostic learning in the regime $\tau = O(d/m)$. In particular, we will demonstrate that a straightforward multiplicative Chernoff argument can \emph{not} yield a result as general as \cref{thm:main-result}. 

To begin, notice that the true errors of $N$ hypotheses can indeed be confidently estimated to within error $\tau$ using $O(\log(N/\delta)/\tau)$ samples, due to multiplicative Chernoff. However, if $\tau=O(\log(N/\delta)/m)$, then the above yields a vacuous bound. An additive Chernoff bound would also in this case imply an additive $ O(\log(N/\delta)/m)$ error term. In the paper, we consider classes of finite VC dimension $d$ but arbitrary, possibly infinite cardinality $N$, thereby preventing us from employing this bound. Even if one were to bound $N$ as $O((m/d)^d)$ using Sauer-Shelah, this would imply that multiplicative Chernoff is unhelpful when $\tau = O((d \log(m/d)+ \log(1/\delta))/m)$. This is a wide regime containing, for instance, $\tau = O(d/m)$, a setting in which we demonstrate that our algorithm is optimal. Furthermore, an additive Chernoff bound would incur a suboptimal additive $O(d\log(m/d)/m)$ term.

However, let us assume that a Chernoff bound argument for infinite hypothesis classes could yield \cref{thm:main} for all $\tau$ (e.g., using an $\epsilon$-covering argument). 
In this case, one would have demonstrated that an empirical risk minimizer $h'$ has error $ 2.1\tau+O(\sqrt{\frac{\tau(d+\ln{\left(1/\delta \right)})}{m}})+\frac{d+\ln{\left(1/\delta \right)}}{m}.$ 
Then, taking $\tau=0$ (i.e, considering the realizable case), would demonstrate that ERM results in an error of $O(\frac{d+\ln{\left(1/\delta \right)}}{m})$. However, this is not generally true, owing to the worst-case lower bound on ERM learners of $O(\frac{d\ln{\left(1/\varepsilon \right)}+\ln{\left(1/\delta \right)}}{m})$, due to \citep[Theorem~11]{bousquet2020proper}. This strongly suggests that a Chernoff bound argument using a union bound over each function in the hypothesis class (or a discretization of it) cannot yield a result as general as Theorem 1.1, which holds across all $\tau$. 

Finally, we remark that our learner for \cref{thm:main} is agnostic to $\tau$ (i.e., is oblivious to the value of $\tau$), whereas the previous Chernoff procedure requires knowledge of $\tau$. This is a notable distinction, as precise knowledge of $\tau$ will typically be unavailable to the learner. In particular, ERM-based estimates of $\tau$ will be uninformative in the regime $\tau = o(d/m)$.

\appendix 
\crefalias{section}{appendix}

\section{Analysis of $\widehat{\cA}_{\rt}$}

As described in the proof sketch, we require a bound on $\ls_{\cD}^{10/243}\cA(S;\emptyset)$ in order to upper bound $ \ls_{\cD}^{11/243}(\widehat{\cA}_{\rt})$.  
Thus, we now present our error bound for \cref{alg:splittingwith27} when running $ \erm $ on each dataset generated on $\cS(\rS,\emptyset)$. (We assume that $ |\rS| =3^{k}$ for $ k\in\mathbb{N}$, at the cost of discarding a constant fraction of training points.)

\begin{lemma}\label{lem:upperbound}
    There exists a universal constant $ c \geq 1$ such that: For any hypothesis class $ \cH $ of VC dimension $ d $, distribution $ \cD $ over $ \cX\times\cY,$ failure parameter $ 0<\delta<1 $, training sequence size $ m=3^{k} $ for $ k\geq 6 $, and training sequence $  \rS\sim \cD^{m},$ with probability at least $ 1-\delta $ over $ \rS $ one has that  
    \begin{align*}
        \ls_{\cD}^{10/243}\big(\erm(\rS;\emptyset)\big)  
        \leq c\tau+\frac{c\left(d+\ln{(e/\delta )}\right)}{m}.
    \end{align*}
\end{lemma}

Let us defer the proof of \Cref{lem:upperbound} 
for the moment, and proceed with presenting the main theorem of this section, assuming the claim of \Cref{lem:upperbound}. 

\begin{theorem}\label{thm:main}
    There exists a universal constant $ c \geq1$ such that: For any hypothesis class $ \cH $ of VC dimension $ d $, distribution $ \cD $ over $ \cX\times\cY,$ failure parameter $ 0<\delta<1 $, training sequence size $ m=3^{k} $ for $ k\geq 6 $, training sequence $  \rS\sim \cD^{m},$ and sampling size $ t\geq4\cdot 243^{2}\ln{\left(2m/(\delta(d+\ln{\left(1/\delta \right)})) \right)},$ we have with probability at least $ 1-\delta $ over $ \rS $ and the randomness $ \rt $  used to draw $ \widehat{\erm}_{\rt}(\rS) $ that:
    \begin{align*}
     \ls_{\cD}^{11/243}(\widehat{\erm}_{\rt}(\rS) )\leq  c\tau+\frac{c\left(d+\ln{(e/\delta )}\right)}{m}.
    \end{align*}
\end{theorem}
\begin{proof}
Let $ \widehat{\erm}_{\rt}(S;\emptyset)=\{\rh_{1},\ldots,\rh_{t}  \}$, considered as a multiset, and recall that the $\rh_{i} $ are drawn uniformly at random from $ \erm(S;\emptyset)=\{ \erm(S') \}_{S'\in\cS(S;\emptyset)} $, which is likewise treated as a multiset.
Let $ E_{\rS} $ denote the event 
\[ E_{\rS}= \left \{(x,y) \; \Big| \sum_{h\in \erm(\rS;\emptyset)}\frac{\ind\{ h(x)\not=y \}}{|\erm(\rS;\emptyset)|} \geq \frac{10}{243}  \right \} \]
and 
\[ \bar{E}_{\rS}= \left \{(x,y) \; \Big| \sum_{h\in \erm(\rS;\emptyset)}\frac{\ind\{ h(x)\not=y \}}{|\erm(\rS;\emptyset)|} < \frac{10}{243}  \right \} \]
its complement. Now fix a realization $S$ of $ \rS $. Using the fact that $ \p\left[A\right]=\p\left[A\cap B\right]+\p\left[A\cap \bar{B}\right]$, we have that 
\begin{align}\label{eq:uppsampledbound1}
\ls_{\cD}^{11/243}(\widehat{\erm}_{\rt}(S;\emptyset) )
    &= \p_{(\rx,\ry)\sim \cD}\left[\sum_{i=1}^{t} \ind\{   \rh_{t}(\rx)\not=\ry\}/t\geq \frac{11}{243}, E_{S}\right] \notag \\ 
    & \qquad +\p_{(\rx,\ry)\sim \cD}\left[\sum_{i=1}^{t} \ind\{   \rh_{t}(\rx)\not=\ry\}/t\geq\frac{11}{243}, \bar{E}_{S}\right] \notag
    \\
    &\leq \p_{(\rx,\ry)\sim\cD}\big [E_{S}\big]+\p_{(\rx,\ry)\sim \cD}\left[\sum_{i=1}^{t} \ind\{   \rh_{t}(\rx)\not=\ry\}/t\geq\frac{11}{243}, \bar{E}_{S}\right]. 
\end{align} 
We now bound each term in the above, i.e., in \Cref{eq:uppsampledbound1}.

In pursuit of bounding the second term, consider a labeled example $ (x,y)\in\bar{E}_{S}$. We may assume that $ \bar{E}_{S} $ is non-empty, as otherwise the term is simply $0$.
Now, for any such labeled example $(x,y) \in \bar{E}_{S}$ we have that  $ \sum_{i=1}^{t} \ind\{ \rh_{i}(x)=y \} $  has expectation 
\[ \mu_{(x,y)}:=\e_{\rt}\left[\sum_{i=1}^{t} \ind\{ \rh_{i}(x)=y \}\right]=t\cdot\left(\sum_{h\in \erm(\rS;\emptyset)} \frac{\ind\{ h(x)=y \}}{|\erm(\rS;\emptyset)|} \right)\geq (1-\nicefrac{10}{243} \cdot t ) \geq t/2, \] 
where the final inequality follows from the fact that $ (x,y)\in \bar{E}_{S}$. (Recall that we use the boldface symbol $\rt$ to denote the randomness underlying the random variables $ \rh_{1},\ldots,\rh_{t}$.)  
Now, since $ \ind\{ \rh_{i}(x)\not=y \} $ is a collection of i.i.d.\ $\{0,1\}-$random variables, we have by the Chernoff inequality that 
\begin{align*}
 \p_{\rt} \left[\sum_{i=1}^{t} \frac{\ind\{ \rh_{i}(x)=y \}}{t} \leq \left (1-\frac{1}{243} \right)\mu_{(x,y)}\right] &\leq \exp{\left(\frac{-\mu_{(x,y)}}{2\cdot 243^{2}}\right)} \\
 & \leq \frac{\delta\big(d+\ln{\left(1/\delta \right)}\big)}{2m}, 
\end{align*}
where the final inequality follows from the fact that $ \mu_{(x,y)}\geq \nicefrac t2 $ and  
\[ t\geq 4\cdot 243^{2}\ln\left(\frac{2m}{\delta\big(d+\ln{\left(1/\delta \right)}\big)} \right). \]  
The above implies that with probability at least $ 1-\frac{\delta(d+\ln{\left(1/\delta \right)})}{2m}$, we have  
\[ \sum_{i=1}^{t} \frac{\ind\{ \rh_{i}(x)=y \}}{t} > \left( 1-\frac{1}{10} \right) \mu_{(x,y)} \geq \left( 1- \frac{1}{243} \right) \left( 1 - \frac{10}{243} \right) = 1 - \frac{11}{243}. \] 
This further implies that $ \sum_{i=1}^{t} \ind\{ \rh_{i}(x)\not=y \}/t< 11/243$  with probability at most $ \frac{\delta(d+\ln{\left(1/\delta \right)})}{2m}$.
As we demonstrated this fact for any pair $(x,y)\in \bar{E}_{S}$, an application of Markov's inequality yields that 
\begin{align}\label{eq:uppsampledbound2}
 &\p_{\rt}\left[\p_{(\rx,\ry)\sim \cD}\left[\sum_{i=1}^{t} \ind\{   \rh_{t}(\rx)\not=\ry\}/t\geq\frac{11}{243}, \bar{E}_{S}\right]\geq \frac{(d+\ln{\left(1/\delta \right)})}{m}\right] \hspace{2cm} \notag 
 \\
& \qquad \quad \leq \frac{(d+\ln{\left(1/\delta \right)})\e_{(\rx,\ry)\sim \cD}\left[ \p_{\rt} \left[\sum_{i=1}^{t} \ind\{   \rh_{t}(\rx)\not=\ry\}/t\geq\frac{11}{243}\right] \ind \{   \bar{E}_{S}\} \right]}{m} \notag \\
& \qquad \quad \leq \frac{\delta}{2}. 
\end{align}
Note that the first inequality follows from an application of Markov's inequality and the observation that $ \bar{E}_{S} $ depends only upon $ (\rx,\ry)$ and $ \rt $, which are independent from one another, meaning we can swap the order of expectation. 
The final inequality follows from the bound on the probability of $ \sum_{i=1}^{t} \ind\{   \rh_{t}(\rx)\not=\ry\}/t\geq\frac{11}{243}  $ happening over $ \rt $,  for $ x,y\in\bar{E}_{S}$.

Thus, we conclude that with probability at least $1 - \delta/2$ over $\rt$, the random draw of the hypothesis in $ \widehat{\erm}_{\rt}(S;\emptyset) $ is such that  
\[ \p_{(\rx,\ry)\sim \cD}\left[\sum_{i=1}^{t} \frac{\ind\{   \rh_{t}(\rx)\not=\ry\}}{t} \geq \frac{11}{243} , \bar{E}_{S}\right]\leq \frac{d+\ln{\left(1/\delta \right)}}{2m}. \]  
Furthermore, as we showed this for any realization $S$  of $\rS$ (and $ \rt $ and $ \rS $ are independent), we conclude that, with probability at least $ 1-\delta/2$ over both $ \rt $ and $ \rS $,
\[ \p_{(\rx,\ry)\sim \cD}\left[\sum_{i=1}^{t} \frac{\ind\{   \rh_{t}(\rx)\not=\ry\}}{t} \geq \frac{11}{243}, \bar{E}_{\rS}\right]\leq \frac{d+\ln{\left(1/\delta \right)}}{m}.\] 
Furthermore, by \cref{lem:upperbound}, we have that with probability at least $ 1-\delta/2 $ over $ \rS $, 
\begin{align}\label{eq:uppsampledbound3}
    \p_{(\rx,\ry)\sim\cD}\left[E_{\rS}\right]\leq   c\tau +\frac{c\left(d+\ln{(2e/\delta )}\right)}{m}.
\end{align}
Furthermore, as this event does not depend on the randomness $ \rt $ employed in drawing hypotheses for $ \widehat{\erm}_{\rt}(\rS),$ we conclude that the above also holds with probability at least $ 1-\delta/2 $ over both $ \rS $ and $ \rt.$ 
Now, applying a union bound over the event in \Cref{eq:uppsampledbound2} and \Cref{eq:uppsampledbound3}, combined with the bound on $ \ls_{\cD}(\widehat{\erm}_{\rt}(\rS)) $ in \Cref{eq:uppsampledbound1}, we get that, with probability at least $ 1-\delta $ over $ \rS $ and $ \rt $, it holds that 
\begin{align*}
    \ls_{\cD}^{11/243}(\widehat{\erm}_{\rt}(\rS))\leq c\tau +\sqrt{\frac{c\tau \left(d+\ln{(2e/\delta )}\right)}{m}}+\frac{(1+c)\left(d+\ln{(2e/\delta )}\right)}{m}.
\end{align*}
As $c$ is permitted by any absolute constant, this concludes the proof. 
\end{proof}

We now proceed to give the proof of \Cref{lem:upperbound}. Doing so will require two additional results, the first of which relates the empirical error of a hypothesis $h$ to its true error.

\begin{lemma}[\cite{Understandingmachinelearningfromtheory} Lemma B.10]\label{lem:additiveerrorhstar}
Let $\cD$ be a distribution over $\cX \times \{-1, 1\}$, $h \in \{-1, 1\}^\cX$ be a hypothesis, $\delta \in (0, 1)$ a failure parameter, and $m \in \N$ a natural number. Then, 
    \begin{align*}
        \p_{\rS \sim \cD^m}\left[\ls_{\rS}(h)\leq \ls_{\cD}(h) +\sqrt{\frac{2\ls_{\cD}(h)\ln{(1/\delta )}}{3m}}+\frac{2\ln{(1/\delta )}}{m} \right] 
        \geq 1-\delta,
    \end{align*}
and 
\begin{align*}
    \p_{\rS \sim \cD^m} \left[\ls_{\cD}(h)\leq \ls_{\rS}(h) +\sqrt{\frac{2\ls_{\rS}(h)\ln{(1/\delta )}}{m}}+\frac{4\ln{(1/\delta )}}{m} \right] 
    \geq 1-\delta.
\end{align*}  
\end{lemma}

The second result we require is one which bounds the error of $ \erm(S;T)$ for arbitrary training sets $T$ (i.e., not merely $T = \emptyset$, as we have previously considered). 

\begin{theorem}\label{thm:upperbound}
    There exists, universal constant $ c\geq1 $ such that: For any a hypothesis class $\cH$ of VC dimension $ d $, any distribution $ \cD $ over $ \cX\times\{  -1,1\}  $, any failure parameter $\delta \in (0, 1)$, any training sequence size $ m=3^{k} $, any training sequence $ T\in (\cX\times\{-1,1  \})^{\star} $, and a random training sequence $ \rS\sim\cD^{m},$ it holds with probability at least $ 1-\delta $ over $ \rS $ that:        
    \begin{align*}
        \ls_{\cD}^{10/243}(\erm(\rS;T))
        \leq  \max\limits_{S'\in\cS(\rS;T)}\frac{c\Sigma_{\not=}(\hs\negmedspace,S')}{m/s_{\sqcap}}+\frac{c\left(d+\ln{(1/\delta )}\right)}{m},
    \end{align*}
    where
    \[ \Sigma_{\not=}(\hs\negmedspace,S') =\sum\limits_{(x,y)\in S'} \ind\{ \hs(x)\not=y \}. \] 
\end{theorem}

Let us first give the proof of \Cref{lem:upperbound} assuming  \Cref{thm:upperbound}, and subsequently offer the proof of \Cref{thm:upperbound}.

\begin{proofof}{\Cref{lem:upperbound}}
First note that for each $i \in\{ 1,\ldots,27\}$, 
\begin{align*}
\max\limits_{S'\in \cS(\rS_{i,\sqcup}; \rS_{i,\sqcap})} \frac{\Sigma_{\not=}(\hs\negmedspace,S')}{m/s_{\sqcap}} & =\max\limits_{S'\in \cS(\rS_{i,\sqcup};\rS_{i,\sqcap})} \sum\limits_{(x,y)\in S'}\frac{\ind\{ \hs(x)\not=y \}}{(m/s_{\sqcap})} \\
&\leq \frac{|\rS_{i,\sqcap}\sqcup\rS_{i,\sqcup}|}{(m/s_{\sqcap})}
\sum_{(x,y)\in \rS_{i,\sqcap}\sqcup\rS_{i,\sqcup}}\frac{\ind\{ \hs(x)\not=y \}}{|\rS_{i,\sqcap}\sqcup  \rS_{i,\sqcup}|} \\ 
&= 2\ls_{\rS_{i}}(\hs). 
\end{align*}
The inequality follows from the fact that any $S'\in \cS(\rS_{i,\sqcup};\rS_{i,\sqcap})$ satisfies $S'\sqsubset \rS_{i,\sqcap}\sqcup \rS_{i,\sqcup}$ and the final equality uses the facts that $ |\rS_{i,\sqcap}\sqcup\rS_{i,\sqcup}|=|\rS_{i}|=m/3^{3}$ and $ s_{\sqcap}=3^{3}/(1-1/27) $.  
Thus, invoking \Cref{lem:additiveerrorhstar} over $\rS_{i}  $ for $ i\in\{  1,\ldots,27\}  $ with failure parameter $  \delta/28$, we have by a union bound that with probability at least $ 1-27\delta/28 $ over $ \rS $, each $ i\in\{  1,\ldots,27\}$ satisfies
\begin{align}\label{eq:upperboundlem1}
\hspace{-1.5 cm}
    \max\limits_{S'\in \cS(\rS_{i,\sqcup};\rS_{i,\sqcap})}\frac{\Sigma_{\not=}(\hs\negmedspace,S')}{m/s_{\sqcap}}
 \leq  
2
 \left(
    \tau+
 \left(\sqrt{\tau\frac{6\ln{(28/\delta )}}{3m}}
 +\frac{6\ln{(28/\delta )}}{m}\right)
 \right)\leq 4 \left(
    \tau+\frac{6\ln{(28/\delta )}}{m}\right)
\end{align}
where we have used that $ \sqrt{ab}\leq a+b.$ 
Furthermore, by \Cref{thm:upperbound} we have that with probability at least $ 1-\delta/28 $ over $ \rS $, 
\begin{align*}
   \ls_{\cD}^{10/243}(\cA(\rS;\emptyset))\leq\max\limits_{S'\in\cS(\rS;T)}\frac{c\Sigma_{\not=}(\hs\negmedspace,S')}{m/s_{\sqcap}}+\frac{c\left(d+\ln{(28/\delta )}\right)}{m}
\end{align*}
Again invoking a union bound, we have with probability at least $ 1-\delta $ over $ \rS $ that  
\begin{align*}
\ls_{\cD}^{10/243}(\erm(\rS;\emptyset)) &\leq \max\limits_{S'\in\cS(\rS;T)}\frac{c\Sigma_{\not=}(\hs\negmedspace,S')}{m/s_{\sqcap}}+\frac{c\left(d+\ln{(28/\delta )}\right)}{m}
\\
&\leq 
4 c\left(
    \tau+\frac{6\ln{(28/\delta )}}{m}\right)+\frac{c\left(d+\ln{(28/\delta )}\right)}{m}
    \\
&\leq 7c\tau+\frac{7c\left(d+\ln{(28/\delta )}\right)}{m},
\end{align*}
where the first inequality follows from \Cref{eq:upperboundlem1}. This concludes the proof. 
\end{proofof}

We now direct our attention to proving \Cref{thm:upperbound}. We will make use of another set of two lemmas, the first of which permits us to make a recursive argument over $ \erm $-calls based on sub-training sequences created in \Cref{alg:splittingwith27}.

\begin{lemma}\label{lem:recursivelemma}
Let $ S, T\in (\cX\times \cY)^{*} $ with $ |S|=3^{k}$ for $ k\geq 6$, and let $\cD$ be a distribution over $\cX \times \cY$. Then, 
\begin{align*}
\hspace{-0.6 cm}
\ls_{\cD}^{10/243}\big(\cS(S;T)\big) \; \leq \; 5687 \max_{\stackrel{i,j\in \{1,\ldots,27  \}}{i\not=j} }\max_{h'\in \erm(S_{j,\sqcup};S_{j,\sqcap}\sqcup T)}\p_{(\rx,\ry)\sim\cD}\left[h'(\rx)\not=\ry,\avg(\erm(S_{i,\sqcup}; S_{i,\sqcap}\sqcup T))(\rx,\ry) \geq \frac{10}{243}\right]. 
\end{align*}
\end{lemma}
\begin{proof}
Let $(x,y)$ be an example such that  
\[ \sum_{h\in \erm(S;T)} \frac{\ind\{ h(x)\not=y \}}{|\erm(S;T)|} \geq \frac{10}{243}. \]
As $k\geq 6$, \Cref{alg:splittingwith27} calls itself when called with $(S;T)$. Furthermore, as each of the 27 calls produce an equal number of subtraining sequences, it must be the case that
\begin{align*}
    \frac{10}{243} \leq \avg \big( \erm(S;T) \big) (x,y) = \sum_{i=1}^{27}\frac{1}{27}\sum_{h\in \erm(S_{i,\sqcup}; S_{i,\sqcap} \sqcup T)} \frac{\ind\left\{ h(x)\not=y \right\} }{|\erm(S_{i,\sqcup}; S_{i,\sqcap} \sqcup T)|}. 
\end{align*} 
This in turn implies that there exists an $\hat{i} \in [27]$ satisfying the above inequality, i.e., such that
\begin{align*}
    \frac{10}{243}\leq\avg(\erm(S_{\hat{i},\sqcup}; S_{\hat{i},\sqcap}\sqcup T))(x,y)= \sum_{h\in \erm(S_{\hat{i},\sqcup}; S_{\hat{i},\sqcap} \sqcup T)} \frac{\ind\left\{ h(x)\not=y \right\} }{|\erm(S_{\hat{i},\sqcup}; S_{\hat{i},\sqcap} \sqcup T)|} \nonumber.
\end{align*}
We further observe that for any $i\in [27]$, 
\begin{align*}
    \frac{10}{243} &\leq \sum_{j=1}^{27}\frac{1}{27}\sum_{h\in \erm(S_{j,\sqcup}; S_{j,\sqcap} \sqcup T)} \frac{\ind\left\{ h(x)\not=y \right\} }{|\erm(S_{j,\sqcup}; S_{j,\sqcap} \sqcup T)|}
    \\
    &\leq
    \sum_{j\in \{1,\ldots,27  \}\backslash i }\frac{1}{27}\sum_{h\in \erm(S_{j,\sqcup}; S_{j,\sqcap} \sqcup T)} \frac{\ind\left\{ h(x)\not=y \right\} }{|\erm(S_{j,\sqcup}; S_{j,\sqcap} \sqcup T)|} +\frac{1}{27}.
\end{align*} 
This implies, again for any arbitrary choice of $i \in [27]$, that 
\begin{align*}
  \frac{1}{243}\leq
    \sum_{j\in \{1,\ldots,27  \}\backslash i }\frac{1}{27}\sum_{h\in \erm(S_{j,\sqcup}; S_{j,\sqcap} \sqcup T)} \frac{\ind\left\{ h(x)\not=y \right\} }{|\erm(S_{j,\sqcup}; S_{j,\sqcap} \sqcup T)|} \nonumber.
\end{align*}
Simply multiplying both sides by $27 / 26$, we have that 
\begin{align}\label{eq:relationerror}
    \frac{1}{234}\leq
      \sum_{j\in \{1,\ldots,27  \}\backslash i }\frac{1}{26}\sum_{h\in \erm(S_{j,\sqcup}; S_{j,\sqcap} \sqcup T)} \frac{\ind\left\{ h(x)\not=y \right\} }{|\erm(S_{j,\sqcup}; S_{j,\sqcap} \sqcup T)|}.
\end{align}
Using the above, we can conclude that when $(x,y)$ is such that $\avg(\erm(S;T))(x,y)\geq \frac{10}{243}$, then there exists an $ i\in [27]$ with $\avg(\erm(S_{i,\sqcup}; S_{i,\sqcap}\sqcup T))(x,y) \geq \frac{10}{243}$. Then by \Cref{eq:relationerror}, at least a $\nicefrac{1}{234}-$fraction of hypotheses in $ \bigsqcup_{j\in \{1,\ldots,27  \} \backslash i}\erm(S_{j,\sqcup}; S_{j,\sqcap} \sqcup T) $ err on $ (x,y).$ Thus, if we let $ \rI $ be drawn uniformly at random from $ \{1,\ldots,27  \} $ and $ \rh $ be drawn uniformly at random from $ \bigsqcup_{j\in \{1,\ldots,27  \} \backslash \rI}\erm(S_{j,\sqcup}; S_{j,\sqcap} \sqcup T) $, then by the law of total probability we have that  
\begin{align*}
 &\p_{\rI,\rh}\left[\rh(x)\not=y,\avg(\erm(S_{\rI,\sqcup}; S_{\rI,\sqcap}\sqcup T))(x,y) \geq \frac{10}{243}\right] 
 \\
 & \qquad = \p_{\rI,\rh}\left[\rh(x)\not=y| \avg(\erm(S_{\rI,\sqcup}; S_{\rI,\sqcap}\sqcup T))(x,y) \geq \frac{10}{243}\right]\p_{\rh}\left[\avg(\erm(S_{\rI,\sqcup}; S_{\rI,\sqcap}\sqcup T))(x,y)\geq \frac{10}{243}\right]
 \\
 &\qquad \geq \frac{1}{234} \cdot \frac{10}{243} \\ 
 &\qquad \geq \frac{1}{5687}. 
\end{align*}
This implies in turn that 
\begin{align}\label{eq:relationerror1}
\hspace{-1.5 cm} 5687\p_{\rI,\rh}\left[\rh(x)\not=y,\avg(\erm(S_{\rI,\sqcup}; S_{\rI,\sqcap}\sqcup T))(x,y)\geq \frac{10}{243}\right]\geq \ind \left \{\avg(\erm(S;T))(x,y) \geq \frac{10}{243} \right \}.  
\end{align}
Taking expectations with respect to $(\rx,\ry)\sim \cD $, we have
\begin{align*}
5687\e_{\rI,\rh}\left[\p_{(\rx,\ry)\sim\cD}\left[\rh(\rx)\not=\ry,\avg(\erm(S_{\rI,\sqcup}; S_{\rI,\sqcap}\sqcup T))(\rx,\ry) \geq \frac{10}{243} \right]\right]\geq \ls_{\cD}^{\frac{10}{243}} (\erm(S;T)).
\end{align*}
As $\rh\in \bigsqcup_{j\in \{1,\ldots,27  \} \backslash \rI}\erm(S_{j,\sqcup}; S_{j,\sqcap} \sqcup T)$,  it follows that 
\begin{align}\label{eq:relationerror2}
&\e_{\rI,\rh}\left[\p_{(\rx,\ry)\sim\cD}\left[\rh(\rx)\not=\ry,\avg(\erm(S_{\rI,\sqcup}; S_{\rI,\sqcap}\sqcup T))(\rx,\ry) \geq \frac{10}{243} \right]\right] \\
&\qquad \leq \e_{\rI}\left[\max_{h'\in  \bigsqcup_{j\in \{1,\ldots,27  \} \backslash \rI}\erm(S_{j,\sqcup}; S_{j,\sqcap} \sqcup T) }\p_{(\rx,\ry)\sim\cD}\left[h'(\rx)\not=\ry,\avg(\erm(S_{\rI,\sqcup}; S_{\rI,\sqcap}\sqcup T))(\rx,\ry) \geq \frac{10}{243}  \right]\right]. \nonumber
\end{align}
And as $\rI\in\{1,\ldots,27  \}$, then clearly
\begin{align}\label{eq:relationerror3}
& \e_{\rI}\left[\max_{h'\in  \sqcup_{j\in \{1,\ldots,27  \} \backslash \rI}\erm(S_{j,\sqcup}; S_{j,\sqcap} \sqcup T) }\p_{(\rx,\ry)\sim\cD}\left[h'(\rx)\not=\ry,\avg(\erm(S_{\rI,\sqcup}; S_{\rI,\sqcap}\sqcup T))(\rx,\ry) \geq \frac{10}{243}  \right]\right] \nonumber
\\
&\qquad \leq\max_{i\in\{1,\ldots,27  \} } \max_{h'\in  \sqcup_{j\in \{1,\ldots,27  \} \backslash i}\erm(S_{j,\sqcup}; S_{j,\sqcap} \sqcup T) }\p_{(\rx,\ry)\sim\cD}\left[h'(\rx)\not=\ry,\avg(\erm(S_{i,\sqcup}; S_{i,\sqcap}\sqcup T))(\rx,\ry) \geq \frac{10}{243}\right]. 
\end{align} 
By combining \Cref{eq:relationerror1,eq:relationerror2,eq:relationerror3}, we conclude that
\begin{align*}
    \ls_{\cD}^{\frac{10}{243}} &(\erm(S;T))
    \\
    &\leq 5687\max_{i\in\{1,\ldots,27  \} } \max_{h'\in  \sqcup_{j\in \{1,\ldots,27  \} \backslash i}\erm(S_{j,\sqcup}; S_{j,\sqcap} \sqcup T) }\p_{(\rx,\ry)\sim\cD}\left[h'(\rx)\not=\ry,\avg(\erm(S_{i,\sqcup}; S_{i,\sqcap}\sqcup T))(\rx,\ry) \geq \frac{10}{243} \right] \nonumber
    \\
    &=5687 \max_{\stackrel{i,j\in \{1,\ldots,27  \}}{i\not=j} }\max_{h'\in \erm(S_{j,\sqcup};S_{j,\sqcap}\sqcup T)}\p_{(\rx,\ry)\sim\cD}\left[h'(\rx)\not=\ry,\avg(\erm(S_{i,\sqcup}; S_{i,\sqcap}\sqcup T))(\rx,\ry) \geq \frac{10}{243}\right], 
\end{align*}
which completes the proof.
\end{proof}

The second lemma employed in the proof of \Cref{thm:upperbound} is the standard uniform convergence property for VC classes. 

\begin{lemma}[\cite{Understandingmachinelearningfromtheory}, Theorem 6.8]\label{lem:fundamentalheoremoflearning}
There exists a universal constant $ C>1 $ such that for any distribution $ \cD  $ over $ \cX\times \left\{  -1,1\right\}  $ and any hypothesis class $ \cH \subseteq \{  -1,1\}^{\cX} $ with finite VC-dimension $ d $, it holds with probability at least $ 1-\delta $ over $ \rS\sim \cD^m$ that for all $ h\in \cH $:  
\begin{align*}
 \ls_{\cD}(h)\leq \ls_{\rS}(h)+\sqrt{\frac{C(d+\ln{\left(e/\delta \right)})}{m}}.
\end{align*}    
\end{lemma}

We now present the proof of \Cref{thm:upperbound}, which concludes the section. 

\begin{proofof}{\Cref{thm:upperbound}}    
We induct on $ k\geq 1 $. 
In particular, we will demonstrate that for each $ k\geq 1$ and $\rS \sim \cD^m$ with $m = 3^k$, and for any $\delta \in (0, 1)$, $T \in (\cX\times \cY)^{*}$, one has with probability at least $ 1-\delta $ over $ \rS $  that
\begin{align}\label{eq:upperbound5}
\hspace{-0.25 cm}
    \ls^{10/243}_{\cD}(\erm(\rS;T))\leq  \max\limits_{S'\in\cS(\rS;T)} \frac{12000 \cdot \Sigma_{\not=}(\hs\negmedspace,S')}{m/s_{\sqcap}}+\sqrt{\cb \frac{C\left(d+\ln{(e/\delta )}\right) \frac{12000\Sigma_{\not=}(\hs\negmedspace,S')}{m/s_{\sqcap}}}{m}} +\cc\frac{C\left(d+\ln{(e/\delta )}\right)}{m},
\end{align}
where $s_{\sqcap} = \frac{|S|}{|S_{i, \sqcap}|} = \frac{27}{1 - \nicefrac{1}{27}}$ is the previously defined constant, $C \geq 1$ is the constant from \Cref{lem:fundamentalheoremoflearning}, and $ \cb $ and $ \cc $ are the following constants:  
\begin{align*}
    &\cb=\left(5687^2\cdot4\cdot3^{6} \ln{(\ch e)}s_{\sqcap}\right)^{2}, 
    \\
    &\cc=3^{12}\ln{(\ch e )}^{2}\sqrt{\cb} \, s_{\sqcap}\,. 
\end{align*}
Note that applying $ \sqrt{ab}\leq a+b$ to \Cref{eq:upperbound5} would in fact suffice to complete the proof of \Cref{thm:upperbound}.  

Thus it remains only to justify \Cref{eq:upperbound5}. 
For any choice of $\delta \in (0, 1)$ and $T\in\left(\cX\times \cY\right)^{*} $, first observe observe that if $ k\leq 12,$ the claim follows immediately from the fact that the right hand side of \Cref{eq:upperbound5} is at least 1. (Owing to the fact that $\cc\geq 3^{12}$.)

We now proceed to the inductive step. For the sake of brevity, we will often suppress the distribution from which random variables are drawn when writing expectations and probabilities, e.g., $ \p_{\rS}$ rather than $\p_{\rS\sim \cD^{m}}$.
Now fix a choice of $T \in \left (\cX \times \cY \right)^{*} $, $\delta \in (0, 1)$, and $k > 12$. 
Let $ a_{\rS} $ equal the right-hand side of \Cref{eq:upperbound5}, i.e.,
\begin{align}\label{eq:upperbound-2}
    a_{\rS}=\max\limits_{S'\in\cS(\rS;T)} \frac{12000 \cdot \Sigma_{\not=}(\hs\negmedspace,S')}{m/s_{\sqcap}}+\sqrt{\cb \frac{C\left(d+\ln{(e/\delta )}\right) \frac{12000\Sigma_{\not=}(\hs\negmedspace,S')}{m/s_{\sqcap}}}{m}} +\cc\frac{C\left(d+\ln{(e/\delta )}\right)}{m}.  
\end{align}
Then invoking \Cref{lem:recursivelemma} and a union bound, we have that
\begin{align}\label{eq:upperbound-1}
    &\p_{\rS}\left[\ls_{\cD}^{10/243}(\erm(\rS;T))> a_{\rS}\right]
    \\ 
    &\quad \leq\p_{\rS}\left[5687\max_{\stackrel{i,j\in \{ 1,\ldots,27 \}}{i\not=j} } \max_{h\in \erm(\rS_{j,\sqcup};\rS_{j,\sqcap}\sqcup T)}\p_{\rx,\ry}\left[\avg(\erm(\rS_{i,\sqcup};\rS_{i,\sqcap}\sqcup T))(\rx,\ry)\geq\frac{10}{243},h(\rx)\not=\ry\right] >a_{\rS}\right]\nonumber
    \\
    &\quad \leq\sum_{\stackrel{i,j\in \{ 1,\ldots,27 \}}{i\not=j} }\p_{\rS}\left[5687\max_{h\in \erm(\rS_{j,\sqcup};\rS_{j,\sqcap}\sqcup T)}\p_{\rx,\ry}\left[\avg(\erm(\rS_{i,\sqcup};\rS_{i,\sqcap}\sqcup T))(\rx,\ry)\geq\frac{10}{243},h(\rx)\not=\ry\right] >a_{\rS}\right].
\end{align}
Thus it suffices to show that for $i \neq j \in [27]$,
\begin{align}\label{eq:upperbound0}
    \p_{\rS}\left[5687\max_{h\in \erm(\rS_{j,\sqcup};\rS_{j,\sqcap}\sqcup T)}\p_{\rx,\ry}\left[\avg(\erm(\rS_{i,\sqcup};\rS_{i,\sqcap}\sqcup T))(\rx,\ry)\geq\frac{10}{243},h(\rx)\not=\ry\right] >a_{\rS}\right]\leq  \frac{\delta}{26\cdot27},
\end{align}
as one can immediately apply this inequality with \Cref{eq:upperbound-1}. 
Then it remains to establish \Cref{eq:upperbound0}. As the pairs $(\rS_{1,\sqcup},\rS_{1,\sqcap}) , \ldots , (\rS_{27,\sqcup},\rS_{27,\sqcap})$ are all i.i.d., it suffices to demonstrate the inequality for, say, $ j=1 $ and $ i=2 $. To this end, fix arbitrary realizations $(S_k)_{3 \leq k \leq 27}$ of the random variables $(\rS_k)_{3 \leq k \leq 27}$; we will demonstrate the claim for any such realization.  

First note that if we happen to have realizations $ S_{2,\sqcup},S_{2,\sqcap} $ of $  \rS_{2,\sqcup},\rS_{2,\sqcap}$ such that 
\begin{align*}
 \ls_{\cD}^{10/243}(\erm(S_{2,\sqcup};S_{2,\sqcap}\sqcup T))=\p_{\rx,\ry}\left[\avg(\erm(S_{2,\sqcup};S_{2,\sqcap}\sqcup T))(\rx,\ry)\geq \frac{10}{243}\right]\leq \cc\frac{C\left(d+\ln{(e/\delta )}\right)}{5687m}\leq \frac{a_{\rS}}{5687},
\end{align*} 
then we are done by monotonicity of measures, as 
\begin{align}\label{eq:upperbound9}
5687\max_{h\in \erm(\rS_{1,\sqcup};\rS_{1,\sqcap}\sqcup T)}\p_{\rx,\ry}\left[\avg(\erm(S_{2,\sqcup};S_{2,\sqcap}\sqcup T))(\rx,\ry)\geq \frac{10}{243},h(\rx)\not=\ry\right]\leq  a_{\rS}.
\end{align}
Furthermore, consider any realization $ S_{2,\sqcap} $ of $ \rS_{2,\sqcap} $. 
We note that by $ m=|\rS|=3^{k}$ for $ k>12$, and by \Cref{alg:splitting27if}, it holds that $ |\rS_{2,\sqcup}| =3^{k-6}=m/3^{6}.$ 
Thus, we may invoke the inductive hypothesis with $\erm( \rS_{2,\sqcup}  ;  S_{2,\sqcap} \sqcup T)$ and failure parameter $\delta/\ch$ in order to conclude that with probability at least $ 1-\delta/\ch $ over $ \rS_{2,\sqcup} $,
\begin{align}\label{eq:upperbound6}
\ls_{\cD}^{10/243}(\erm(\rS_{2,\sqcup};S_{2,\sqcap}\sqcup T))
& \leq
\max\limits_{S'\in\cS(\rS_{2,\sqcup};S_{2,\sqcap}\sqcup T)} 
\frac{12000\Sigma_{\not=}(\hs\negmedspace,S')}{m/(3^6 s_{\sqcap})} \notag \\
&\qquad +\sqrt{\cb \frac{C\left(d+\ln{(\ch e/\delta  )}\right)
\frac{12000\Sigma_{\not=}(\hs\negmedspace,S')}{m/(3^6 s_{\sqcap})}}{m/3^6}}
+\cc\frac{C\left(d+\ln{\left(\ch e/\delta  \right)}\right)}{m/3^6}. 
\end{align}
Furthermore, for any $a,b,c,d>0 $, we have that 
\begin{align}\label{eq:upperbound-7}
a+\sqrt{abc}+cd
 \leq a+a\sqrt{b} +c\sqrt{b}+ cd
 =(1+\sqrt{b})\cdot a+(d+\sqrt{b})\cdot c,
\end{align}
where the inequality follows from the fact that $\sqrt{abc}\leq \max(\sqrt{ba^2},\sqrt{bc^2})\leq a\sqrt{b}+c\sqrt{b}$.
Now, combining \cref{eq:upperbound6} and \cref{eq:upperbound-7} (with $ b=\cb,d=\cc $), we obtain
\begin{align}\label{eq:upperbound10}
    \ls_{\cD}^{10/243}(\erm(\rS_{2,\sqcup};S_{2,\sqcap}\sqcup T)) & \leq  (1+\sqrt{\cb})\max\limits_{S'\in\cS(\rS_{2,\sqcup};S_{2,\sqcap}\sqcup T)} 
    \frac{12000\Sigma_{\not=}(\hs\negmedspace,S')}{m/(3^6 s_{\sqcap})} \notag \\ 
    & \qquad +(\cc+\sqrt{\cb}) \frac{C\left(d+\ln{(\ch e/\delta  )}\right)}{m/3^6} \notag
    \\
    & \leq2\sqrt{\cb}\max\limits_{S'\in\cS(\rS_{2,\sqcup};S_{2,\sqcap}\sqcup T)} 
    \frac{12000\Sigma_{\not=}(\hs\negmedspace,S')}{m/(3^6 s_{\sqcap})}+2\cc \frac{C\left(d+\ln{(\ch e/\delta  )}\right)}{m/3^6}.
\end{align}
Note that the second inequality makes use of the fact that $ \cb\geq1 $ and $ \cb\leq \cc.$  
We thus conclude that for any realization $ S_{2,\sqcap} $ of $ \rS_{2,\sqcap} $, the above inequality holds with probability at least $1-\delta/\ch$ over $ \rS_{2,\sqcup} $. Further, as $\rS_{2,\sqcap} $ and $ \rS_{2,\sqcup} $ are independent, the inequality also holds with probability at least $ 1-\delta/\ch $ over $ \rS_{2,\sqcap},\rS_{2,\sqcup}$.

We now let 
\[ a_{\rS_{2}}=2\sqrt{\cb}\max\limits_{S'\in\cS(\rS_{2,\sqcup};\rS_{2,\sqcap}\sqcup T)} 
\frac{12000\Sigma_{\not=}(\hs\negmedspace,S')}{m/(3^6 s_{\sqcap})}+2\cc \frac{C\left(d+\ln{(\ch e/\delta  )}\right)}{(m/3^6)} \]
and consider the following three events over $ \rS_{2}=(\rS_{2,\sqcap},\rS_{2,\sqcup} )$: 
\begin{align*}
 E_{1}=&\Bigg\{\cc\frac{C\left(d+\ln{(e/\delta )}\right)}{5687m}<    \ls_{\cD}^{10/243}(\erm(S_{2,\sqcup};S_{2,\sqcap}\sqcup T))\leq a_{\rS_{2}} \Bigg\},
 \\
 E_{2}=&\left\{ \cc\frac{C\left(d+\ln{(e/\delta )}\right)}{5687m}\geq   \ls_{\cD}^{10/243}(\erm(S_{2,\sqcup};S_{2,\sqcap}\sqcup T)) \right\},
 \\
 E_{3}=& \left\{  \ls_{\cD}^{10/243}(\erm(S_{2,\sqcup};S_{2,\sqcap}\sqcup T))
 > a_{\rS_{2}} \right\}.
\end{align*}
By \cref{eq:upperbound9}, we have that for $\rS_{2,\sqcap},\rS_{2,\sqcup} \in E_{2} $, the bound in \cref{eq:upperbound5} holds. 
Furthermore, from the comment below \cref{eq:upperbound10}, we have that $ \rS_{2,\sqcap},\rS_{2,\sqcup} \in E_{3} $ happens with probability at most $ \delta/\ch $ over $ \rS_{2,\sqcap},\rS_{2,\sqcup} $. For brevity, let $ a_{\rS} $ denote the right-hand side of \cref{eq:upperbound5}. Then, using the law of total probability along with independence of  $ \rS_{1}$ and  $\rS_{2} $, we can conclude that 
\begin{align}\label{eq:upperbound16}
    &\p_{\rS_{1},\rS_{2}}\left[5687\negmedspace\negmedspace\negmedspace\negmedspace\negmedspace\negmedspace\max_{h\in \erm(\rS_{1,\sqcup};\rS_{1,\sqcap}\sqcup T)}\p_{\rx,\ry}\left[\avg(\erm(\rS_{2,\sqcup};\rS_{2,\sqcap}\sqcup T))(\rx,\ry)\geq \frac{10}{243},h(\rx)\not=\ry\right] >a_{\rS}\right]
    \\
    &\leq     
    \e_{\rS_{2}}\left[\p_{\rS_{1}}\left[5687\negmedspace\negmedspace\negmedspace\negmedspace\negmedspace\negmedspace\max_{h\in \erm(\rS_{1,\sqcup};\rS_{1,\sqcap}\sqcup T)}\p_{\rx,\ry}\left[\avg(\erm(\rS_{2,\sqcup};\rS_{2,\sqcap}\sqcup T))(\rx,\ry)\geq \frac{10}{243},h(\rx)\not=\ry\right] >a_{\rS}\right]\Bigg| E_{1}\right]\p\left[E_{1}\right]\nonumber
    \\
    &+
    \e_{\rS_{2}}\left[\p_{\rS_{1}}\left[5687\negmedspace\negmedspace\negmedspace\negmedspace\negmedspace\negmedspace\max_{h\in \erm(\rS_{1,\sqcup};\rS_{1,\sqcap}\sqcup T)}\p_{\rx,\ry}\left[\avg(\erm(\rS_{2,\sqcup};\rS_{2,\sqcap}\sqcup T))(\rx,\ry)\geq \frac{10}{243},h(\rx)\not=\ry\right] >a_{\rS}\right]\Bigg| E_{2}\right]\p\left[E_{2}\right]\nonumber
    \\
    &+
    \e_{\rS_{2}}\left[\p_{\rS_{1}}\left[5687\negmedspace\negmedspace\negmedspace\negmedspace\negmedspace\negmedspace\max_{h\in \erm(\rS_{1,\sqcup};\rS_{1,\sqcap}\sqcup T)}\p_{\rx,\ry}\left[\avg(\erm(\rS_{2,\sqcup};\rS_{2,\sqcap}\sqcup T))(\rx,\ry)\geq \frac{10}{243},h(\rx)\not=\ry\right] >a_{\rS}\right]\Bigg| E_{3} \right]\p\left[E_{3}\right]\nonumber
    \\
    &\leq\e_{\rS_{2}}\left[\p_{\rS_{1}}\left[5687\negmedspace\negmedspace\negmedspace\negmedspace\negmedspace\negmedspace\max_{h\in \erm(\rS_{1,\sqcup};\rS_{1,\sqcap}\sqcup T)}\p_{\rx,\ry}\left[\avg(\erm(\rS_{2,\sqcup};\rS_{2,\sqcap}\sqcup T))(\rx,\ry)\geq \frac{10}{243},h(\rx)\not=\ry\right] >a_{\rS}\right]\Bigg| E_{1}\right]
    +0 
    +\delta/\ch .\nonumber
\end{align} 
Note that the second inequality follows from \cref{eq:upperbound9}, \cref{eq:upperbound10} and $ \p\left[E_{1}\right] \leq 1$.
Thus, if we can bound the first term of the final line by $ 2\delta/\ch $, it will follow that \cref{eq:upperbound0} holds with probability at least $ 1-\delta/(26\cdot27) $, as claimed.

To this end, consider a realization $ S_{2} $ of $ \rS_{2}\in E_{1} $.
For such an $ S_{2} $, we have that 
\[ \ls_{\cD}^{10/243}(\erm(S_{2,\sqcup};S_{2,\sqcap}\sqcup T))= \p_{\rx,\ry}\left[\avg(\erm(S_{2,\sqcup};S_{2,\sqcap}\sqcup T))(\rx,\ry)\geq \frac{10}{243}\right]>0. \] 
Then, again invoking the law of total probability, we have that 
\begin{align}\label{eq:upperbound1}
    &5687\max_{h\in \erm(\rS_{1,\sqcup};\rS_{1,\sqcap}\sqcup T)}\p_{\rx,\ry}\left[ \avg(\erm(S_{2,\sqcup};S_{2,\sqcap}\sqcup T))(\rx,\ry)\geq \frac{10}{243}, h(\rx)\not=\ry \right] \nonumber
    \\
 & \qquad = \;
 5687\max_{h\in \erm(\rS_{1,\sqcup};\rS_{1,\sqcap}\sqcup T)}\p_{\rx,\ry}\left[h(\rx)\not=\ry|\avg(\erm(S_{2,\sqcup};S_{2,\sqcap}\sqcup T))(\rx,\ry)\geq \frac{10}{243}\right] \nonumber \\
 & \qquad \qquad \times 
 \ls_{\cD}^{10/243}(\erm(S_{2,\sqcup};S_{2,\sqcap}\sqcup T)). 
\end{align}
Now let $ A = \{(x,y)\in(\cX\times\cY) \mid \avg(\erm(S_{2,\sqcup};S_{2,\sqcap}\sqcup T))(x,y)\geq \frac{10}{243}\}$ and  $ \rN_{1}=|\rS_{1,\sqcap}\sqcap A|$.    
As $ S_{2} \in E_{1} $, we have that 
\[ \p_{\rx,\ry}[A]=\ls_{\cD}^{10/243} \big( \erm(S_{2,\sqcup};S_{2,\sqcap}\sqcup T) \big) \geq \frac{\cc C(d+\ln{(e/\delta )})}{5687 \cdot m}. \]
Then, owing to the fact that $ \rS_{1,\sqcap}\sim \cD^{(m/s_{\sqcap})}$ --- note that $ m/s_{\sqcap}=|\rS|/(|\rS|/|\rS_{\sqcap}|)=|\rS_{\sqcap}| $ --- this implies that 
\begin{align*}
    \e_{\rS_{1,\sqcap}}[\rN_{1}]
    =\p_{\rx,\ry\sim \cD}[A] \cdot (m/s_{\sqcap})
    = \frac{\cc C\big(d+\ln{(e/\delta )}\big)}{5687 \cdot s_{\sqcap}}.
\end{align*} 
Thus, by a multiplicative Chernoff bound, we have
\begin{align}\label{eq:upperbound13}
\p_{\rS_{1,\sqcap}\sim \cD^{(m/s_{\sqcap})}}\left[\rN_{1}>\e_{\rS_{1,\sqcap}\sim \cD^{(m/s_{\sqcap})}}[\rN_{1}]/2\right] &\geq 1- \exp{\left(-\frac{\cc C(d+\ln{(e/\delta )})}{8\cdot 5687 s_{\sqcap}\cdot}  \right)} \notag \\
&\geq 1-(\delta/e)^{19}. 
\end{align} 
Note that the second inequality uses the fact that $ \cc= 3^{12} \ln{(\ch e )}^{2}\sqrt{\cb} s_{\sqcap}$.
Now let $ N_{1} $ be any realization of $\rN_1$ such that 
\[ N_1 > \frac 12 \cdot  \e_{\rS_{1,\sqcap}\sim \cD^{(m/s_{\sqcap})}}[\rN_{1}] = \frac 12 \cdot  \ls_{\cD}^{10/243} \big( \erm(S_{2,\sqcup};S_{2,\sqcap}\sqcup T) \big) (m/s_{\sqcap}). \]  
Notice that $ \rS_{1,\sqcap} \sqcap A \sim \cD^{N_{1}}(\; \cdot \; | \avg(\erm(S_{2,\sqcup};S_{2,\sqcap}\sqcup T))(\rx,\ry)\geq \frac{10}{243})$. 

Now, combining \cref{eq:upperbound1} and \cref{lem:fundamentalheoremoflearning}, we have that with probability at least $ 1-\delta/\cj $ over $ \rS_{1,\sqcap}\sqcap A$,
\begin{align}\label{eq:upperbound3}
  &5687\max_{h\in \erm(\rS_{1,\sqcup};\rS_{1,\sqcap}\sqcup T)}\p_{\rx,\ry} \left[ \avg(\erm(S_{2,\sqcup};S_{2,\sqcap}\sqcup T))(\rx,\ry)\geq \frac{10}{243}, h(\rx)\not=\ry \right] 
    \\
  &\leq
    5687\max_{h\in \erm(\rS_{1,\sqcup};\rS_{1,\sqcap}\sqcup T)}\left(\ls_{ \rS_{1,\sqcap}\sqcap A}(h)+\sqrt{C\left(d+\ln{(\ch e/\delta )}\right)/N_1}\right) \ls_{\cD}^{10/243}(\erm(S_{2,\sqcup};S_{2,\sqcap}\sqcup T))\nonumber
\end{align}
We now bound each of the two terms on the right hand side of \cref{eq:upperbound3}, considered after multiplying out the term associated with $\ls_{\cD}^{10/243}$. Beginning with the first term, and recalling that $ N_{1}=|\rS_{1,\sqcap}\sqcap A|$, we have  
\begin{align}\label{eq:upperbound17}
\ls_{ \rS_{1,\sqcap}\sqcap A}(h) & \cdot \ls_{\cD}^{10/243}(\erm(S_{2,\sqcup};S_{2,\sqcap}\sqcup T)) \nonumber \\
    &=\max_{h\in \erm(\rS_{1,\sqcup};\rS_{1,\sqcap}\sqcup T)}\sum_{(x,y)\in \rS_{1,\sqcap}\sqcap A}\frac{\ind\{ h(x)\not=y \}}{N_{1}}  \ls_{\cD}^{10/243}(\erm(S_{2,\sqcup};S_{2,\sqcap}\sqcup T))\nonumber
    \\
    &\leq
    \max_{h\in \erm(\rS_{1,\sqcup};\rS_{1,\sqcap}\sqcup T)}\sum_{(x,y)\in \rS_{1,\sqcap}\sqcap A}\frac{2\ind\{ h(x)\not=y \}}{m/s_{\sqcap}} \nonumber \\
    &\leq
    \max\limits_{S'\in \cS(\rS_{1,\sqcup};\rS_{1,\sqcap}\sqcup T)}\sum_{(x,y)\in \rS_{1,\sqcap}\sqcap A}\frac{2\ind\{ \erm(S')(x)\not=y \}}{m/s_{\sqcap}} \nonumber
    \\
    &\leq
     \max\limits_{S'\in \cS(\rS_{1,\sqcup};\rS_{1,\sqcap}\sqcup T)}\sum\limits_{(x,y)\in S'}\frac{2\ind\{ \erm(S')(x)\not=y \}}{m/s_{\sqcap}}  \nonumber \\
    &\leq
    \max\limits_{S'\in \cS(\rS_{1,\sqcup};\rS_{1,\sqcap}\sqcup T)}\frac{2\Sigma_{\not=}(\hs\negmedspace,S')}{ m/s_{\sqcap}}.
\end{align}
Note that the first inequality uses the fact that $ N_1\geq \ls_{\cD}^{10/243}(\erm(S_{2,\sqcup};S_{2,\sqcap}\sqcup T))(m/s_{\sqcap})/2$ and the second inequality uses that $h\in\erm(\rS_{1,\sqcup};\rS_{1,\sqcap}\sqcup T)$, meaning there exists an $ S'\in\cS(\rS_{1,\sqcup};\rS_{1,\sqcap}\sqcup T)$ such that $h=\erm(S')$. The third inequality follows from the fact that $\rS_{1,\sqcap}\sqcap A \sqsubset \tilde{S} $ for any $ \tilde{S}\in \cS(\rS_{1,\sqcup};\rS_{1,\sqcap}\sqcup T)$ (and especially for $S'$) and the final inequality from both the $ \cerm $-property of $ \cA $  on $ S' $ and the definition of $ \Sigma_{\not=}(\hs\negmedspace,S').$

We now bound the second term of \cref{eq:upperbound3}.
In what follows, let $ \beta= C\left(d+\ln{(\ch e/\delta )}\right)$. 
We will in the first inequality use that  $N_1 \geq \frac 12 \cdot  \ls_{\cD}^{10/243}(\erm(S_{2,\sqcup};S_{2,\sqcap}\sqcup T))(m/s_{\sqcap})$: 
\begin{align}\label{eq:upperbound18}
 \ls_{\cD}^{10/243} & (\erm(S_{2,\sqcup};S_{2,\sqcap}\sqcup T)) \sqrt{C\left(d+ \ln{(\ch e/\delta )}\right)/N_1}  \\
    &\leq\sqrt{\frac{2\beta\ls_{\cD}^{10/243}(\erm(S_{2,\sqcup};S_{2,\sqcap}\sqcup T))}{m/s_{\sqcap}}} \nonumber
    \\
    &\leq
\sqrt{\frac{4\beta   \left(\sqrt{\cb}\max\limits_{S'\in\cS(S_{2,\sqcup};S_{2,\sqcap}\sqcup T)} 
\frac{12000\Sigma_{\not=}(\hs\negmedspace,S')}{m/3^6 s_{\sqcap}}+\cc \frac{C\left(d+\ln{(\ch e/\delta  )}\right)}{m/3^6}  \right)}{m/s_{\sqcap}}}\tag{by definition of $ E_{1} $ and \cref{eq:upperbound10}}
    \\
    &\leq
\sqrt{\frac{4\beta\sqrt{\cb}\max\limits_{S'\in\cS(S_{2,\sqcup};S_{2,\sqcap}\sqcup T)} 
\frac{12000\Sigma_{\not=}(\hs\negmedspace,S')}{m/(3^6 s_{\sqcap})}}{m/s_{\sqcap}}}\tag{by $ \sqrt{a+b}\leq \sqrt{a}+\sqrt{b} $ and definition of $ \beta $ }
    + 
\sqrt{\frac{4\beta^{2}\cc \frac{1}{m/3^6}}{m/s_{\sqcap}}}\nonumber
    \\
    &\leq
   \sqrt{4\cdot 3^{6}\sqrt{\cb}\ln{\left(\ch e \right)} s_{\cap}}
   \sqrt{\frac{C\left(d+\ln{(e/\delta )}\right) \max\limits_{S'\in\cS(S_{2,\sqcup};S_{2,\sqcap}\sqcup T)}\frac{12000\Sigma_{\not=}(\hs\negmedspace,S')}{m/ s_{\sqcap}} }{m}} \nonumber \\
   & \qquad +
   \sqrt{4\cdot3^{6}\cc \left(\ln{\left(\ch e \right)}\right)^{2}s_{\cap}}
    \frac{C\left(d+\ln{(e/\delta )}\right) }{m}, \nonumber 
\end{align}
where the last inequality follows from $ \beta=C(d+\ln{(\ch e/\delta )})\leq \ln{(\ch e)}C(d+\ln{(e/\delta )}) $ and rearrangement.
We now bound each of the constant terms under the square roots. 
Beginning with the first term, we have 
\begin{align}\label{eq:upperbound11}
    \sqrt{4\cdot 3^{6}\sqrt{\cb}\ln{\left(\ch e \right)} s_{\cap}}= \sqrt{5687^2 \cdot4\cdot 3^{6}\sqrt{\cb}\ln{\left(\ch e \right)} s_{\cap}}/5687
    \leq
    \sqrt{\cb}/5687,
\end{align}
where the inequality follows from $\cb=\left(5687^2\cdot4\cdot3^{6} \ln{(\ch e)}s_{\sqcap}\right)^{2}.$ 
For the second term, we have that
\begin{align}\label{eq:upperbound12}
    \sqrt{4\cdot3^{6}\cc \left(\ln{\left(\ch e \right)}\right)^{2}s_{\cap}}= \sqrt{5687^2\cdot4\cdot3^{6}\cc \left(\ln{\left(\ch e \right)}\right)^{2}s_{\cap}}/5687
    \leq
    \cc ,
\end{align}
where the inequality follows by $\cc=  3^{12}  \ln{(\ch e )}^{2}\sqrt{\cb} s_{\sqcap}
    $ and $\cb=\left(5687^2\cdot4\cdot3^{6} \ln{(\ch e)}s_{\sqcap}\right)^{2}.$ 
Then we conclude from \cref{eq:upperbound18} that 
    \begin{align}\label{eq:upperbound19}
        &\sqrt{\frac{C\left(d+\ln{(\ch e/\delta )}\right)}{N_1}} \ls_{\cD}^{10/243}(\erm(S_{2,\sqcup};S_{2,\sqcap}\sqcup T))
        \\
        &\; \; \leq
       \frac{1}{5687}\left(\sqrt{\cb\frac{C\left(d+\ln{(e/\delta )}\right) \max\limits_{S'\in\cS(S_{2,\sqcup};S_{2,\sqcap}\sqcup T)}\frac{12000\Sigma_{\not=}(\hs\negmedspace,S')}{m/ s_{\sqcap}} }{m}} \nonumber
 + 
        \cc\frac{C\left(d+\ln{(e/\delta )}\right) }{m}\right),
    \end{align}

Thus, by applying \cref{eq:upperbound17} and \cref{eq:upperbound19} to \cref{eq:upperbound3}, we obtain that for any realization $ N_{1} $ of $ \rN_1\geq \frac 12 \cdot \ls_{\cD}^{10/243}(\erm(S_{2,\sqcup};S_{2,\sqcap}\sqcup T))(m/s_{\sqcap})$, it holds with probability at least $1-\delta/\cj $ over $ \rS_{1,\sqcap}\sqcap A \sim \cD^{N_{1}}(\; \cdot \;| \erm(S_{2,\sqcup};S_{2,\sqcap}\sqcup T)(x)\not=y)$\footnote{I.e., the restriction of $\CD$ to those $(x, y)$ pairs satisfying the given condition.} that
\begin{align}\label{eq:upperbound15}
&5687\max_{h\in \erm(\rS_{1,\sqcup};\rS_{1,\sqcap}\sqcup T)}\p_{\rx,\ry} \left[ \avg(\erm(S_{2,\sqcup};S_{2,\sqcap}\sqcup T))(\rx,\ry)\geq \frac{10}{243}, h(\rx)\not=\ry \right] 
\\
& \quad \leq \max\limits_{S'\in \cS(\rS_{1,\sqcup};\rS_{1,\sqcap}\sqcup T)}\frac{12000\Sigma_{\not=}(\hs\negmedspace,S')}{m/s_{\sqcap}} +\sqrt{\cb\frac{C\left(d+\ln{(e/\delta )}\right) \max\limits_{S'\in\cS(S_{2,\sqcup};S_{2,\sqcap}\sqcup T)}\frac{12000\Sigma_{\not=}(\hs\negmedspace,S')}{m/ s_{\sqcap}} }{m}} \nonumber \\
& \qquad + \cc\frac{C\left(d+\ln{(e/\delta )}\right) }{m} \nonumber \\
& \quad \leq \max\limits_{S'\in \cS(\rS; T)}\frac{12000\Sigma_{\not=}(\hs\negmedspace,S')}{m/s_{\sqcap}} +\sqrt{\cb\frac{C\left(d+\ln{(e/\delta )}\right) \frac{12000\Sigma_{\not=}(\hs\negmedspace,S')}{m/ s_{\sqcap}} }{m}} \nonumber
+ \cc\frac{C\left(d+\ln{(e/\delta )}\right) }{m} \nonumber \\
&\quad =a_{\rS}.
\end{align} 
Note that the second inequality follows from the fact that $ S'\in \cS(\rS_{i,\sqcup};\rS_{i,\sqcap}\sqcup T) $ for $ i=1,2 $, meaning $ S'\in \cS(\rS;T)$. The equality follows simply from the definition of $ a_{\rS}$ in \cref{eq:upperbound-2}.  

Now, combining the above observations, we can conclude that for any realization $S_{2} \in E_{1}$ of $\rS_{2}$,
\begingroup
\allowdisplaybreaks
\begin{align*}
&\p_{\rS_{1}}\Bigg[5687\max_{h\in \erm(\rS_{1,\sqcup};\rS_{1,\sqcap}\sqcup T)}\p_{(\rx,\ry)}\left[\avg(\erm(S_{2,\sqcup};S_{2,\sqcap}\sqcup T))(\rx,\ry)\geq \frac{10}{243},h(\rx)\not=\ry\right] \leq a_{\rS} \Bigg]
\\
& \quad \geq\p_{\rS_{1}}\Bigg[5687\max_{h\in \erm(\rS_{1,\sqcup};\rS_{1,\sqcap}\sqcup T)}\p_{(\rx,\ry)}\left[\avg(\erm(S_{2,\sqcup};S_{2,\sqcap}\sqcup T))(\rx,\ry)\geq \frac{10}{243},h(\rx)\not=\ry\right] \leq
a_{\rS} \\ 
&\qquad \Bigg| \rN_1\geq \ls_{\cD}^{10/243}(\erm(S_{2,\sqcup};S_{2,\sqcap}\sqcup T))(m/s_{\sqcap})/2 \Bigg]  \times \p_{\rS_{1}}\left[\rN_1\geq \ls_{\cD}^{10/243}(\erm(S_{2,\sqcup};S_{2,\sqcap}\sqcup T))(m/s_{\sqcap})/2\right] 
\\
&\geq  \left( 1- \frac{\delta}{\ch} \right) \left( 1- \left(\frac{\delta}{e} \right)^{19} \right) \\ 
&\geq 1- 2\frac{\delta}{2^{12}}.
\end{align*} 
\endgroup
(Note that the first term on the right side of the first inequality is distributed across two lines of text.) In particular, 
 the first inequality follows from the law of total expectation, and the second from \cref{eq:upperbound13} (see the comment below the equations) and \cref{eq:upperbound15}.
Thus, the above implies that the term in \cref{eq:upperbound16} which conditions upon $\rS_{2}\in E_{1}$ is bounded by $ \delta/2^{11}$. Altogether, we can conclude that 
\begin{align*}
    \p_{\rS_{1},\rS_{2}}\left[5687\max_{h\in \erm(\rS_{1,\sqcup};\rS_{1,\sqcap}\sqcup T)}\p_{(\rx,\ry)}\left[\avg(\erm(\rS_{2,\sqcup};\rS_{2,\sqcap}\sqcup T))(\rx,\ry)\geq \frac{10}{243},h(\rx)\not=\ry\right] >a_{\rS}\right]
    \leq \frac{\delta}{2^{10}}. 
\end{align*} 
As $ \nicefrac{\delta}{2^{10}} \leq \nicefrac{\delta}{26\cdot 27}$, we arrive at \cref{eq:upperbound0}, which as previously argued concludes the proof due to the fact that $(\rS_{1,\sqcup},\rS_{1,\sqcap}), \ldots, (\rS_{27,\sqcup},\rS_{27,\sqcap})$ are i.i.d.\ 
\end{proofof}

\section{Augmentation of $\widehat{\cA}_{\rt}$ Through Tie-breaking}

Let us assume we are given a training sequence $ \rS $ of size $ m=3^{k} $ for $ k\geq1 $.  
We then take $ \rS $ and split it into three disjoint, equal-sized training sequences $ \rS_{1},$ $ \rS_{2},$ and $ \rS_{3}.$ 
We denote the sizes of $ \rS_{1},\rS_{2}$ and $\rS_{3} $ as  $ m'=m/3 $. 
On $ \rS_{1}$ and $ \rS_{2}$, we train  $\widehat{\cA}_{\rt_{1}}(S_{1})$ and $\widehat{\cA}_{\rt_{2}}(\rS_{2}),$ where we recall that $ \rt_{1} $ and $ \rt_{2} $ denote the randomness used to draw the hypothesis in  $\widehat{\cA}_{\rt_{1}}(S_{1})$ and $\widehat{\cA}_{\rt_{2}}(\rS_{2})$ from, respectively, $ \cA(\rS_{1};\emptyset) $  and $ \cA(\rS_{2};\emptyset).$

We now evaluate $\widehat{\cA}_{\rt_{1}}(\rS_{1})$ and $\widehat{\cA}_{\rt_{2}}(\rS_{2})$ on $ \rS_{3} $ and consider all the examples $ (x,y)\in \rS_{3} $,  where $ \avg(\widehat{\cA}_{t_{1}}(S_{1}))(x,y)\geq11/243 $ or $\avg(\widehat{\cA}_{t_{2}}(S_{2}))(x,y)\geq11/243$. Denote the set of all such examples as $\rS_{3}^{\not=}.$ 
We now run the $ \cerm $-algorithm $ \cA $ on  $ \rS_{3}^{\not=} $ to obtain $\htie =\cA(\rS_{3}^{\not=}).$

For a point $ x $,  let $ \tie^{11/243}\left(\widehat{\cA}_{\rt_{1}}(\rS_{1}),\widehat{\cA}_{\rt_{2}}(\rS_{2}); \htie \right)(x) $ be equal to the label $ y $ if both
\[ \sum_{h\in \widehat{\cA}_{\rt_{1}}(\rS_{1})}\ind\{h(x)=y\}/|\widehat{\cA}_{\rt_{1}}(\rS_{1})|\geq 232/243 \] and 
\[ \sum_{h\in \widehat{\cA}_{\rt_{2}}(\rS_{2})}\ind\{h(x)=y\}/|\widehat{\cA}_{\rt_{2}}(\rS_{2})|\geq 232/243. \]
Otherwise, we set it to $ \htie(x).$
In other words, if both $ \widehat{\cA}_{\rt_{1}}(\rS_{1}) $ and $ \widehat{\cA}_{\rt_{2}}(\rS_{2}) $ have at least $ 232/243 $ of their hypotheses agreeing on the same label $ y $, we output that label; otherwise, we output the label of $ \htie (x).$ 

Notice that if there were a true label $ y $ and point $ x $, such that we ended up outputting the answer of $\htie(x),$ then at least one of $ \widehat{\cA}_{\rt_{1}}(\rS_{1}) $ and $ \widehat{\cA}_{\rt_{2}}(\rS_{2}) $ has more than $ 11/243 $ incorrect answers, not equal to $ y $ on $ x $, which we know by \cref{lem:upperbound} is unlikely. 
Furthermore, in the former case and the tie erring on $(x, y)$, then both $ \widehat{\cA}_{\rt_{1}}(\rS_{1}) $ and $ \widehat{\cA}_{\rt_{2}}(\rS_{2}) $ err with at least a $ 232/243 $ fraction of their hypotheses, which again is unlikely by \cref{lem:upperbound}. Thus, both cases of possible error are unlikely, which we exploit in order to demonstrate the following theorem.         

\begin{theorem}\label{thm:tiemajority}
    There exists a universal constant $ c \geq1$ such that for any hypothesis class $ \cH $ of VC dimension $ d $, distribution $ \cD $ over $ \cX \times \cY,$ failure parameter $ 0<\delta<1 $, training sequence size $ m=3^{k} $ for $ k\geq 5 $, training sequence $ \rS\sim \cD^{m},$ and sampling size $ t_{1},t_{2}\geq 4\cdot243^{2}\ln{\big(2m/(\delta(d+\ln{\left(\cur/\delta \right)})) \big)},$ we have, with probability at least $ 1-\delta $ over $ \rS_{1},\rS_{2},\rS_{3}\sim \cD^{m/3} $ and the randomness $ \rt_{1},\rt_{2} $ used to draw $ \widehat{\erm}_{\rt_{1}}(\rS_{1};\emptyset) $ and $ \widehat{\erm}_{\rt_{2}}(\rS_{2};\emptyset) $,  that:
    \begin{align*}
    \ls_{\cD}\left(\tie^{11/243}\left(\widehat{\cA}_{\rt_{1}}(\rS_{1}),\widehat{\cA}_{\rt_{2}}(\rS_{2}), \htie\right)\right)\leq  2.0888\tau+\sqrt{\frac{c\tau \left(d+\ln{(e/\delta )}\right)}{m}}+\frac{c\left(d+\ln{(e/\delta )}\right)}{m}. 
    \end{align*}
\end{theorem}

In the proof of \cref{thm:tiemajority} we will need the following $ \cerm $-theorem. Recall that we take $ \cA $  to be an $ \cerm $-algorithm, meaning $ \cA $  is proper (i.e., it always emits hypotheses in $\cH$), and for any training sequence $S$ that $ \ls_{S}(\cA(S)) = \min_{h\in\cH}\ls_{S}(h).$    

\begin{theorem}\label{thm:ermtheoremunderstanding}[\cite{Understandingmachinelearningfromtheory}~Theorem 6.8]
        There exists a universal constant $ C'>1 $ such that for any distribution $ \cD  $ over $ \cX\times \left\{  -1,1\right\}  $, any hypothesis class $ \cH \subseteq \{  -1,1\}^{\cX} $ with VC dimension $ d $, and any $ \cerm $-algorithm $ \cA $,   it holds with probability at least $ 1-\delta $ over $ \rS\sim \cD^m$ that for all $ h\in \cH $:  
        \begin{align*}
         \ls_{\cD}(\cA(\rS))\leq \inf_{h\in \cH}\ls_{\cD}(h)+\sqrt{\frac{C'(d+\ln{\left(e/\delta \right)})}{m}}.
        \end{align*}    
\end{theorem}

\begin{proofof}{\cref{thm:tiemajority}.}
First note that by the definition of $\tie^{11/243}$, for $ \tie^{11/243}\left(\widehat{\cA}_{\rt_{1}}(\rS_{1}),\widehat{\cA}_{\rt_{2}}(\rS_{2}), \htie\right) $ to err on a fixed example $ (x,y) $, it must be the case that either there exists $y' \neq y \in\left\{  -1,1\right\}$ such that
\begin{align*}
& \sum_{h\in \widehat{\cA}_{\rt_{1}}(\rS_{1})}\ind\{h(x)=y'\}/|\widehat{\cA}_{\rt_{1}}(\rS_{1})|\geq 232/243 \text{  and } \sum_{h\in \widehat{\cA}_{\rt_{2}}(\rS_{2})}\ind\{h(x)=y'\}/|\widehat{\cA}_{\rt_{2}}(\rS_{2})|\geq 232/243
 \\
& \Leftrightarrow \avg(\widehat{\cA}_{\rt_{1}}(\rS_{1}))(x,y)\geq \frac{232}{243},\avg(\widehat{\cA}_{\rt_{2}}(\rS_{2}))(x,y)\geq \frac{232}{243}
\end{align*} 
or the case that for all $y'\in\left\{  -1,1\right\}$,
\begin{align*}
& \Bigg( \sum_{h\in \widehat{\cA}_{\rt_{1}}(\rS_{1})}\ind\{h(x)=y'\}< 232/243 \text{ or} \sum_{h\in \widehat{\cA}_{\rt_{2}}(\rS_{2})}\ind\{h(x)=y'\}< 232/243 \Bigg)  \text{ and } \htie(x)\not=y
    \\
& \Rightarrow  \left( \avg(\widehat{\cA}_{\rt_{1}}(\rS_{1}))(x,y)\geq \frac{11}{243} \text{ or } \avg(\widehat{\cA}_{\rt_{2}}(\rS_{2}))(x,y)\geq \frac{11}{243} \right)  \text{ and } \htie(x)\not=y
\end{align*} 
Thus, we have that   
\begin{align}\label{eq:secondmajority1}
 \ls_{\cD} \Big(& \tie^{11/243}  \left(\widehat{\cA}_{\rt_{1}}(\rS_{1}),\widehat{\cA}_{\rt_{2}}(\rS_{2}), \htie\right)(\rx)\not=\ry\Big) \nonumber 
 \\
& \leq
 \p_{(\rx,\ry)\sim\cD}\left[\avg(\widehat{\cA}_{\rt_{1}}(\rS_{1}))(\rx,\ry)\geq \frac{232}{243},\avg(\widehat{\cA}_{\rt_{2}}(\rS_{2}))(\rx,\ry)\geq \frac{232}{243}\right] \nonumber
 \\
& \qquad  +\p_{(\rx,\ry)\sim\cD}\left[ \htie\not=\ry, \avg(\widehat{\cA}_{\rt_{1}}(\rS_{1}))(\rx,\ry)\geq \frac{11}{243} \text{ or } \avg(\widehat{\cA}_{\rt_{2}}(\rS_{2}))(\rx,\ry)\geq \frac{11}{243}\right]. 
\end{align}
We now bound each of these terms separately. 
The first term we will soon bound by 
\begin{align}\label{eq:secondmajority-1}
& \hspace{-1.7cm} \p_{(\rx,\ry)\sim\cD}\left[\avg(\widehat{\cA}_{\rt_{1}}(\rS_{1}))(\rx,\ry)\geq \frac{232}{243},\avg(\widehat{\cA}_{\rt_{2}}(\rS_{2}))(\rx,\ry)\geq \frac{232}{243}\right] \nonumber  \\ 
& \hspace{-1.7cm} \qquad \leq 1.0888\tau+\sqrt{\frac{16cC\tau(d+2\ln{\left(3\cu/\delta \right)})}{m'}}+\frac{6cC(d+\ln{\left(e\cu /\delta \right)})}{m'} 
\end{align}
with probability at least $1- 82\delta/\cu,$ over $ \rS_{1},\rS_{2},\rt_{1} $ and $ \rt_{2} $. Likewise, the second term we will soon bound by 
\begin{align}\label{eq:secondmajority-2}
    \p_{(\rx,\ry)\sim\cD}\left[ \htie(\rx)\not=\ry, \widehat{\cA}_{\rt_{1}}(\rS_{1})(\rx)\not=\widehat{\cA}_{\rt_{2}}(\rS_{2})(\rx)\right]\leq  
    \tau+ \sqrt{\frac{4\tau cC'(d+\ln{\left(\cu/\delta \right)})}{m'}}+\frac{5cC'\ln{\left(\cu e/\delta \right)}}{m}
\end{align}
with probability $ 1-4\delta/\cu $ at least over $ \rS_{1},\rS_{2}, \rS_{3},\rt_{1} $ and $ \rt_{2} $, where $ c,C,C'\geq 1 $ are universal constants and $ \cu= 86$. 
Applying a union bound over the above two events establishes the claim of the theorem.

Let us begin by pursuing \cref{eq:secondmajority-1}. 
To this end, consider any realizations of $ S_{1},S_{2},t_{1}$ and $ t_{2} $ of $ \rS_{1},$ $ \rS_{2}, \rt_{1}$ and $ \rt_{2}.$ 
We note that for an example $ (x,y) $ with $ \avg(\widehat{\cA}_{\rt_{1}}(\rS_{1}))(x,y)\geq 232/243$ and $\avg(\widehat{\cA}_{t_{2}}(S_{2}))(x,y)\geq 232/243$, it must be the case that both $ \widehat{\cA}_{t_{1}}(S_{1}) $ and $ \widehat{\cA}_{t_{2}}(S_{2}) $  have at least a $ 232/243 $-fraction of hypotheses which err at $(x, y)$. 
Now let $ \rh $ be a random hypothesis drawn from $ \widehat{\cA}_{t_{1}}(S_{1}).$ Then with probability at least $ 232/243,$ $ \rh(x)\not=y.$ 
Thus, for any such example $ (x,y) $, we conclude that  
\begin{align*}
\hspace{-0.6 cm}
 \p_{\rh}\left[\rh(x)\not=y,\avg(\widehat{\cA}_{t_{2}}(S_{2}))(x,y)\geq \frac{232}{243}\right]\geq \frac{232}{243}\ind\left\{ \avg(\widehat{\cA}_{t_{1}}(S_{1}))(x,y)\geq \frac{232}{243},\avg(\widehat{\cA}_{t_{2}}(S_{2}))(x,y)\geq \frac{232}{243} \right\}. 
\end{align*}

Multiplying both sides of the above equation by $ 243/232 $ and taking expectation with respect to $ (\rx,\ry)\sim\cD $ on both sides, we obtain
\begin{align}\label{eq:secondmajority2}
& \p_{(\rx,\ry)\sim\cD}\left[\avg(\widehat{\cA}_{t_{1}}(S_{1}))(\rx,\ry)\geq \frac{232}{243},\avg(\widehat{\cA}_{t_{2}}(S_{2}))(\rx,\ry)\geq \frac{232}{243}\right]   \\ 
& \qquad \leq \frac{243}{232}\e_{\rh}\left[\p_{(\rx,\ry)\sim \cD}\left[\rh(\rx)\not=\ry,\avg(\widehat{\cA}_{t_{2}}(S_{2}))(\rx,\ry)\geq \frac{232}{243}\right]\right].
\end{align}
Now by $ \rh $ being drawn from $ \widehat{\cA}_{t_1}(S_{1}),$ which in turn is drawn from $ \cA(S_{1};\emptyset),$ we conclude that $ \rh $ is contained in $ \cA(S_{1};\emptyset).$ Thus, 
\[ \e_{\rh}\left[\p_{(\rx,\ry)\sim \cD}\left[\rh(x)\not=y,\avg(\widehat{\cA}_{t_{2}}(S_{2}))(\rx,\ry)\geq 232/243 \right]\right] \] 
can be upper bounded by $ \max_{h\in \cA(S_{1};\emptyset)}\p_{(\rx,\ry)\sim \cD}\left[h(x)\not=y,\avg(\widehat{\cA}_{t_{2}}(S_{2}))(\rx,\ry)\geq 232/243\right]$. 
Using this observation and substituting it into \cref{eq:secondmajority2}, we obtain that
\begin{align*}
& \p_{(\rx,\ry)\sim\cD}\left[\avg(\widehat{\cA}_{t_{1}}(S_{1}))(\rx,\ry)\geq \frac{232}{243},\avg(\widehat{\cA}_{t_{2}}(S_{2}))(\rx,\ry)\geq \frac{232}{243}\right] \\
& \qquad \leq\frac{243}{232}\max_{h\in \cA(S_{1};\emptyset)}\p_{(\rx,\ry)\sim \cD}\left[h(\rx)\not=\ry,\avg(\widehat{\cA}_{t_{2}}(S_{2}))(\rx,\ry)\geq \frac{232}{243}\right] \\
& \qquad \leq\frac{243}{232}\max_{h\in \cA(S_{1};\emptyset)}\p_{(\rx,\ry)\sim \cD}\left[h(\rx)\not=\ry,\avg(\widehat{\cA}_{t_{2}}(S_{2}))(\rx,\ry)\geq \frac{11}{243}\right].
\end{align*}
As we demonstrated the above inequality for any realizations of $ S_{1},S_{2},t_{1}$ and $ t_{2} $ of $ \rS_{1},$ $ \rS_{2}, \rt_{1}$ and $ \rt_{2},$ the inequality also holds for the random variables. 
We now demonstrate that the right-hand side of the above expression can be bounded by
\begin{align}\label{eq:secondmajority0}
& \frac{243}{232}\max_{h\in \cA(S_{1};\emptyset)}\p_{(\rx,\ry)\sim \cD}\left[h(\rx)\not=\ry,\avg(\widehat{\cA}_{\rt_{2}}(\rS_{2}))(\rx,\ry)\geq \frac{11}{243}\right] \nonumber \\ 
& \qquad  \leq 1.0888\tau+\sqrt{\frac{16cC\tau(d+2\ln{\left(3\cu/\delta \right)})}{m'}}+\frac{6cC(d+\ln{\left(e\cu /\delta \right)})}{m'}
\end{align}
with probability at least $ 1-82\delta/\cu $ over $ \rS_{1} $, $ \rS_{2}$, and $ \rt_{2}.$ We denote  the above event  over $ \rS_{1},\rS_{2} $, and $ \rt_{2}$ as  $ E_{G}.$   
In pursuit of \cref{eq:secondmajority0}, we now consider the following events over $ \rS_{2}  $ and $ \rt_{2} $:  
\begin{align*}
E_{1} &= \left\{ \ls_{\cD}^{11/243}(\widehat{\cA}_{\rt_{2}}(\rS_{2}))\leq  \frac{c\ln{\left(\cu e/\delta \right)}}{m'} \right\}
 \\
E_{2} &= \left\{ \frac{c\ln{\left(\cu e/\delta \right)}}{m'} < \ls_{\cD}^{11/243}(\widehat{\cA}_{\rt_{2}}(\rS_{2}))\leq  c\tau+\frac{c\left(d+\ln{(\cu e/\delta )}\right)}{m'}\right\}
 \\
E_{3} &= \left\{ \ls_{\cD}^{11/243}(\widehat{\cA}_{\rt_{2}}(\rS_{2}))>  c\tau+\frac{c\left(d+\ln{(\cu e/\delta )}\right)}{m'}\right\},  
\end{align*}
where $c$ is at least the constant of \cref{thm:main} and also greater than $ c\geq 2\cdot 10^{6}s_{\sqcap}$.
We first notice that if $ S_{2} $ and $ t_{2} $   are realizations of $ \rS_{2} $ and $ \rt_{2} $ in $ E_{1},$ then by monotonicity of measures, we have that  
\begin{align}\label{eq:secondmajority9} 
    \frac{243}{232}\max_{h\in \cA(S_{1};\emptyset)}\p_{(\rx,\ry)\sim \cD}\left[h(\rx)\not=\ry,\avg(\widehat{\cA}_{t_{2}}(S_{2}))(\rx,\ry)\geq \frac{11}{243}\right]\leq \frac{2c \ln{\left(\cu e/\delta \right)}}{m'},
\end{align}
which would imply the event $ E_{G} $ in \cref{eq:secondmajority0}.  

We now consider realizations $ S_{2} $ and $ t_{2} $ of $ \rS_{2} $ and $ \rt_{2} $ in $ E_{2}.$ 
Let $ \rS_{1,i,\sqcap}$ denote  $ (\rS_{1})_{i,\sqcap} $ and  $ \rS_{1,i,\sqcup}=(\rS_{1})_{i,\sqcup}$ for $ i\in \{  1,\ldots,27\}.$
Using this notation, we have that 
\begin{align*}
& \frac{243}{232}\max_{h\in \cA(S_{1};\emptyset)}\p_{(\rx,\ry)\sim \cD}\left[h(\rx)\not=\ry,\avg(\widehat{\cA}_{t_{2}}(S_{2}))(\rx,\ry)\geq \frac{11}{243}\right]
    \\
&\quad =\frac{243}{232}\max_{i\in\{  1,\ldots,27\} } \max_{h\in \cA(\rS_{1,i,\sqcup};\rS_{1,i,\sqcap})}\p_{(\rx,\ry)\sim \cD}\left[h(\rx)\not=\ry,\avg(\widehat{\cA}_{t_{2}}(S_{2}))(\rx,\ry)\geq \frac{11}{243}\right].
\end{align*}

We now bound the $\max$ term associated to each $i \in [27]$. To this end, fix such an $i$ and let $ A=\{ (x,y)\mid \avg(\widehat{\cA}_{t_{2}}(S_{2}))(x,y)\geq 11/243\}$. Also, let $\rN_{i}$ denote the number of examples in $ \rS_{1,i,\sqcap} $ landing in $ A  $, i.e., $ \rN_{i}=| \rS_{1,i,\sqcap}\sqcap A|.$  
Now by $ S_{2} $ and $ t_{2} $ being realizations of $ \rS_{2} $ and $ \rt_{2} $ in $ E_{2}$, we have that $ \p\left[A\right]=\p_{(\rx,\ry)\sim\cD}[A]=\ls_{\cD}(\widehat{\cA}_{t_{2}}(S_{2}))\geq \frac{c\ln{\left(\cu e/\delta \right)}}{m'}.$
Thus, as $\rS_{1,i,\sqcap}\sim \cD^{m'/s_{\sqcap}}$, we have that $ \e\left[N_{i}\right]=\p[A]m'/s_{\sqcap}$. 
Furthermore, this implies by Chernoff that with probability at least $ 1-\delta/\cu $ over $ \rS_{1,i,\sqcap} $,  
\begin{align*}
 \p_{\rN_{i}}\left[\rN_{i}> (1-1/1000)\p[A]m'/s_{\sqcap}\right] \geq 1- \exp{\left(-\p[A]m'/(2\cdot 1000^{2}s_{\sqcap} )\right)} \geq 1-\delta/\cu
\end{align*}
where the last inequality follows by $ \p[A]\geq \frac{c\ln{\left(\cu e/\delta \right)}}{m'}$ and $ c\geq 2\cdot 10^{6}s_{\sqcap}.$  
Let now $ \cD(\cdot \mid A) $ be the conditional distribution of $ A $, i.e., for an event $ E $ over $ \cX\times \cY  $, we have that $ \cD(E\mid A)=\p_{(\rx,\ry)\sim \cD}\left[(\rx,\ry)\in E\cap A\right]/\p_{(\rx,\ry)\sim \cD}\left[(\rx,\ry)\in A\right].$ 
Since $ \rS_{1,i,\sqcap}\sim \cD $, it follows that $ \rS_{1}\sqcap A \sim \cD(\cdot \mid A)^{\rN_{i}}.$  
Consider now a realization $ N_{i} $ of $ \rN_{i}$ with 
\[ N_i \geq (1-1/1000)\p[A]m'/s_{\sqcap}=(1-1/1000)\ls_{\cD}(\widehat{\cA}_{t_{2}}(S_{2}))m'/s_{\sqcap}. \]   
Then by the law of total probability and definition of $\cD(\, \cdot \,| A)$, we have that 
\begin{align*}
&\max_{h\in \cA(\rS_{1,i,\sqcup};\rS_{1,i,\sqcap})}\p_{(\rx,\ry)\sim \cD}\left[h(x)\not=y,\avg(\widehat{\cA}_{t_{2}}(S_{2}))(\rx,\ry)\geq \frac{11}{243}\right] \\
&=\max_{h\in \cA(\rS_{1,i,\sqcup};\rS_{1,i,\sqcap})}\p_{(\rx,\ry)\sim \cD}\left[h(x)\not=y\mid\avg(\widehat{\cA}_{t_{2}}(S_{2}))(\rx,\ry)\geq \frac{11}{243}\right]\p_{(\rx,\ry)\sim \cD}\left[\avg(\widehat{\cA}_{t_{2}}(S_{2}))(\rx,\ry)\geq \frac{11}{243}\right] \\
&= \max_{h\in \cA(\rS_{1,i,\sqcup};\rS_{1,i,\sqcap})}\p_{(\rx,\ry)\sim \cD(\cdot\mid A)}\left[h(x)\not=y\right]\ls_{\cD}(\widehat{\cA}_{t_{2}}(S_{2})),
\end{align*}
where the last equality follows by the definition of $ \cD(\cdot | A).$ 
Furthermore, by \cref{lem:fundamentalheoremoflearning}, we have with probability at least $ 1-\delta/\cu $ over $ \rS_{1,i,\sqcap}\sqcap A\sim \cD(\cdot\mid A)^{N_{i}} $ that
\begin{align}\label{eq:secondmajority4}
& \max_{h\in \cA(\rS_{1,i,\sqcup};\rS_{1,i,\sqcap})}\p_{(\rx,\ry)\sim \cD(\cdot\mid A)}\left[h(x)\not=y\right]\ls_{\cD}(\widehat{\cA}_{t_{2}}(S_{2})) \nonumber \\
& \quad \leq \ls_{\cD}(\widehat{\cA}_{t_{2}}(S_{2}))\left(\max_{h\in \cA(\rS_{1,i,\sqcup};\rS_{1,i,\sqcap})}\ls_{\rS_{1,i,\sqcap}\sqcap A}(h)+\sqrt{\frac{C(d+\ln{\left(e\cu/\delta \right)})}{N_{i}}}\right),
\end{align}
where  $ C\geq1 $ is the universal constant of \cref{lem:fundamentalheoremoflearning}.   
We now bound each term, starting with the first. 
Now, $ \max_{h\in \cA(\rS_{1,i,\sqcup};\rS_{1,i,\sqcap})}\ls_{\rS_{1,i,\sqcap}\sqcap A}(h) $ is equal to $ \max_{h\in \cA(\rS_{1,i,\sqcup};\rS_{1,i,\sqcap})} \Sigma_{\not=}(h,\rS_{1,i,\sqcap}\sqcap A)/N_{i}.$ 
And as any $ h\in \cA(\rS_{1,i,\sqcup};\rS_{1,i,\sqcap}) $ is equal to $ h=\cA(S') $ for some $ S'\in \cS(\rS_{1,i,\sqcup};\rS_{1,i,\sqcap}),$ we get that $ \max_{h\in \cA(\rS_{1,i,\sqcup};\rS_{1,i,\sqcap})} \Sigma_{\not=}(h,\rS_{1,i,\sqcap}\sqcap A)$ is equal to $ \max_{S'\in \cS(\rS_{1,i,\sqcup};\rS_{1,i,\sqcap})} \Sigma_{\not=}(\cA(S'),\rS_{1,i,\sqcap}\sqcap A).$ 
Furthermore, since any $ S'\in \cS(\rS_{1,i,\sqcup};\rS_{1,i,\sqcap}) $ contains the training sequence $ \rS_{1,i,\sqcap} \sqcap A$, we get that 
$ \max_{S'\in \cS(\rS_{1,i,\sqcup};\rS_{1,i,\sqcap})} \Sigma_{\not=}(\cA(S'),\rS_{1,i,\sqcap}\sqcap A) \leq  \max_{S'\in \cS(\rS_{1,i,\sqcup};\rS_{1,i,\sqcap})} \Sigma_{\not=}(\cA(S'),S').$

Now we have that $ \cA(S') $ is a $ \cerm $-algorithm run on $ S',$ thus $ \cA(S') \in \cH $ and any other hypothesis in $ \cH $ has a larger empirical error on $ S' $ than $ \cA(S'), $ including $ \hs$. We thus have that $ \max_{S'\in \cS(\rS_{1,i,\sqcup};\rS_{1,i,\sqcap})} \Sigma_{\not=}(\cA(S'),S') \leq \max_{S'\in \cS(\rS_{1,i,\sqcup};\rS_{1,i,\sqcap})} \Sigma_{\not=}(\hs,S'),$ which, since $ S' $ is contained in $ \rS_{1,i}=(\rS_{1})_{i} $, can further be upper bounded by $ \Sigma_{\not=}(\hs,\rS_{1,i})$. We can thus conclude that 
\[ \max_{h\in \cA(\rS_{1,i,\sqcup};\rS_{1,i,\sqcap})}\ls_{\rS_{1,i,\sqcap}\sqcap A}(h)\leq  \Sigma_{\not=}(\hs,\rS_{1,i})/N_{i}=m'\ls_{\rS_{1,i}}(\hs)/(3^3N_{i}). \] Note we have used that $ |\rS_{1,i}|=m'/3^{3}.$  
Now, using the fact that we have crystallized an event $N_i$ with
$ N_{i}\geq  (1-1/1000)\ls_{\cD}(\widehat{\cA}_{t_{2}}(S_{2}))m'/s_{\sqcap},$ we have that the first term of \cref{eq:secondmajority4} (after factoring out the multiplication) can be bounded as follows:
\begin{align}\label{eq:secondmajority5}
\ls_{\cD}(\widehat{\cA}_{t_{2}}(S_{2}))\max_{h\in \cA(\rS_{1,i,\sqcup};\rS_{1,i,\sqcap})}\ls_{\rS_{1,i,\sqcap}\sqcap A}(h)
& \leq
\ls_{\cD}(\widehat{\cA}_{t_{2}}(S_{2}))\frac{m'}{3N_{i}}\ls_{\rS_{1,i}}(\hs) \nonumber \\ 
& \leq \frac{1000s_{\sqcap}}{3^{3}999}\ls_{\rS_{1,i}}(\hs) \nonumber 
\\
& \leq \frac{1000\cdot27}{999\cdot26}\ls_{\rS_{1,i}}(\hs)
\end{align}     
where we, in the last inequality, have used that $ s_{\sqcap}=\frac{3^{3}}{1-\frac{1}{27}}$. This concludes the bound on the first term in \cref{eq:secondmajority4}.

Now, using that $ N_{i}\geq  (1-1/1000)\ls_{\cD}(\widehat{\cA}_{t_{2}}(S_{2}))m'/s_{\sqcap},$ we get that the second term in \cref{eq:secondmajority4} can be bound as follows:
\begin{align*}
\sqrt{\frac{C(d+\ln{\left(e\cu/\delta \right)})}{N_{i}}}\ls_{\cD}(\widehat{\cA}_{t_{2}}(S_{2}))
& \leq 
\sqrt{\frac{1000C(d+\ln{\left(e\cu/\delta \right)})\ls_{\cD}(\widehat{\cA}_{t_{2}}(S_{2}))s_{\sqcap}}{999m'}}
\nonumber  \\
& \leq\sqrt{\frac{3^{3}\cdot27\cdot 1000C(d+\ln{\left(e\cu/\delta \right)})\ls_{\cD}(\widehat{\cA}_{t_{2}}(S_{2}))}{26\cdot999m'}} \nonumber \\
& \leq\sqrt{\frac{29C(d+\ln{\left(e\cu/\delta \right)})\ls_{\cD}(\widehat{\cA}_{t_{2}}(S_{2}))}{m'}}.
\end{align*}
In the second inequality we have used that $ s_{\sqcap}=\frac{3^{3}}{1-\frac{1}{27}}.$ 
We now use the assumption that we had realizations $ S_{2} $ and $ t_{2} $ of  $ \rS_{2} $ and $ \rt_{2} $ in $ E_{2} $, i.e., such that $\ls_{\cD}(\widehat{\cA}_{t_{2}}(S_{2}))\leq  c\tau+\frac{c\left(d+\ln{(\cu e/\delta )}\right)}{m'}$. This allows us to conclude that
\begin{align}\label{eq:secondmajority6}
\sqrt{\frac{4C(d+\ln{\left(e\cu/\delta \right)})\ls_{\cD}(\widehat{\cA}_{t_{2}}(S_{2}))}{m'}}
   &\leq \sqrt{\frac{4C(d+\ln{\left(e\cu/\delta \right)})\left(c\tau+\frac{c\left(d+\ln{(\cu e/\delta )}\right)}{m'}\right)}{m'}}
   \\
   &\leq \sqrt{\frac{4cC\tau(d+\ln{\left(e\cu/\delta \right)})}{m'}}+\frac{4cC(d+\ln{\left(e\cu/\delta \right)})}{m'},
\end{align}
where we have used the inequality $ \sqrt{a+b}\leq \sqrt{a}+\sqrt{b} $ for $ a,b\geq0 $ in the last step.

Then, to summarize, we have seen that the event $\rN_{i}\geq (1-1/1000)\p[A]m'/s_{\sqcap}=(1-1/1000)\ls_{\cD}(\widehat{\cA}_{t_{2}}(S_{2}))m'/s_{\sqcap}$ occurs with probability at least $ 1-\delta/\cu $, and that conditioned on this event, \cref{eq:secondmajority4} holds with probability at least $ 1-\delta/\cu $. Consequently, we have that with probability at least $ 1-2\delta/\cu $ over $ \rS_{1,i} $ each of \cref{eq:secondmajority4}, \cref{eq:secondmajority5} and \cref{eq:secondmajority6} hold. Then, with probability at least $ 1-2\delta/\cu $ over $ \rS_{1,i} $ we have that
\begin{align*}
& \frac{243}{232}\max_{h\in \cA(\rS_{1,i,\sqcup};\rS_{1,i,\sqcap})}\p_{(\rx,\ry)\sim \cD}\left[h(x)\not=y,\avg(\widehat{\cA}_{t_{2}}(S_{2}))(\rx,\ry)\geq \frac{11}{243}\right] \\
& \quad \leq \frac{243}{232} \frac{1000\cdot27}{999\cdot26}\ls_{\rS_{1,i}}(\hs)+\sqrt{\frac{4cC\tau(d+\ln{\left(e\cu/\delta \right)})}{m'}}+\frac{4cC(d+\ln{\left(e\cu/\delta \right)})}{m'}
    \\
& \quad \leq 1.0888\ls_{\rS_{1,i}}(\hs)+\sqrt{\frac{4cC\tau(d+\ln{\left(e\cu/\delta \right)})}{m'}}+\frac{4cC(d+\ln{\left(e\cu/\delta \right)})}{m'}.
\end{align*} 
Now, invoking a union bound over $ i\in \{ 1,\ldots,27 \} $, we have that with probability at least $ 1-54\delta/\cu $ over $ \rS_{1} $, it holds that
\begin{align}\label{eq:secondmajority7}
& \frac{243}{232}\max_{h\in \cA(\rS_{1};\emptyset)}\p_{(\rx,\ry)\sim \cD}\left[h(x)\not=y,\avg(\widehat{\cA}_{t_{2}}(S_{2}))(\rx,\ry)\geq \frac{11}{243}\right] \nonumber \\
&\quad \leq 1.0888\max_{i\in\{  1,\ldots,27\} }\ls_{\rS_{1,i}}(\hs)+\sqrt{\frac{4cC\tau(d+\ln{\left(e\cu/\delta \right)})}{m'}}+\frac{4cC(d+\ln{\left(e\cu/\delta \right)})}{m'}. 
\end{align}
Furthermore, by \cref{lem:additiveerrorhstar} and another union bound over $ \rS_{1,1},\ldots,\rS_{1,27} $, we have that with probability at least $ 1-27\delta/\cu $ over $ \rS_{1}\sim \cD^{m'}$ it holds that

\begin{align}\label{eq:secondmajority8}
    \max_{i\in\{  1,\ldots,27\} }\ls_{\rS_{1,i}}(\hs)\leq \tau+\sqrt{\frac{2\tau \ln{(\cu /\delta )}}{3m'}}+\frac{2\ln{(\cu /\delta )}}{m'}.
\end{align}
Thus, by applying the union bound over the events in \cref{eq:secondmajority7} and \cref{eq:secondmajority8}, we get that with probability at least $ 1-81\delta/\cu $ over $ \rS_{1} $, it holds that
\begin{align}\label{eq:secondmajority10}
& \frac{243}{232}\max_{h\in \cA(\rS_{1};\emptyset)}\p_{(\rx,\ry)\sim \cD}\left[h(x)\not=y,\avg(\widehat{\cA}_{t_{2}}(S_{2}))(\rx,\ry)\geq \frac{11}{243}\right] \nonumber
    \\
& \quad \leq 1.0888\left(\tau+\sqrt{\frac{2\tau \ln{(\cu /\delta )}}{3m'}}+\frac{2\ln{(\cu /\delta )}}{m'}\right)+\sqrt{\frac{4cC\tau(d+\ln{\left(e\cu/\delta \right)})}{m'}}+\frac{4cC(d+\ln{\left(e\cu/\delta \right)})}{m'} \nonumber 
    \\
& \quad \leq  1.0888\tau+\sqrt{\frac{16cC\tau(d+2\ln{\left(3\cu/\delta \right)})}{m'}}+\frac{6cC(d+\ln{\left(e\cu /\delta \right)})}{m'}.
\end{align}
Note that this suffices to give the event $ E_{G} $ in \cref{eq:secondmajority0}.
We remark that we demonstrated the above for any realizations $ S_{2} $ and $ t_{2} $ of $ \rS_{2} $ and $ \rt_{2}$ in $ E_{2}.$

We now notice that by \cref{thm:main}, the fact that $ t_{1}\geq 4\cdot243^{2}\ln{\left(2m/(\delta(d+\ln{\left(\cur/\delta \right)})) \right)}$, and the choice of $ \cu=\cur $), we have that $ \p_{\rS_{2},\rt_{2}}[E_{3}]\leq \delta/\cu.$ 
Combining this with the conclusion below \cref{eq:secondmajority9} and \cref{eq:secondmajority10}, and with the fact that $ \rS_{1},$  $ \rS_{2} $ $ \rt_{2} $ are independent, we have that 
\begin{align*}
 \p_{\rS_{1},\rS_{2},\rt_{2}}\left[E_{G}\right]
& =\e_{\rS_{2},\rt_{2}}\left[\p_{\rS_{1}}\left[E_{G}\right]\ind\{   E_{1}\}\right]+\e_{\rS_{2},\rt_{2}}\left[\p_{\rS_{1}}\left[E_{G}\right]\ind\{   E_{2}\}\right]+\e_{\rS_{2},\rt_{2}}\left[\p_{\rS_{1}}\left[E_{G}\right]\ind\{   E_{3}\}\right]
 \\
 & \geq  \p_{\rS_{2},\rt_{2}}\left[ E_{1}\right]+(1-81\delta/\cu)\p_{\rS_{2},\rt_{2}}\left[ E_{2}\right] \\
 &\geq(1-81\delta/\cu)(1-\p_{\rS_{2},\rt_{2}}[E_{3}]) \\
 & \geq1-82\delta/\cu.
\end{align*}
Note that first equality follows from $ E_{1},E_{2},E_{3} $ partitioning the outcomes of $ \rS_{2} $ and $ \rt_{2}$ and the first inequality follows from the conclusions below \cref{eq:secondmajority9} and \cref{eq:secondmajority10}, which state that $ E_{G} $ holds with probability $ 1 $ on $ E_{1} $ and with probability at least $ 1-81\delta/\cu$ on $ E_{2} $. The second inequality again follows from $ E_{1},E_{2},E_{3} $ partitioning the outcomes of $ \rS_{2} $ and $ \rt_{2}$ and the bound $ \p_{\rS_{2},\rt_{2}}[E_{3}]\leq \delta/\cu,$ which shows \cref{eq:secondmajority-1}.

We now proceed to show \cref{eq:secondmajority-2}, i.e., that with probability at least $ 1-4\delta/\cu $ over $ \rS_{1},\rS_{2},\rS_{3},\rt_{1},\rt_{2},$ it holds that
\begin{align}\label{eq:secondmajority11}
& \p_{(\rx,\ry)\sim\cD}\left[ \htie\not=\ry, \avg(\widehat{\cA}_{\rt_{1}}(\rS_{1}))(\rx,\ry)\geq \frac{11}{243} \text{ or } \avg(\widehat{\cA}_{\rt_{2}}(\rS_{2}))(\rx,\ry)\geq \frac{11}{243}\right] \nonumber \\
& \quad \leq 
    \tau+ \sqrt{\frac{4\tau cC'(d+\ln{\left(\cu/\delta \right)})}{m'}}+\frac{5cC'\ln{\left(\cu e/\delta \right)}}{m}.
\end{align}
We denote this event $ E_{F}.$ 
Towards proving the claim, we consider the following event over $ \rS_{1},\rS_{2},\rt_{1}$ and $ \rt_{2}$,
\begin{align*}
        E_{4} = \left\{ \p_{(\rx,\ry)\sim \cD}\left[ \avg(\widehat{\cA}_{\rt_{1}}(\rS_{1}))(\rx,\ry)\geq \frac{11}{243} \text{ or } \avg(\widehat{\cA}_{\rt_{2}}(\rS_{2}))(\rx,\ry)\geq \frac{11}{243}\right]\leq  \frac{c\ln{\left(\cu e/\delta \right)}}{m'} \right\}. 
\end{align*}
As previously mentioned, we take $ c $ to be at least the constant of \cref{thm:main} and to also satisfy $ c\geq 2\cdot10^{6}s_{\sqcap} $.

Now, if we have realizations $ S_{1},S_{2},t_{1} $ and $ t_{2} $ of $ \rS_{1},$ $ \rS_{2},$ $ \rt_{1} $ and $ \rt_{2} $ in $ E_{4},$ then by monotonicity of measures, we obtain that
\begin{align*}
    \p_{(\rx,\ry)\sim\cD}\left[\htie\not=\ry,  \avg(\widehat{\cA}_{t_{1}}(S_{1}))(\rx,\ry)\geq \frac{11}{243} \text{ or } \avg(\widehat{\cA}_{t_{2}}(S_{2}))(\rx,\ry)\geq \frac{11}{243}\right]\leq \frac{c\ln{\left(\cu e/\delta \right)}}{m'},
\end{align*}
which would imply the event $ E_{F} $ in \cref{eq:secondmajority11}.
Now consider a realization of $ S_{1},S_{2},t_{1} $ and $ t_{2} $ of $ \rS_{1},$ $ \rS_{2},$ $ \rt_{1} $ and $ \rt_{2} $ on the complement $ \bar{E}_{4} $ of $ E_{4}.$ 
We then have that 
\[ \p_{(\rx,\ry)\sim\cD}\left[ \avg(\widehat{\cA}_{t_{1}}(S_{1}))(\rx,\ry)\geq 11/243 \text{ or } \avg(\widehat{\cA}_{t_{2}}(S_{2}))(\rx,\ry)\geq 11/243\right] \geq \frac{c\ln{\left(\cu e/\delta \right)}}{m'}. \] 
We recall that $ \rS_{3}^{\not=},$ are the examples in $(x,y)\in \rS_{3}$ for which  
\[ \avg(\widehat{\cA}_{t_{1}}(S_{1}))(x,y)\geq \frac{11}{243} \text{ or } \avg(\widehat{\cA}_{t_{2}}(S_{2}))(x,y)\geq \frac{11}{243}. \] 
In the following, let $ \cD(\, \cdot \, | \not=) $ be the conditional distribution of $ \cD $ given that $\avg(\widehat{\cA}_{t_{1}}(S_{1}))(x,y)\geq \frac{11}{243}$ or $\avg(\widehat{\cA}_{t_{2}}(S_{2}))(x,y)\geq \frac{11}{243}$. That is, for an event $ B $ over $ (\cX\times \cY),$ we have
\begin{align*}
  \cD(B|\not=)=\frac{\p_{(\rx,\ry)\sim \cD}\left[(\rx,\ry)\in B, \avg(\widehat{\cA}_{t_{1}}(S_{1}))(\rx,\ry)\geq \frac{11}{243} \text{ or } \avg(\widehat{\cA}_{t_{2}}(S_{2}))(\rx,\ry)\geq \frac{11}{243}\right]}{\p_{(\rx,\ry)\sim \cD}\left[ \avg(\widehat{\cA}_{t_{1}}(S_{1}))(\rx,\ry)\geq \frac{11}{243} \text{ or } \avg(\widehat{\cA}_{t_{2}}(S_{2}))(\rx,\ry)\geq \frac{11}{243}\right]}
\end{align*}     
Thus we have that $ \rS_{3}^{\not=}\sim \cD(B|\not=).$ 
We now notice by $ \rS_{3}\sim \cD $ that
\begin{align*}
\e_{\rS_{3}}\left[|\rS_{3}^{\not=}|\right]=m'\p_{(\rx,\ry)\sim\cD}\left[\avg(\widehat{\cA}_{t_{1}}(S_{1}))(\rx,\ry)\geq \frac{11}{243} \text{ or } \avg(\widehat{\cA}_{t_{2}}(S_{2}))(\rx,\ry)\geq \frac{11}{243}\right].
\end{align*}  
Using a Chernoff bound, this implies 
\begin{align*}
& \hspace{-1cm} \p_{\rS_{3}}\left[|\rS_{3}^{\not=}|\geq m'\p_{(\rx,\ry)\sim\cD}\left[\avg(\widehat{\cA}_{t_{1}}(S_{1}))(\rx,\ry)\geq \frac{11}{243} \text{ or } \avg(\widehat{\cA}_{t_{2}}(S_{2}))(\rx,\ry)\geq \frac{11}{243}\right]/2\right]
 \\
&  \geq 1- \exp{\left(-m'\p_{(\rx,\ry)\sim\cD}\left[\avg(\widehat{\cA}_{t_{1}}(S_{1}))(\rx,\ry)\geq \frac{11}{243} \text{ or } \avg(\widehat{\cA}_{t_{2}}(S_{2}))(\rx,\ry)\geq \frac{11}{243}\right]/8 \right)} \\
&\geq 1-\delta/\cu.  
\end{align*}   
Note that we are also using the facts that 
\[ \p_{(\rx,\ry)\sim\cD}\left[\avg(\widehat{\cA}_{t_{1}}(S_{1}))(\rx,\ry)\geq \frac{11}{243} \text{ or } \avg(\widehat{\cA}_{t_{2}}(S_{2}))(\rx,\ry)\geq \frac{11}{243}\right] \geq \frac{c\ln{\left(\cu e/\delta \right)}}{m'} \]
and that $ c\geq 2\cdot1000^{2} s_{\sqcap}$.   
Now consider an outcome of $ N_{\not=}=|\rS_{3}^{\not=}| $ where 
\[ N_{\not=} \geq \frac{m'}{2} \p_{(\rx,\ry)\sim\cD}\left[\avg(\widehat{\cA}_{t_{1}}(S_{1}))(\rx,\ry)\geq \frac{11}{243} \text{ or } \avg(\widehat{\cA}_{t_{2}}(S_{2}))(\rx,\ry)\geq \frac{11}{243}\right]. \] 
Now, by \cref{thm:ermtheoremunderstanding} we have that since $ \htie=\cA(\rS_{3}^{\not=})$ and $ \rS_{3}^{\not=}\sim \cD(\cdot | \not=) $, then with probability at least $ 1-\delta/\cu $ over $ \rS_{3}^{\not=} $,
\begin{align*}
\p_{(\rx,\ry)\sim\cD(\cdot|\not=)}\left[ \htie(\rx)\not=\ry\right] &\leq \inf_{h\in\cH}\ls_{\cD(\cdot|\not=)}(h)+\sqrt{\frac{C'(d+\ln{\left(\cu/\delta \right)})}{N_{\not=}}} \\
& \leq \ls_{\cD(\cdot|\not=)}(\hs)+\sqrt{\frac{C'(d+\ln{\left(\cu/\delta \right)})}{N_{\not=}}}.   
\end{align*}
Note that the first inequality uses \cref{thm:ermtheoremunderstanding} (and $ C'>1 $  is the universal constant of \cref{thm:ermtheoremunderstanding}), and the second inequality uses that $ \hs\in \cH $ so it has error greater than the infimum. 
Now, using the law of total expectation, we have that  
\begin{align*}
& \p_{(\rx,\ry)\sim\cD}\left[ \htie(\rx)\not=\ry,\avg(\widehat{\cA}_{t_{1}}(S_{1}))(\rx,\ry)\geq \frac{11}{243} \text{ or } \avg(\widehat{\cA}_{t_{2}}(S_{2}))(\rx,\ry)\geq \frac{11}{243}\right]\\
& \quad = \p_{(\rx,\ry)\sim\cD(\cdot|\not=)}\left[ \htie(\rx)\not=\ry\right]\p_{(\rx,\ry)\sim\cD}\left[\avg(\widehat{\cA}_{t_{1}}(S_{1}))(\rx,\ry)\geq \frac{11}{243} \text{ or } \avg(\widehat{\cA}_{t_{2}}(S_{2}))(\rx,\ry)\geq \frac{11}{243}\right]
    \\
& \quad \leq \left(\ls_{\cD(\cdot|\not=)}(\hs)+\sqrt{\frac{C'(d+\ln{\left(\cu/\delta \right)})}{N_{\not=}}}\right)   \\
& \qquad \times \p_{(\rx,\ry)\sim\cD}\left[\avg(\widehat{\cA}_{t_{1}}(S_{1}))(\rx,\ry)\geq \frac{11}{243} \text{ or } \avg(\widehat{\cA}_{t_{2}}(S_{2}))(\rx,\ry)\geq \frac{11}{243}\right]
\end{align*}
We now bound each term in the above. 
First, 
\begin{align*}
& \ls_{\cD(\cdot|\not=)}(\hs)  \p_{(\rx,\ry)\sim\cD}\left[\avg(\widehat{\cA}_{t_{1}}(S_{1}))(\rx,\ry)> 11/243 \text{ or } \avg(\widehat{\cA}_{t_{2}}(S_{2}))(\rx,\ry)> 11/243\right] \\
& \quad =\p_{\rx,\ry\sim\cD}\left[\hs(\rx)\not=\ry,\widehat{\cA}_{t_{1}}(S_{1})(\rx)\not=\widehat{\cA}_{t_{2}}(S_{2})(\rx)\right] 
\end{align*}
which is less than $ \tau.$ 
Furthermore, for the second term, we have by 
\[ N_{\not=}\geq \frac{m'}{2}\p_{(\rx,\ry)\sim\cD}\left[\avg(\widehat{\cA}_{t_{1}}(S_{1}))(\rx,\ry)> 11/243 \text{ or } \avg(\widehat{\cA}_{t_{2}}(S_{2}))(\rx,\ry)> 11/243\right] \] 
that 
\begin{align*}
& \sqrt{\frac{C'(d+\ln{\left(\cu/\delta \right)})}{N_{\not=}}}\p_{(\rx,\ry)\sim\cD}\left[\avg(\widehat{\cA}_{t_{1}}(S_{1}))(\rx,\ry)\geq \frac{11}{243} \text{ or } \avg(\widehat{\cA}_{t_{2}}(S_{2}))(\rx,\ry)\geq \frac{11}{243}\right]
    \\
& \quad \leq \sqrt{\frac{2C'(d+\ln{\left(\cu/\delta \right)})\left(\ls_{\cD}^{11/243}(\widehat{\cA}_{t_{1}}(S_{1})) +\ls_{\cD}^{11/243}(\widehat{\cA}_{t_{2}}(S_{2}))\right)}{m'}}
    \end{align*}    
where the first inequality follows from plugging in that 
\[ N_{\not=}\geq \frac{m'}{2} \p_{(\rx,\ry)\sim\cD}\left[\avg(\widehat{\cA}_{t_{1}}(S_{1}))(\rx,\ry)> 11/243 \text{ or } \avg(\widehat{\cA}_{t_{2}}(S_{2}))(\rx,\ry)> 11/243\right] \]  in the denominator, and the second inequality follows from a union bound over the event 
\[ \avg(\widehat{\cA}_{t_{1}}(S_{1}))(\rx,\ry)> 11/243 \text{ or } \avg(\widehat{\cA}_{t_{2}}(S_{2}))(\rx,\ry)> 11/243 .\]  
Thus, we have shown that with probability at least $ 1-2\delta/\cu $ over $\rS_{3}$, it holds that 
\begin{align*}
    \p_{(\rx,\ry)\sim\cD}\left[ \htie(\rx)\not=\ry,\avg(\widehat{\cA}_{t_{1}}(S_{1}))(\rx,\ry)\geq \frac{11}{243} \text{ or } \avg(\widehat{\cA}_{t_{2}}(S_{2}))(\rx,\ry)\geq \frac{11}{243}\right]
    \\
    \leq \tau+ \sqrt{\frac{2C'(d+\ln{\left(\cu/\delta \right)})\left(\ls_{\cD}(\widehat{\cA}_{t_{1}}(S_{1})) +\ls_{\cD}(\widehat{\cA}_{t_{2}}(S_{2}))\right)}{m'}}
\end{align*}
which, in particular, also implies that with probability at least $ 1-2\delta/\cu $ over $ \rS_{3} $ 
\begin{align*}
    \p_{(\rx,\ry)\sim\cD}\left[ \htie(\rx)\not=\ry,\avg(\widehat{\cA}_{t_{1}}(S_{1}))(\rx,\ry)\geq \frac{11}{243} \text{ or } \avg(\widehat{\cA}_{t_{2}}(S_{2}))(\rx,\ry)\geq \frac{11}{243}\right]
    \\
    \leq \tau+ \sqrt{\frac{2C'(d+\ln{\left(\cu/\delta \right)})\left(\ls_{\cD}(\widehat{\cA}_{t_{1}}(S_{1})) +\ls_{\cD}(\widehat{\cA}_{t_{2}}(S_{2}))\right)}{m'}}+\frac{c\ln{\left(\cu e/\delta \right)}}{m}.
\end{align*}
Thus, since the above also upper bounds 
\[ \p_{(\rx,\ry)\sim\cD}\left[ \htie(\rx)\not=\ry,\avg(\widehat{\cA}_{t_{1}}(S_{1}))(\rx,\ry)\geq \frac{11}{243} \text{ or } \avg(\widehat{\cA}_{t_{2}}(S_{2}))(\rx,\ry)\geq \frac{11}{243}\right] \]
for any realization $ S_{1},S_{2},t_{1} $ and $ t_{2} $ of $ \rS_{1},\rS_{2},\rt_{1} $, and $ \rt_{2},$ on $ E_{4},$ (with probability $ 1 $), we conclude that for any realization  $S_{1},S_{2},t_{1} $, and $ t_{2} $ of $ \rS_{1},\rS_{2},\rt_{1} $, and $ \rt_{2},$ it holds with probability at least $ 1-2\delta/\cu $ over $ \rS_{3} $ that 
\begin{align*}
    \p_{(\rx,\ry)\sim\cD}\left[ \htie(\rx)\not=\ry,\avg(\widehat{\cA}_{t_{1}}(S_{1}))(\rx,\ry)\geq \frac{11}{243} \text{ or } \avg(\widehat{\cA}_{t_{2}}(S_{2}))(\rx,\ry)\geq \frac{11}{243}\right]
    \\
    \leq \tau+ \sqrt{\frac{2C'(d+\ln{\left(\cu/\delta \right)})\left(\ls_{\cD}(\widehat{\cA}_{t_{1}}(S_{1})) +\ls_{\cD}(\widehat{\cA}_{t_{2}}(S_{2}))\right)}{m'}}+\frac{c\ln{\left(\cu e/\delta \right)}}{m}.
\end{align*}
Let this event be denoted $ E_{5} $. 
Now, let $ E_{6} $ be the event that
\begin{align*}
 E_{6}= \left \{ \ls_{\cD}(\widehat{\cA}_{t_{1}}(S_{1})) +\ls_{\cD}(\widehat{\cA}_{t_{2}}(S_{2})) \leq 2 \left(c\tau+\frac{c\left(d+\ln{(\cu e/\delta )}\right)}{m'} \right) \right\} 
\end{align*} 
which, by \cref{thm:main}, the fact that $ t_{1},t_{2}\geq 4\cdot243^{2}\ln{\left(2m/(\delta(d+\ln{\left(\cur/\delta \right)})) \right)} $, and a union bound, holds with probability at least $ 1-2\delta/\cu $. (We let $ c>1 $ be a universal constant at least as large as the universal constant of \cref{thm:main}, and with $ c\geq 2\cdot10^{6}s_{\sqcap} $.) 
Now, notice that for realizations $S_{1},S_{2},t_{1} $ and $ t_{2} $ of $ \rS_{1},\rS_{2},\rt_{1} $ and $ \rt_{2},$ on $ E_{6} $,  it holds with probability at least $ 1-2\delta/\cu $ over $ \rS_{3} $ that 
\begin{align*}
& \p_{(\rx,\ry)\sim\cD}\left[ \htie(\rx)\not=\ry,\avg(\widehat{\cA}_{t_{1}}(S_{1}))(\rx,\ry)\geq \frac{11}{243} \text{ or } \avg(\widehat{\cA}_{t_{2}}(S_{2}))(\rx,\ry)\geq \frac{11}{243}\right]
    \\
& \qquad \leq \tau+ \sqrt{\frac{2C'(d+\ln{\left(\cu/\delta \right)})\left(\ls_{\cD}(\widehat{\cA}_{t_{1}}(S_{1})) +\ls_{\cD}(\widehat{\cA}_{t_{2}}(S_{2}))\right)}{m'}}+\frac{c\ln{\left(\cu e/\delta \right)}}{m}
    \\
& \qquad \leq\tau+ \sqrt{\frac{4C'(d+\ln{\left(\cu/\delta \right)})\left(c\tau+\frac{c\left(d+\ln{(\cu e/\delta )}\right)}{m'}\right)}{m'}}+\frac{c\ln{\left(\cu e/\delta \right)}}{m}
    \\
& \qquad \leq
    \tau+ \sqrt{\frac{4\tau cC'(d+\ln{\left(\cu/\delta \right)})}{m'}}+\frac{5cC'\ln{\left(\cu e/\delta \right)}}{m}.
\end{align*}
where the first inequality follows by $ E_{5},$ the second inequality by $ E_{6},$ and the third inequality by $ \sqrt{a+b}\leq \sqrt{a}+\sqrt{b}.$ 
We notice that the above event is $ E_{F} $ in \cref{eq:secondmajority11}. 
Thus, by the above holding with probability at least $ 1-2\delta/\cu $ for any outcome of $ S_{1},S_{2},t_{1} $ and $ t_{2} $ of $ \rS_{1},\rS_{2},\rt_{1} $ and $ \rt_{2}$ on $ E_{6} $, and by $ \rS_{1},\rS_{2}$ and $\rS_{3} $ being independent  we conclude that 
\begin{align*}
 \p_{\rS_{1},\rS_{2},\rS_{3},\rt_{1},\rt_{2}}\left[ E_{F}\right] 
 & \geq\e_{\rS_{1},\rS_{2},\rS_{3},\rt_{1},\rt_{2}}\left[ \p_{\rS_{3}}\left[E_{F},E_{5}\right]\ind\{  E_{6}\} \right] \\
 &=\e_{\rS_{1},\rS_{2},\rS_{3},\rt_{1},\rt_{2}}\left[ \p_{\rS_{3}}\left[E_{5}\right]\ind\{  E_{6}\} \right]
 \\
 &\geq\e_{\rS_{1},\rS_{2},\rS_{3},\rt_{1},\rt_{2}}\left[ (1-2\delta/\cu)\ind\{  E_{6}\} \right] \\
 & \geq (1-2\delta/\cu)^{2} \\ & \geq 1-4\delta/\cu. 
\end{align*} 
Note that the equality follows from our previous reasoning, i.e., that for outcomes of $ S_{1},S_{2},t_{1} $ and $ t_{2} $ of $ \rS_{1},\rS_{2},\rt_{1} $ and $ \rt_{2}$ on $ E_{6} $, and outcomes $ S_{3} $  of $ \rS_{3} $ on $ E_{5} $, $ E_{F} $ holds. The second inequality follows from the fact that $E_{5}$ holds with probability at least $ 1-2\delta/\cu $ over $ \rS_{3} $  for any realizations $ S_{1},S_{2},t_{1} $ and $ t_{2} $ of $ \rS_{1},\rS_{2},\rt_{1} $ and $ \rt_{2}$. The third inequality follows from the fact that $ E_{6} $ holds with probability at least $ 1-2\delta/\cu $ for $ \rS_{1},\rS_{2}, \rt_{1} $ and $ \rt_{2},$ which concludes the proof of \cref{eq:secondmajority-2}, as desired.      
\end{proofof}

\section{Best-of-both-worlds Learner}

In this section we demonstrate that splitting a  training sample $ \rS \sim \cD^{m} $ into $\{ \rS_i \sim \cD^{m/3}\}_{i \in [3]}$ followed by running the algorithm of \citet{hanneke2024revisiting} on $ \rS_{1} $, running the algorithm of \cref{thm:main} on $ \rS_{2} $, and selecting the one $\hmin$ with smallest empirical error on $ \rS_{3} $  gives the following error bound:
\begin{align*}
\ls_{\cD}(\hmin) = \min\Bigg(&2.1 \cdot \tau+O\left(\sqrt{\frac{\tau(d+\ln{\left(1/\delta \right)})}{m}}+\frac{\left(d+\ln{\left(1/\delta \right)}\right)}{m}\right), \\
&\tau+O\left(\sqrt{\frac{\tau(d+\ln{\left(1/\delta \right)})}{m}}+\frac{\ln^{5}{\left( m/d\right)}\left(d+\ln{\left(1/\delta \right)}      \right)}{m} \right)\Bigg).
\end{align*}

In pursuit of the above, recall the error bound of \cite[Theorem~3]{hanneke2024revisiting}, which establishes the existence of a learner $ \tilde{\cA} $ which with probability at least $ 1-\delta  $ over $ \rS \sim \cD^m$ incurs error at most
\begin{align*}
    \ls_{\cD}(\tilde{\cA}(\rS))\leq \tau+\sqrt{\frac{c'\tau(d+\ln{\left(1/\delta \right)})}{m}}+\frac{c'\ln^{5}{\left( m/d\right)}\left(d+\ln{\left(1/\delta \right)}      \right)}{m},
\end{align*}
for a universal constant $c' \geq 1$. 
We will in the following use $ \widehat{\cA}_{\rt} $ to denote the algorithm of \Cref{thm:tiemajority}. In what follows let $m'=m/3.$

By invoking the previous bounds on  $ \widehat{\cA}_{\rt} $ and $ \tilde{\cA} $ with $\delta = \delta/4$, and further employing a union bound, we have that with probability at least $1 - \delta/2$ over $\rS_1$ and $\rS_2$, both $\tilde{\cA}$ and $\widehat{\cA}_{\rt}$ will emit hypotheses satisfying: 
\begin{align}\label{eq:bestofbothworld0}
 \ls_{\cD}(\tilde{\cA}(\rS_{1})) &\leq \tau+\sqrt{\frac{c'\tau(d+\ln{\left(4/\delta \right)})}{m'}}+\frac{c'\ln^{5}{\left( m'/d\right)}\left(d+\ln{\left(4/\delta \right)}      \right)}{m'} \\
 \ls_{\cD}(\widehat{\cA}_{\rt}(\rS_{2};\emptyset)) &\leq 2.1 \cdot \tau+\sqrt{\frac{c\tau(d+\ln{\left(4/\delta \right)})}{m'}}+\frac{c\left(d+\ln{\left(4/\delta \right)}\right)}{m'}, \label{eq:bestofbothworld0.5}
\end{align}
where $c$ is the universal constant of \Cref{thm:tiemajority}. Let $G$ denote the event that $\rS$ satisfies the previous condition, which, as we have noted, has probability at least $1 - \delta / 2$. 

Now consider realizations $ S_{1} $ and $ S_{2} $  of $\rS_1$ and $\rS_2$ with $(\rS_1,\rS_2) \in G$. Further, let $ \hmin=\argmin_{h'\in\{   \tilde{\cA}(S_{1}),\widehat{\cA}(S_{2};\emptyset)\}}\left(\ls_{\rS_{3}}(h')\right)$. 
We now invoke \Cref{lem:additiveerrorhstar} on $ \rS_{3}$ with the classifiers $ \tilde{\cA}(S_{1})$ and $\widehat{\cA}_{t}(S_{2};\emptyset)$ (and failure probability $\delta/8$) along with a union bound to see that with probability at least  $ 1-\delta/2 $  over $ \rS_{3} $, we have that both choices of  $ h\in\{   \tilde{\cA}(S_{1}),\widehat{\cA}(S_{2};\emptyset)\}$ satisfy 

\begin{align}\label{eq:bestofbothworld2}
    \ls_{\cD}(h)\leq \ls_{\rS_{3}}(h) +\sqrt{\frac{2\ls_{\rS_{3}}(h)\ln{(8/\delta )}}{m'}}+\frac{4\ln{(8/\delta )}}{m'},
    \end{align}
and
\begin{align}\label{eq:bestofbothworld1}
\ls_{\rS_{3}}(h)\leq \ls_{\cD}(h) +\sqrt{\frac{2\ls_{\cD}(h)\ln{(8/\delta )}}{3m'}}+\frac{2\ln{(8/\delta )}}{m'}\leq 2\ls_{\cD}(h')+\frac{4\ln{\left(8/\delta \right)}}{m'}. 
\end{align}
Note that the final inequality follows from the fact that $ \sqrt{ab}\leq a+b $ for $ a,b>0.$ 
Then, using the previous inequalities along with the definition of $\hmin$, we have that: 
\begin{align}\label{eq:bestofbothworld4}
 \ls_{\cD}(\hmin) &\leq  \ls_{\rS_{3}}(\hmin) +\sqrt{\frac{2\ls_{\rS_{3}}(\hmin)\ln{(8/\delta )}}{m'}}+\frac{4\ln{(8/\delta )}}{m'}
 \\
&\leq \min_{h\in\{   \tilde{\cA}(S_{1}),\widehat{\cA}(S_{2};\emptyset)\}}\left(\ls_{\rS_{3}}(h) +\sqrt{\frac{2\ls_{\rS_{3}}(h)\ln{(8/\delta )}}{m'}}+\frac{4\ln{(8/\delta )}}{m'}\right)
 \\
& \leq \min_{h\in\{   \tilde{\cA}(S_{1}),\widehat{\cA}(S_{2};\emptyset)\}}\Bigg(\ls_{\cD}(h) +\sqrt{\frac{2\ls_{\cD}(h)\ln{(8/\delta )}}{m'}}+\frac{6\ln{(8/\delta )}}{m'}  \nonumber \\
& \qquad + \sqrt{\frac{2\left( 2\ls_{\cD}(h)+4\ln{\left(8/\delta \right)}/m' \right)\ln{\left(8/\delta \right)}}{m'}}\Bigg)
 \\
& \leq  \min_{h\in\{   \tilde{\cA}(S_{1}),\widehat{\cA}(S_{2};\emptyset)\}}\left(\ls_{\cD}(h) + 3 \sqrt{\frac{2\ls_{\cD}(h)\ln{(8/\delta )}}{m'}}+\frac{9\ln{(8/\delta )}}{m'}\right).
\end{align}
In particular, the first inequality follows from \cref{eq:bestofbothworld2} applied to $\hmin$, the second from the definition of $\hmin$, the third from \cref{eq:bestofbothworld1}, and the fourth from the fact that $ \sqrt{a+b}\leq \sqrt{a}+\sqrt{b}$ for $ a,b>0.$ 
Now using the equations in \cref{eq:bestofbothworld0} we have that 
\begin{align}\label{eq:bestofbothworld3}
\vspace{-1 cm}
    \ls_{\cD}(\tilde{\cA}(\rS_{1}))\leq \tau+\sqrt{\frac{c'\tau(d+\ln{\left(4/\delta \right)})}{m'}}+\frac{c'\ln^{5}{\left( m'/d\right)}\left(d+\ln{\left(4/\delta \right)}      \right)}{m'} 
    \leq 2\tau+\frac{2c'\ln^{5}{\left( m'/d\right)}\left(d+\ln{\left(4/\delta \right)}      \right)}{m'}
\end{align}
since $ \sqrt{ab}\leq a+b $ for $ a,b>0$. (We are further assuming that $m = \Omega(d)$, a standard condition.) Then, using \cref{eq:bestofbothworld3} and the fact that $ \sqrt{a+b}\leq \sqrt{a}+\sqrt{b} $ for $ a,b>0$, we have 
\begin{align*}
 \sqrt{\frac{2\ls_{\cD}(\tilde{\cA}(\rS_{1}))\ln{\left(8/\delta \right)}}{m'}} 
 \leq  \sqrt{\frac{2\left(2\tau+2\frac{c'\ln^{5}{\left( m'/d\right)}\left(d+\ln{\left(4/\delta \right)}      \right)}{m'}\right)\ln{\left(8/\delta \right)}}{m'}}
\\
\leq \sqrt{\frac{4\tau\ln{\left(8/\delta \right)}}{m'}}+
\frac{2c'\ln^{5}{\left( m'/d\right)}\left(d+\ln{\left(8/\delta \right)}      \right)}{m'}. 
\end{align*}
Using the previous inequality along with \cref{eq:bestofbothworld0}, we have that 
\begin{align*}
&\ls_{\cD}(\tilde{\cA}(\rS_{1})) + 3 \sqrt{\frac{2\ls_{\cD}(\tilde{\cA}(\rS_{1}))\ln{(8/\delta )}}{m'}}+\frac{9\ln{(8/\delta )}}{m'}   \\
& \quad \leq  \tau+\sqrt{\frac{c'\tau(d+\ln{\left(4/\delta \right)})}{m'}}+\frac{c'\ln^{5}{\left( m'/d\right)}\left(d+\ln{\left(4/\delta \right)}    \right)}{m'}+ 3 \sqrt{\frac{4\tau\ln{\left(8/\delta \right)}}{m'}} \\
& \qquad \quad  + \frac{6 c'\ln^{5}{\left( m'/d\right)}\left(d+\ln{\left(8/\delta \right)} \right)}{m'} + \frac{9\ln{\left(8/\delta \right)}}{m'} \\
& \quad \leq \tau + 7 \sqrt{\frac{c'\tau(d+\ln{\left(8/\delta \right)})}{m'}}+\frac{16 c'\ln^{5}{\left( m'/d\right)}\left(d+\ln{\left(8/\delta \right)}      \right)}{m'}. 
\end{align*}
Similarly, using \cref{eq:bestofbothworld0.5} and $ \sqrt{ab}\leq a+b $  we have that 
\begin{align*}
    \ls_{\cD}(\widehat{\cA}_{\rt}(\rS_{2};\emptyset))\leq  3.1 \tau+\frac{2c\left(d+\ln{\left(4/\delta \right)}\right)}{m'}.
\end{align*}
Again using the fact that $\sqrt{a+b}\leq \sqrt{a}+\sqrt{b} $ for $ a,b>0$, we have that 
\begin{align*}
    \sqrt{\frac{2\ls_{\cD}(\widehat{\cA}_{\rt}(\rS_{2};\emptyset))\ln{(8/\delta )}}{m'}}
    \leq\sqrt{\frac{6.2 \tau\ln{\left(8/\delta \right)}}{m'}}+\frac{2c(d+\ln{\left(8/\delta \right)})}{m'}. 
\end{align*}
Applying \cref{eq:bestofbothworld0.5} then yields
\begin{align*}
& \ls_{\cD}(\widehat{\cA}_{\rt}(\rS_{2};\emptyset)) + 3\sqrt{\frac{2\ls_{\cD}(\widehat{\cA}_{\rt}(\rS_{2};\emptyset))\ln{(8/\delta )}}{m'}}+\frac{9\ln{(8/\delta )}}{m'} \\
& \quad \leq 2.1 \tau+\sqrt{\frac{c\tau(d+\ln{\left(4/\delta \right)})}{m'}}+\frac{c\left(d+\ln{\left(4/\delta \right)}\right)}{m'}+ 3 \sqrt{\frac{6.2 \tau\ln{\left(8/\delta \right)}}{m'}}+\frac{6c(d+\ln{\left(8/\delta \right)})}{m'}+\frac{9\ln{\left(8/\delta \right)}}{m'} \\
& \quad \leq 2.1 \tau+ 9\sqrt{\frac{c\tau(d+\ln{\left(8/\delta \right)})}{m'}}+\frac{16 c(d+\ln{\left(8/\delta \right)})}{m'}.
\end{align*}
Now plugging in the above expressions into \cref{eq:bestofbothworld4} which is exactly the minimum of   $ \tilde{\cA}(\rS_{1}) $ and $ \widehat{\cA}_{\rt}(\rS_{2};\emptyset) $ we arrive at 
\begin{align}\label{eq:bestofbothworld5}
\ls_{\cD}(\hmin)= \min\Bigg(&2.1 \tau + 9 \sqrt{\frac{c\tau(d+\ln{\left(1/\delta \right)})}{m'}}+\frac{16 c\left(d+\ln{\left(1/\delta \right)}\right)}{m'}, \nonumber \\
& \tau + 7 \sqrt{\frac{c'\tau(d+\ln{\left(1/\delta \right)})}{m'}}+\frac{16 c'\ln^{5}{\left( m'/d\right)}\left(d+\ln{\left(1/\delta \right)}      \right)}{m'} \Bigg)
\end{align}
with probability at least $ 1-\delta/2 $ over $ \rS_{3} $ for any realizations  $ S_{1} $ and $ S_{2} $ of $ \rS_{1} $ and $ \rS_{2} $   in $ G $, now let $ T $ denote the event of \cref{eq:bestofbothworld5}, then we have by independence of $ \rS_{1},\rS_{2},\rS_{3} $ that 
\begin{align*}
 \p_{\rS\sim \cD^{m}}[T]
 \geq
 \e_{\rS_{1},\rS_{2}\sim \cD^{m/3}}\left[\p_{\rS_{3}\sim \cD^{m/3}}\left[\ind\{ T \} \right]\ind\{  G\} \right]
 \geq 
 \e_{\rS_{1},\rS_{2}\sim \cD^{m/3}}\left[(1-\delta/2)\ind\{  G\} \right]
\end{align*}
where the first inequality follows by independence of $ \rS_{1},\rS_{2},\rS_{3} $ and $ G $ only depending upon $ \rS_{1},\rS_{2}$ and the second inequality by \cref{eq:bestofbothworld5} holding (the event of $ T $ ) with probability at least $ 1-\delta/2$ over $ \rS_{3}$ for any $ \rS_{1},\rS_{2} $ in $ G,$. Lastly, note that 
\[  \e_{\rS_{1},\rS_{2}\sim \cD^{m/3}} \left[(1-\delta/2)\ind\{  G\} \right] \geq (1-\delta/2)(1-\delta/2)\geq1-\delta, \]
due to $G$ having probability at least $ 1-\delta/2$ over $ \rS_{1},\rS_{2} $ by \cref{eq:bestofbothworld0}, which concludes the proof.

\end{document}